\documentclass[runningheads]{llncs}

\usepackage{eccv}

\usepackage{multirow}
\usepackage{xcolor,colortbl}
\definecolor{tabfirst}{rgb}{1, 0.6, 0.6} % red
\definecolor{tabsecond}{rgb}{1, 0.8, 0.6} % orange
\definecolor{tabthird}{rgb}{1, 1, 0.6} % yellow

\usepackage{eccvabbrv}

\usepackage{graphicx}
\usepackage{booktabs}

\usepackage[accsupp]{axessibility}  % Improves PDF readability for those with disabilities.

\usepackage{cite}

\usepackage{hyperref}

\usepackage{orcidlink}

\begin{document}

% ---------------------------------------------------------------
% TODO REVIEW: Replace with your title
\title{Benchmarking Efficient \& Effective Camera Pose Estimation Strategies for Novel View Synthesis} 

% TODO REVIEW: If the paper title is too long for the running head, you can set
% an abbreviated paper title here. If not, comment out.
\titlerunning{Abbreviated paper title}

% TODO FINAL: Replace with your author list. 
% Include the authors' OCRID for the camera-ready version, if at all possible.
% Orcid M.R. Oswald: 0000-0002-1183-9958
\author{Jhacson Meza\inst{1,2} \and
Martin R. Oswald\inst{3} \and
Torsten Sattler\inst{2}}

% TODO FINAL: Replace with an abbreviated list of authors.
\authorrunning{Meza et al.}
% First names are abbreviated in the running head.
% If there are more than two authors, 'et al.' is used.

% TODO FINAL: Replace with your institution list.
\institute{Faculty of Electrical Engineering, Czech Technical University in Prague \and
Czech Institute of Informatics, Robotics and Cybernetics, Czech Technical University in Prague \\
%\email{\{jhacson.meza,torsten.sattler\}@cvut.cz}\\ 
\and University of Amsterdam\\
%\email{m.r.oswald@uva.nl}
}

\maketitle

%%%%%%%%%%%%%%%%%%%%%%%%%%%%%%%%%%%%%%%%%%%%%%%%%%%%%%%%%%%%%%%%%%%%%%%
%%%%%%%%%%%%%%%%%%%%%%%%%%%%%%%%%%%%%%%%%%%%%%%%%%%%%%%%%%%%%%%%%%%%%%%
%%%%%%%%%%%%%%%%%%%%%%%%%%%%%%%%%%%%%%%%%%%%%%%%%%%%%%%%%%%%%%%%%%%%%%%

\begin{abstract}
Novel view synthesis (NVS) approaches such as NeRFs or 3DGS can produce photo-realistic 3D scene representation from a set of images with known extrinsic and intrinsic parameters. The necessary camera poses and calibrations are typically obtained from the images via Structure-from-Motion (SfM). Classical SfM approaches rely on local feature matches between the images to estimate both the poses and a sparse 3D model of the scene, using bundle adjustment to refine initial pose, intrinsics, and geometry estimates. In order to increase run-time efficiency, recent SfM systems forgo optimization via bundle adjustment. Instead, they train feed-forward (transformer-based) neural networks to directly regress camera parameters and the 3D structure. While orders of magnitude more efficient, such recent works produce significantly less accurate estimates. To stimulate research on developing SfM approaches that are both efficient \emph{and} effective, this paper develops a benchmark focused on SfM for novel view synthesis. Using existing datasets and two simple strategies for making the reconstruction process more efficient, we show that: (1) simply using fewer features already significantly accelerates classical SfM methods while maintaining high pose accuracy. (2) using feed-forward networks to obtain initial estimates and refining them using classical SfM techniques leads to the best efficiency-effectiveness trade-off. Our benchmark and code are publicly available at \href{https://github.com/jhacsonmeza/sfm-benchmark-nvs}{https://github.com/jhacsonmeza/sfm-benchmark-nvs}.
\end{abstract}

%%%%%%%%%%%%%%%%%%%%%%%%%%%%%%%%%%%%%%%%%%%%%%%%%%%%%%%%%%%%%%%%%%%%%%%
%%%%%%%%%%%%%%%%%%%%%%%%%%%%%%%%%%%%%%%%%%%%%%%%%%%%%%%%%%%%%%%%%%%%%%%
%%%%%%%%%%%%%%%%%%%%%%%%%%%%%%%%%%%%%%%%%%%%%%%%%%%%%%%%%%%%%%%%%%%%%%%

\section{Introduction}
\label{sec:intro}
Given a set of images, Structure-from-Motion (SfM) approaches estimate both the camera poses, \ie, the positions and orientations from which the images were taken, and the 3D structure of the scene (and potentially the intrinsic calibrations of the cameras). 
SfM systems are a vital part of many computer vision applications, \eg, they provide 3D maps 
for visual localization~\cite{Sattler2017PAMI,Sarlin2019CVPR} and the camera poses needed for novel view synthesis (NVS) via Neural Radiance Fields (NeRFs)~\cite{mildenhall2021nerf} or 3D Gaussian Splatting (3DGS)~\cite{kerbl20233d}. 
Ideally, SfM approaches should be both efficient and effective, \ie, they should produce accurate camera parameter and scene structure estimates as run-time-efficient as possible. 

Traditionally, SfM approaches~\cite{schonberger2016structure, pan2024global,Agarwal2009ICCV,Frahm10ECCV,Wu11CVPR}, are typically based on local image features: 
Given local features~\cite{lowe2004distinctive,detone2018superpoint,zhao2023aliked,potje2024xfeat}, \etc, extracted from images, 2D-2D matches between the images are established. 
These 2D-2D matches are then used to estimate the relative poses between the images. 
In turn, the relative poses are leveraged to estimate the absolute poses of the images (up to a similarity transformation) as well as the 3D structure of the scene (by triangulating 3D points from the 2D-2D matches). 
In the absolute pose estimation stage, poses are estimated by either incrementally adding more images to an initial 3D reconstruction (incremental SfM~\cite{schonberger2016structure,Agarwal2009ICCV,Frahm10ECCV}), interleaved with repeated bundle adjustment~\cite{Wu11CVPR,Triggs2000VATP,Safari_2025_CVPR}, or by jointly estimating all poses together (global SfM~\cite{pan2024global,cui2015global,Crandall2011CVPR}). 
Such classical approaches are highly effective in terms of the accuracy of their camera and scene geometry estimates, in large part due to using bundle adjustment. 
However, they can also be rather inefficient, especially when compared to other domains in vision where feed-forward networks are used. 
The (perceived) inefficiency of classical SfM methods has motivated recent work on using feed-forward networks to regress camera and scene parameters~\cite{wang2025vggt}. 
These approaches train transformer architectures on massive amounts of posed images to learn generalizeable pose and structure predictors. 
As estimating scene structure and camera parameters is done via a single pass through a feed-forward network, these approaches are highly efficient. 
Yet, our experimental results show that these methods are significantly less accurate than classical approaches. 

This work proposes a benchmark to foster work on efficient \emph{and} effective SfM techniques. 
We focus on novel view synthesis (NVS) as an application scenario, as NVS requires highly accurate camera pose estimates. 
At the same time, NVS approaches such as 3DGS can produce highly accurate 3D models in a matter of minutes, \ie, efficient camera pose estimation is important as well in this scenario. 
Our benchmark is based on existing datasets for 3D reconstruction~\cite{schops2017multi,knapitsch2017tanks} and NVS~\cite{barron2022mip,barron2023zip}. 
We measure pose estimation time per method to determine efficiency and use pose accuracy and NVS performance to measure effectiveness. 
Besides existing feed-forward-based~\cite{yang2025fast3r,duisterhof2025mast3r,wang2025vggt,wang2025pi,lin2025depth,wang2026vggt} and classical SfM approaches~\cite{Schoenberger2017CVPR,pan2024global}, we evaluate two simple strategies for efficient and effective SfM: 
(1) The number of local features extracted per image directly impacts the run-time efficiency of subsequent matching and reconstruction stages in classical SfM. 
The first strategy thus simply uses fewer features and we benchmark multiple types of (learned) local features commonly used in the literature. 
(2) The second strategy aims to make feed-forward approaches more effective by using their estimates as initialization for classical SfM techniques: 
Given the initial pose estimates, local feature matching is restricted to nearby images. 
The initial estimates are then refined using bundle adjustment using these feature matches.

Concretely, we make the following contributions: 
(1) We design a benchmark to analyze the NVS quality-reconstruction time tradeoff. 
Our benchmark is built around existing easy datasets for 3D reconstruction~\cite{schops2017multi,knapitsch2017tanks} and NVS~\cite{barron2022mip,barron2023zip}. 
We measure reconstruction effectiveness through both the quality of the reconstructed poses and the NVS performance. 
(2) Extensive experiments with two strategies for efficient and effective SfM, multiple (learned) local features, two classical SfM pipelines, and two matching strategies lead to multiple interesting insights: 
(a) Classical approaches can be trivially accelerated by simply using fewer features, often at a limited reduction in effectiveness. Decent reconstruction quality, at significantly reduced run-times, can be achieved using as few as 512 features per image. 
(b) Feed-forward-based methods can be made significantly more effective by refining their pose estimate using local features, while retaining competitive runtimes compared with classical SfM. 
The resulting strategy offers the best efficiency-effectiveness trade-off in our experiments. 
The results show that, even with the advent of feed-forward reconstruction approaches, classical SfM techniques are still relevant. 
(c) Classical SIFT~\cite{lowe2004distinctive} features are surprisingly competitive, and not all learned features outperform SIFT. 
Still, learned features enable faster reconstructions by using fewer features. 
(3) We will make our benchmark, including code for the two strategies, publicly available.

%%%%%%%%%%%%%%%%%%%%%%%%%%%%%%%%%%%%%%%%%%%%%%%%%%%%%%%%%%%%%%%%%%%%%%%
%%%%%%%%%%%%%%%%%%%%%%%%%%%%%%%%%%%%%%%%%%%%%%%%%%%%%%%%%%%%%%%%%%%%%%%
%%%%%%%%%%%%%%%%%%%%%%%%%%%%%%%%%%%%%%%%%%%%%%%%%%%%%%%%%%%%%%%%%%%%%%%

\section{Related Work}
\label{sec:related}

Traditionally, local features have been evaluated using metrics such as the repeatability of the feature detector or how well their descriptors can be matched~\cite{Balntas2017CVPR,Mikolajczyk04IJCV,Tuytelaars08}. 
These works consider features individually. 
However, the distribution of features and matches in images has a large impact on how well the geometric relation between images can be estimated. 
As such, classical metrics are not necessarily a good indicator of performance in practical scenarios such as SfM or visual localization~\cite{Schoenberger2017CVPR}. 

Rather than measuring repeatability rates, descriptor matchability, \etc in isolation, newer benchmarks for local features thus focus on performance in higher-level applications such as relative pose estimation~\cite{Jin2020}, absolute pose estimation~\cite{Sattler2018CVPR,Toft2022TPAMI,Dusmanu2019CVPR,Humenberger2022IJCV}, or Structure-from-Motion~\cite{Schoenberger2017CVPR,IMC2025}. 
They define features as useful if they improve metrics measuring the performance of such higher-level systems, \eg, relative or absolute pose accuracy, or SfM-related statistics (such as the number of images registered in a SfM reconstruction, the number of reconstructed points, the average track length of reconstructed points, \etc). 

Most related to our work are the local feature benchmarks from~\cite{Schoenberger2017CVPR,Jin2020,IMC2025}, which also evaluate the performance of local features in the context of SfM. 
Schoenberger \etal~\cite{Schoenberger2017CVPR} evaluate SfM methods based on SfM metrics (like number of registered images, number of 3D points, number of observations, etc.), with the number of dense Multi-View Stereo (MVS) points as a proxy for pose accuracy due to a lack of ground truth poses. It does not take reconstruction speed into account. We report results on datasets that have highly accurate poses. In addition, we use NVS quality as a better proxy for pose accuracy than outdated MVS techniques. As such, our evaluation protocol represents an updated and more complete evaluation compared to \cite{Schoenberger2017CVPR}, where we additionally consider reconstruction times and feed-forward SfM approaches.

Jin \etal~\cite{Jin2020} do not evaluate full SfM systems but rather relative pose accuracy. They ask the question to what degree learned features and matchers can help to make relative pose estimation more accurate and robust. They use SfM to obtain poses for larger sets of images and treat the relative poses between pairs of images obtained from SfM as ground truth. In contrast to~\cite{Schoenberger2017CVPR,Jin2020}, we aim at benchmarking SfM approaches in terms of their efficiency-run-time tradeoff in the context of NVS. This is a gap we want to fill in this work.

The Image Matching Challeng (IMC)~\cite{IMC2025} aims at robust SfM given a set of images (that can contain unrelated images, as is the case in the context of 3D reconstruction from photo community collections). Their evaluation criteria are (1) the ability to identify outlier images (which is typically not an issue for NVS), and (2) the accuracy of the estimated camera centers. They do not consider NVS quality as a metric, and they do not take run-time into consideration (there is a run-time component as IMC entries need to share code and the code gets executed, with a time limit, but efficiency is not rewarded).

If ground truth poses are not available, the performance of local features in the context of Structure-from-Motion can also be measured indirectly through statistics related to dense 3D reconstruction~\cite{Schoenberger2017CVPR} or Novel View Synthesis (NVS)~\cite{brachmann2024acezero}.  
The underlying principle is that better features will lead to better poses (and better geometry). 
In turn, better pose estimates lead to better 3D reconstructions~\cite{Schoenberger2017CVPR}, as measurable by the number of 3D points in dense MVS models computed from the poses. 
Similarly, better pose estimates lead to better NVS results~\cite{brachmann2024acezero}. 
For our benchmark, we follow the literature and report both pose accuracy (when ground truth poses are available) and NVS quality (when ground truth poses are not available). 

Overall, our benchmark has a different focus than the existing ones, with a wider range of metrics and evaluation criteria. 
We ask the question whether the reconstruction process can be significantly accelerated by using fewer features, and aim to analyze the trade-off between reconstruction time and reconstruction accuracy. 
We believe that these are highly relevant questions in practice and are not aware of benchmarks trying to answer these questions. 
Furthermore, none of the existing benchmarks evaluate feed-forward 3D reconstruction systems, nor do they characterize the accuracy-efficiency trade-off that motivates their use in the first place. Our benchmark addresses this gap by directly comparing feed-forward and classical SfM approaches in terms of reconstruction time, pose accuracy, and downstream NVS quality.

%%%%%%%%%%%%%%%%%%%%%%%%%%%%%%%%%%%%%%%%%%%%%%%%%%%%%%%%%%%%%%%%%%%%%%%
%%%%%%%%%%%%%%%%%%%%%%%%%%%%%%%%%%%%%%%%%%%%%%%%%%%%%%%%%%%%%%%%%%%%%%%
%%%%%%%%%%%%%%%%%%%%%%%%%%%%%%%%%%%%%%%%%%%%%%%%%%%%%%%%%%%%%%%%%%%%%%%

\section{Benchmark}
\label{sec:benchmark}
Our benchmark is divided into two parts. We first investigate the \textit{impact of the number of extracted features} on the sparse reconstruction runtime and the reconstruction quality in classical SfM. For that, we evaluate multiple (learned) local features in terms of camera pose accuracy and NVS with NeRF and 3DGS. In our study, we consider the following learned feature extractors: ALIKED~\cite{zhao2023aliked}, DISK~\cite{tyszkiewicz2020disk}, DoG-HardNet~\cite{mishchuk2017working}, SIFT~\cite{lowe2004distinctive} (OpenCV's implementation), SuperPoint~\cite{detone2018superpoint}, and XFeat~\cite{potje2024xfeat}. 
For feature matching, we use LightGlue (LG)~\cite{lindenberger2023lightglue} and Mutual Nearest Neighbors (MNN). For 3D mapping, we use the incremental SfM pipeline COLMAP~\cite{schonberger2016structure} and the global SfM pipeline GLOMAP~\cite{pan2024global}. Note that OpenCV's SIFT is included in our evaluation, even though it is not a learned keypoint extractor, because of the weights provided by LightGlue for SIFT descriptors. We generate reconstructions extracting 128, 256, 512, 1024, 2048, 4096, and 8192 keypoints from each learned local feature. That means that for each extractor-matcher-mapper triplet, we generate 7 reconstructions of the same scene, one for each number of extracted features. In total, for a single scene we generate 168 sparse reconstructions considering all the number of features, extractors, matchers, and mappers. In addition to the reconstructions obtained with learned features, we also use the feature extractor and matcher from the COLMAP pipeline and use the COLMAP and GLOMAP mappers to get a sparse reconstruction.
We refer to these results as standard COLMAP and standard GLOMAP. 
By default, COLMAP's feature extractor detects 8192 features per image; however, we also vary the number of extracted features as with the learned features. This increases to 182 the total number of reconstructions for a specific scene.

In the second part of our benchmark, we evaluate multiple feed-forward SfM models for the NVS task. We use the camera poses and point clouds provided by the models and run 3DGS on top of them. We also propose a \textit{feed-forward refinement} strategy: 
The pose estimates provided by a feed-forward SfM approach are used to identify other images taken nearby. 
For each image, we extract local features and restrict feature matching to these nearby images. 
The feature matches and initial pose estimates are then used to triangulate a sparse SfM point cloud. 
Finally, we bundle adjust the camera poses, intrinsics, and point positions, and use the refined positions to merge 3D points (if they belong to feature tracks that were initially split into multiple 3D points due to inaccuracies in the initial poses and due to thresholds in the triangulation process) and complete 3D points (\ie, adding observations that were not originally included in the triangulation). 
We iterate this alternating process (bundle adjustment and merging / completing) for a fixed number of iterations or until we see insufficient change in terms of merged / completed 3D points.  
As we will show, this strategy considerably improve NVS effectiveness, both compared to feed-forward approaches and standard COLMAP, while maintaining efficiency.

\subsection{Datasets:}
For our evaluation, we use standard datasets for NVS evaluation and a dataset with accurate absolute camera poses. These datasets are captured in controlled environments with good overlapping between views, consistent illumination conditions, and with no abrupt viewpoint changes. 
We use the ETH3D Multi-View Stereo (MVS) dataset~\cite{schops2017multi}, which consists of 13 training and 12 test indoor and outdoor scenes captured by a DSLR camera. This dataset comes with highly accurate camera poses which we use as ground truths to evaluate the different learned local features in terms of camera accuracy. 
For NVS evaluation, we use the Mip-NeRF 360 dataset~\cite{barron2022mip}, which 
consists of 5 outdoor and 4 indoor scenes captured with the same high-resolution camera with fixed camera exposure. This dataset includes sparse 3D reconstructions obtained with COLMAP, but because their accuracy is limited, we do not consider them for pose accuracy evaluation. We also use the Tanks and Temples dataset~\cite{knapitsch2017tanks} that consists of 21 indoor and outdoor scenes captured as high-resolution video sequences. Further, we use the Zip-NeRF dataset~\cite{barron2023zip}, which is a more large-scale dataset compared to the previous ones, which consists of 4 scenes with thousands of images captured with fisheye cameras.

\subsection{Evaluation metrics:} For camera pose accuracy evaluation, we use ground truth camera poses to estimate absolute pose accuracies as the percentage of views with angular errors and position errors lower than a given threshold. 
For novel view synthesis, we compute the standard PSNR, SSIM, and LPIPS metrics. 
For timings, we measure wall-clock time for each individual stage of the 3D reconstruction process.
For the first part of our benchmark, we measure times for feature extraction, feature matching, geometric verification, and 3D mapping.
The final reconstruction time reported is the sum of these individual stages.
For the second part of our benchmark with the feed-forward models, we report the inference time as the reconstruction time of the models, and for the proposed refinement strategy, we report the sum of the inference, feature extraction, matching, and BA refinement times as the final reconstruction time.

%%%%%%%%%%%%%%%%%%%%%%%%%%%%%%%%%%%%%%%%%%%%%%%%%%%%%%%%%%%%%%%%%%%%%%%
%%%%%%%%%%%%%%%%%%%%%%%%%%%%%%%%%%%%%%%%%%%%%%%%%%%%%%%%%%%%%%%%%%%%%%%
%%%%%%%%%%%%%%%%%%%%%%%%%%%%%%%%%%%%%%%%%%%%%%%%%%%%%%%%%%%%%%%%%%%%%%%

\section{Experimental evaluation}
\label{sec:experiments}
In this section, we describe the experiments we carried out to investigate for efficient camera pose estimation strategies for NVS. We first evaluate the \textit{impact of the number of extracted features} on camera pose accuracy and runtime and in the NVS quality. We then evaluate the 
\textit{feed-forward refinement} strategy, where the camera poses provided by an SfM feed-forward model are refined to improve the NVS quality.
For our experiments, we use a NVIDIA A100 GPU and two AMD Zen 3 EPYC (2.45 GHz) with a total of 128 CPU cores. Considering the large number of reconstructions in the first part of our benchmark with classical SfM, we distributed multiple concurrent reconstructions into these 128 cores.

\subsection{Evaluating the impact of the number of extracted features}
\label{sec:experiments:absolute}

\subsubsection{Camera pose accuracy evaluation: }
To evaluate the impact of the number of extracted features on the accuracy of the camera poses in SfM, we compare multiple learned local features and use millimeter-accurate ground truth camera poses from the ETH3D MVS dataset. 
For each scene, we first estimate the camera rotation and position errors of all the registered views after globally aligning the obtained reconstruction to the ground truth. Then, considering all the views from all the 25 scenes, we estimate the accuracy for each extractor-matcher-mapper triplet as the percentage of views with camera rotation errors and camera position errors under a given threshold with respect to the total number of views across all the scenes. By considering the total number of views we account for non-registered views and penalize those methods that fail to register images in the reconstruction.

\begin{figure*}
    \centering
    \includegraphics[width=1\textwidth]{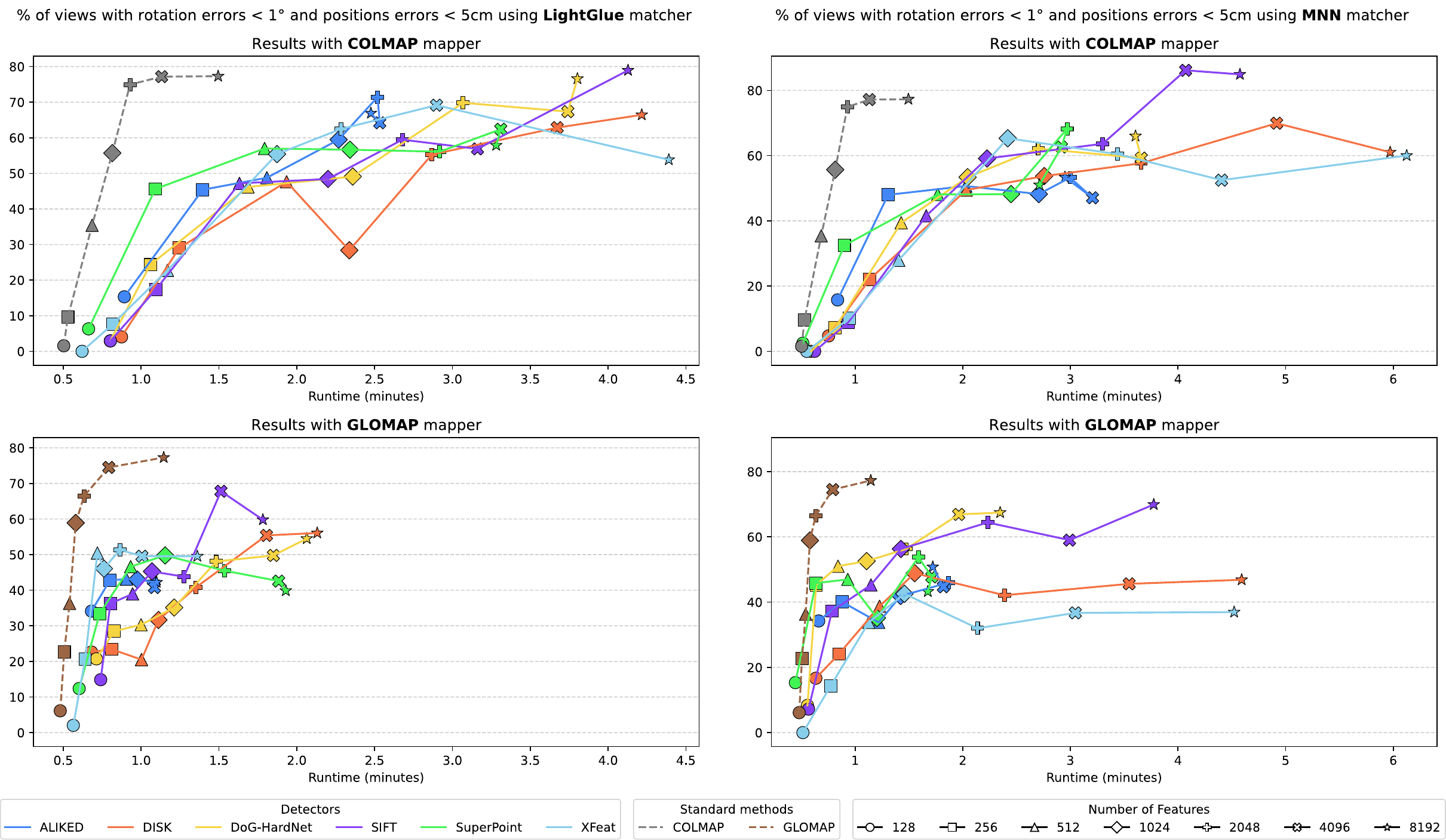}
    \caption{Average absolute camera pose accuracy vs reconstruction time across the 25 scenes from the ETH3D MVS dataset using LightGlue and MNN matchers and COLMAP and GLOMAP mappers. Runtimes are average across all the scenes. For the same matcher, both time and accuracy scales are fixed to allow a direct comparison between the mappers.}
    \label{fig:ETH3D_acc_time}
\end{figure*}

Figure~\ref{fig:ETH3D_acc_time} shows absolute pose accuracy results and the total reconstruction time using LG and MNN matchers, and COLMAP and GLOMAP mappers for camera rotation errors below 1$^\circ$ and camera position errors below 5~cm. We compute the average runtime through all the scenes, and for those cases where the 3D mapping fails (which usually happens for 128 and 256 features), we use a time of 0 minutes in that stage to be able to compute the average and generate the plot. Note that for the same feature matcher, the x- and y-axes ranges are identical for the two mappers to allow for a direct comparison.

As expected, the runtime increases with the number of extracted features. Overall, COLMAP is slightly better than GLOMAP in this camera pose accuracy evaluation. 
Note that some learned features achieve the best accuracy with 2048 features, and others with 4096. 
Interestingly, good results can be achieved already using as little as 256 features per image, resulting in reconstruction times of around 1-2 minutes. 
Overall, SuperPoint is a good choice for ETH3D, independently of the setup (COLMAP \vs GLOMAP, LG \vs MNN). For COLMAP, ALIKED is a (close) second choice.

\subsubsection{Novel view synthesis evaluation:}
\label{subsec:3dgs}
We also evaluate the impact of the number of extracted features on the novel view synthesis task using both 3DGS and NeRF. 
As in the camera pose evaluation, we want to know how the different number of extracted features affect the NVS quality and the reconstruction runtime. For this evaluation, we generate 6,188 SfM reconstructions in total, each of which is used to train both a 3DGS and a NeRF model, resulting in a total of 12,376 trained models.

For the 3DGS evaluation, we use Mip-Splatting~\cite{yu2024mip} because it provides a good trade-off between synthesis quality and training time. 
For our evaluation, we estimate the PSNR, SSIM, and LPIPS metrics for the \textit{Mip-NeRF 360}, \textit{Tanks and Temples}, and \textit{Zip-NeRF} datasets. We have 34 scenes considering all these 3 datasets, which means that we generate 6,188 sparse reconstructions and train 6,188 3DGS models (considering the 182 reconstructions/scene number previously reported in Sec.~\ref{sec:experiments}).
We consistently use the same train and test split of images across the different numbers of extracted keypoints. To compute the metrics for a single scene, when test views are missing because of an unsuccessful registration during the incremental reconstruction process, we assign dummy values (0 for PSNR and SSIM, 1 for LPIPS) to these missing views. 
This penalizes non-registered views in each individual reconstruction used to train a 3DGS model. 
For a given extractor-matcher-mapper tuple, the estimated metrics are aggregated across all the scenes by computing the mean. 
In those cases where the training of a scene fails (because the point cloud was almost empty or very sparse, \eg, for 128 and 256 features), we penalize using the same dummy values mentioned before for the whole scene. Runtimes are aggregated across all the scenes as previously described in the camera pose accuracy evaluation.

\begin{figure*}
    \centering
    \includegraphics[width=1\textwidth]{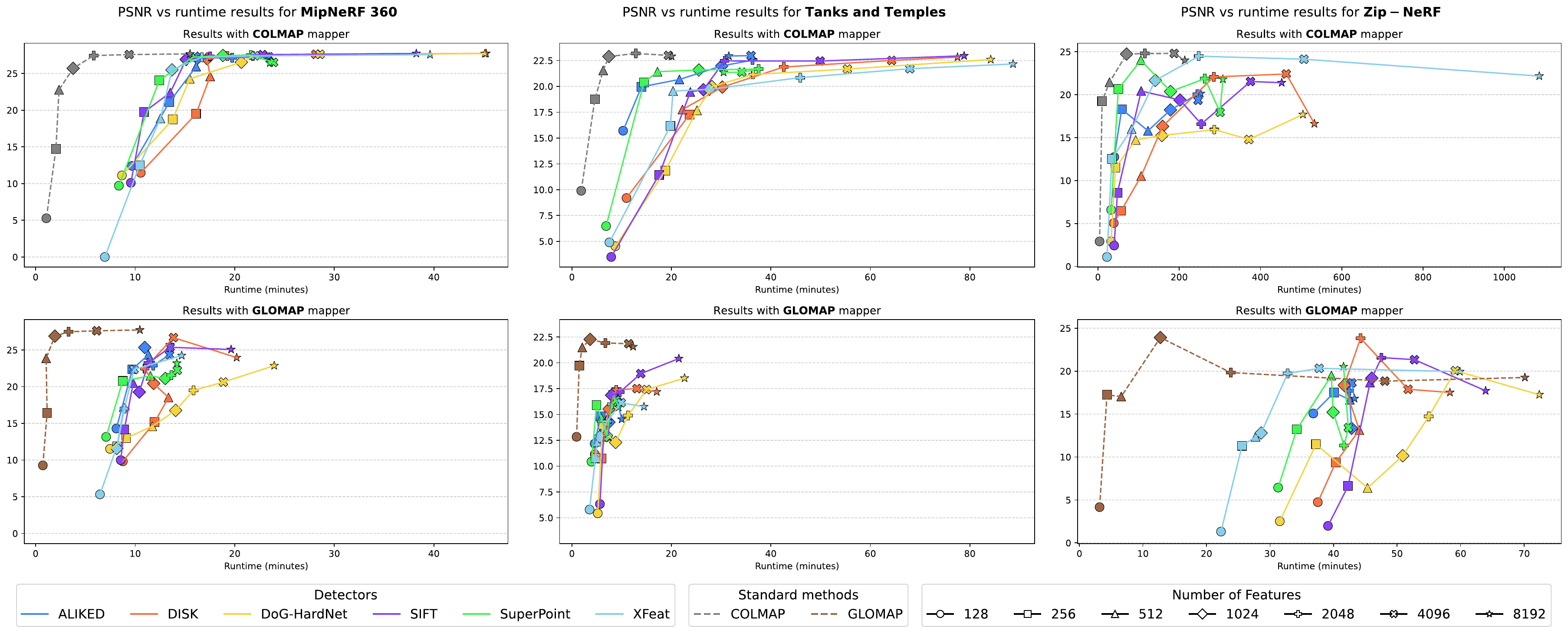}
    \caption{PSNR using 3DGS vs. reconstruction runtime results with the LightGlue matcher for Mip-NeRF 360, Tanks and Temples, and Zip-NeRF datasets. Missing views for the average PSNR computation for a scene are penalized with a value of 0. Scenes that failed to reconstruct or render are also penalize with 0 for the per scene aggregation.}
    \label{fig:3data_LG_psnr_time}
\end{figure*}

Figure~\ref{fig:3data_LG_psnr_time} shows the PSNR vs reconstruction runtime results with 3DGS using LG matcher, and COLMAP and GLOMAP mappers, for the three datasets. To better visualize the runtime and PSNR differences between the two mappers, we use the same x- and y-axis ranges for the same matcher in Mip-NeRF 360 and Tanks and Temples. For Zip-NeRF, only the y-axis is the same. The results for SSIM and LPIPS metrics, and the rendering results using MNN matcher are provided in the supplementary material.

To obtain the standard COLMAP and GLOMAP reconstructions, we use an exhaustive matching approach in all the three datasets. 
For the learned features, we use an exhaustive matching strategy for Mip-NeRF 360 dataset, and a sequential matching approach for Tanks and Temples, and Zip-NeRF datasets. 
Overall, from the plots we can see that standard COLMAP and GLOMAP are faster than using custom features. This is associated with the highly optimized feature extraction and matching algorithms implemented in COLMAP.

From the Mip-NeRF 360 results, we can see that the average runtime difference between 4096 and 8192 extracted features is small for ALIKED and SuperPoint detectors (and in some cases even negligible), while for the other feature extractors the difference tends to be larger. 
This is because these learned features often cannot produce more keypoints beyond a certain number, even if we request more. 
This maximum number of features is typically around 4000, which explains why for 4096 and 8192, the average runtime is almost the same. We can also see that SuperPoint tends to be faster than the other features (especially for fewer keypoints), except when it is used with GLOMAP. 
From the Tanks and Temples results, we can see a similar behavior to the one obtained with Mip-NeRF 360 dataset: SuperPoint often offers the best performance-reconstruction time trade-off. 
Extracting as few as 512 features per image for learned feature extractors is often already sufficient, as using more than 512 features often increases the run-time without significantly increasing the novel view synthesis quality. 
As for the previous experiments, we note that SIFT, when integrated in the same matching pipeline as the learned features (purple lines in the plots), performs worse than the best learned features. 
From the Zip-NeRF plot, we can see that the time gap between standard COLMAP and the learned features is smaller compared to the previous datasets but it is not the case with GLOMAP. This is because of the sequential matching strategy for the learned feature detectors vs COLMAP's exhaustive matcher for the standard COLMAP and GLOMAP results. Following the previous datasets, we again observe that using 512 SuperPoint features per image offers a very good trade-off between reconstruction times and novel view synthesis quality, independently of the used setup. 
We also observe that XFeat is much more competitive compared to previous datasets. 
XFeat extracts lower-dimensional descriptors (64-dimensional, compared to 256-dimensional SuperPoint descriptors and 128-dimensional descriptors for the other features). 
This results in faster matching times, which are noticeable at the scale of the Zip-NeRF dataset, which is a dataset with significantly more images (offering thousands of images per scene) than Mip-NeRF 360 and Tanks and Temples datasets. Also, because of the large scale nature of Zip-NeRF, we can see a more important average reconstruction time difference between the different number of features.  

As in the camera pose evaluation, we observe that 512 features are often enough, offering a good trade-off between sparse reconstruction times and novel view quality. 
This is an interesting observation since Mip-Splatting depends on the quality of the sparse SfM point cloud used to initialize the Gaussians. 
More features typically result in more 3D points, \ie, one would expect to achieve better results using more features. 
Yet, the gains are often rather small. 
Again, SuperPoint features are a good choice. 

\begin{figure*}
    \centering
    \includegraphics[width=1\textwidth]{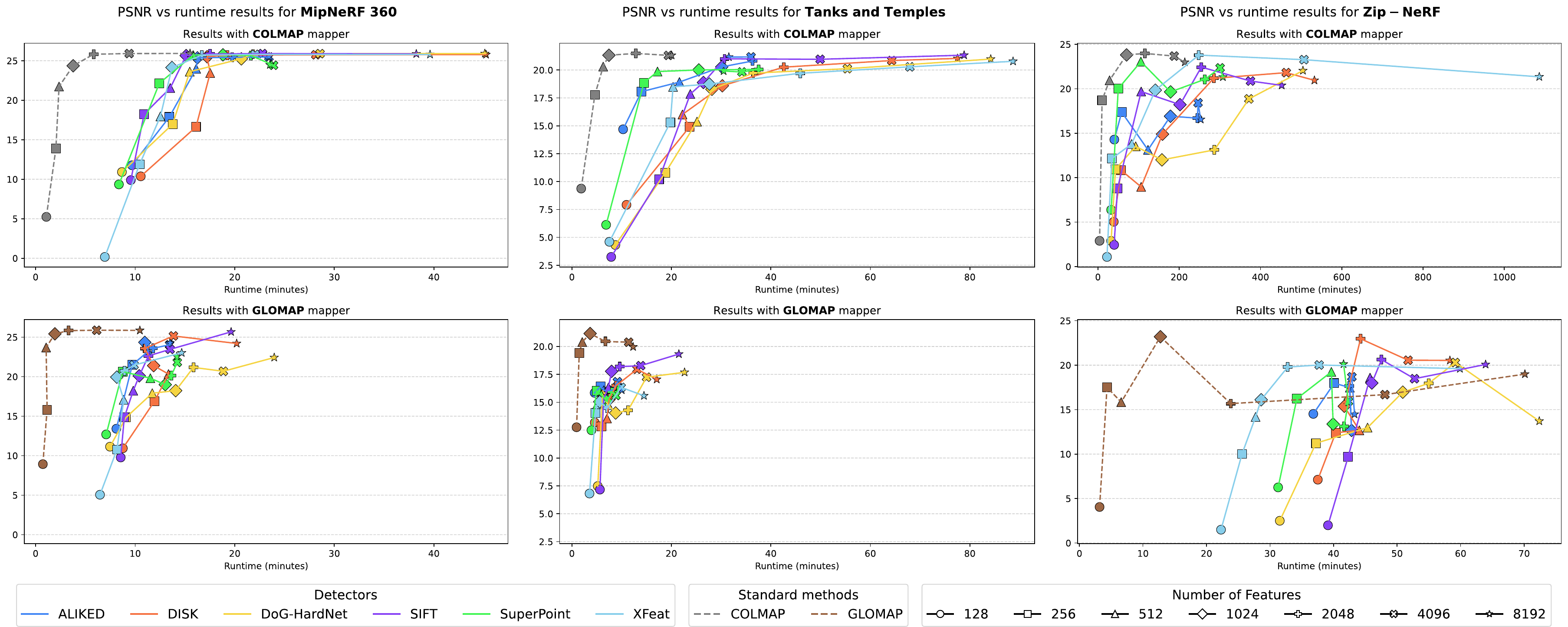}
    \caption{PSNR using NeRF vs. reconstruction runtime results with LightGLue matcher for Mip-NeRF 360, Tanks and Temples, and Zip-NeRF datasets.}
    \label{fig:nerf_3data_psnr_time}
\end{figure*}

Considering that the different number of extracted features affect the final number of 3D points obtained with SfM, and also considering that 3DGS requires the point clouds as input to train a scene, we decided to also evaluate the learned local features for the NVS task using NeRF. In this way, we are only accounting for the quality of the camera poses because this is the only information required by NeRF as input to train a scene. 
For this experiment, we use Nerfacto~\cite{tancik2023nerfstudio} because of its fast training times. We disable camera parameter optimization during training to ensure that we are evaluating the camera poses as we get them after the sparse reconstruction. We evaluate the same 3 datasets that we used on the 3DGS experiment to have a direct comparison.  
We also use the same train and test splits and the same metrics computation strategy (adding dummy values for missing views and failed renderings) that we followed in the 3DGS evaluation. Figure~\ref{fig:nerf_3data_psnr_time} shows the results for the three datasets for both COLMAP and GLOMAP using only the LG matcher. It can be seen that the results obtained with Nerfacto are very similar to those obtained with Mip-Splatting. These results suggest that the renderings for both the learned detectors and the standard methods are mostly dominated by the accuracy of the camera poses because NeRF ignored the 3D point cloud during training.
SSIM and LPIPS metrics, and the results with MNN are reported in the supplementary material.

\subsection{Evaluation of feed-forward SfM}
In this experiment, we want to evaluate multiple feed-forward SfM models. To evaluate the quality of the reconstructions, we use the same camera pose accuracy and NVS evaluation that we performed for the learned local features. In this experiment, we consider Fast3R~\cite{yang2025fast3r}, MASt3R-SfM~\cite{duisterhof2025mast3r}, VGGT~\cite{wang2025vggt}, $\pi^3$~\cite{wang2025pi}, Depth-Anything-3 (DA3)~\cite{lin2025depth}, and VGGT-$\mathrm{\Omega}$~\cite{wang2026vggt}.

Table~\ref{tab:ffn_ETH3D_acc} shows the average camera accuracy results for the feed-forward models using the ETH3D MVS dataset for strict thresholds (\textless1° and \textless5cm) and more permissive ones (\textless5° and \textless10cm). The evaluation and estimation of the camera pose accuracies is the same as the one we followed for the local features. To compare with traditional SfM pipelines, we can also see the results for COLMAP-SuperPoint-LightGlue triplet (referred to in the tables as COLMAP-SP-LG) for 256, 512, and 4096 number of features, and also we include the results for the standard COLMAP and GLOMAP methods using 8192 features. We can see that even though the feed-forward SfM networks are way faster than traditional SfM, the accuracy results are significantly worse. The standard COLMAP and GLOMAP methods provide the best accuracy in this evaluation except for the \textless5° and \textless10cm thresholds, where VGGT-$\mathrm{\Omega}$ provides the best performance.

\begin{table}[t!]
\caption{Average absolute camera pose accuracy across all the scenes from ETH3D MVS dataset for the feed-forward SfM methods and traditional SfM pipelines. We highlight the \colorbox{tabfirst}{best}, \colorbox{tabsecond}{second}, and \colorbox{tabthird}{third} values.}
\centering
\resizebox{0.58\textwidth}{!}{
\begin{tabular}{l|c|c|c}
                    & \textbf{\begin{tabular}[c]{@{}c@{}}\textless{}1° and\\ \textless{}5cm\end{tabular}} & \textbf{\begin{tabular}[c]{@{}c@{}}\textless{}5° and\\ \textless{}10cm\end{tabular}} & \textbf{Time (s)} \\ \hline
Fast3R              & 1.22\%                                                                              & 14.92\%                                                                              & \cellcolor{tabfirst}{1.08}              \\
MASt3R-SfM    & 31.40\%                                                                             & 54.45\%                                                                              & 544.93              \\
VGGT                & 10.80\%                                                                             & 44.32\%                                                                              & \cellcolor{tabthird}{2.12}              \\
$\pi^3$             & 22.38\%                                                                             & 62.36\%                                                                              & 2.71              \\
DA3    & 47.22\%                                                                             & 78.40\%                                                                              & 2.45              \\
VGGT-$\mathrm{\Omega}$                & 55.46\%                                                                             & \cellcolor{tabfirst}{84.63\%}                                                                              & \cellcolor{tabsecond}{1.96}              \\ \hline
COLMAP-SP-LG (256)  & 48.33\%                                                                             & 49.44\%                                                                              & 65.48             \\
COLMAP-SP-LG (512)  & 56.68\%                                                                             & 62.14\%                                                                              & 107.60            \\
COLMAP-SP-LG (4096) & \cellcolor{tabthird}{57.68\%}                                                                             & 73.16\%                                                                              & 198.63            \\ \hline
Standard COLMAP (8192)   & \cellcolor{tabfirst}{77.39\%}                                                                             & \cellcolor{tabthird}{83.07\%}                                                                              & 87.53             \\
Standard GLOMAP (8192)   & \cellcolor{tabsecond}{77.28\%}                                                                             & \cellcolor{tabsecond}{83.41}\%                                                                              & 64.84            
\end{tabular}
}
\label{tab:ffn_ETH3D_acc}
\end{table}

In Table~\ref{tab:ffn_psnr} we have the average PSNR rendering results on Mip-NeRF 360, Tanks and Temples, and Zip-NeRF datasets, using both 3DGS (Mip-Splatting) and NeRF (Nerfacto). PSNR metrics are aggregated across all the scenes in the same way as in the previous experiments with learned features. We also report the average rendering quality results for the same set of traditional SfM configurations as in Table~\ref{tab:ffn_ETH3D_acc}.

Because some feed-forward models require considerable amount of GPU memory and because we have 40GB of VRAM, many scenes failed to be reconstructed with Out-of-Memory (OOM) errors. 
Due to this problem, in addition to PSNR and reconstruction times, we also report the number of scenes successfully reconstructed in the table.
To have a fairer comparison, we also consider VGGT-Long~\cite{deng2025vggt} approach which extends feed-forward reconstruction to larger-scale sequences via chunked inference and loop closure. Following this chunked inference approach, we are able to reconstruct all the scenes from all the datasets without OOM errors and obtain results for VGGT, $\pi^3$, DA3, and VGGT-$\mathrm{\Omega}$. The official VGGT-Long code release supports both VGGT and $\pi^3$, referred to as VGGT-Long and $\pi^3$-Long, respectively. The official DA3 code base also supports this chunked larger-scale inference approach, which is called DA3-Streaming. Finally, based on the VGGT-Long code base, we added support for VGGT-$\mathrm{\Omega}$. We refer to this approach as VGGT-$\mathrm{\Omega}$-Long. In summary, you can find in Table~\ref{tab:ffn_psnr} the results for VGGT, $\pi^3$, DA3, and VGGT-$\mathrm{\Omega}$ and all their chunked versions: VGGT-Long, $\pi^3$-Long, DA3-Streaming, and VGGT-$\mathrm{\Omega}$-Long, respectively.

With this experiment, we can see that MASt3R-SfM does not provide good results compared to the other methods even considering that it optimizes the outputs. If we only consider the feed-forward-based methods, we can usually get the best rendering results with either VGGT-$\mathrm{\Omega}$ or VGGT-$\mathrm{\Omega}$-Long. However, in general, the best PSNR results are obtained with classical SfM. Furthermore, even the rendering results with 256 features with classical SfM are usually better than the feed-forward models, but the reconstruction times are considerably faster with the feed-forward approaches.

\begin{table*}[th!]\caption{Average PSNR results on multiple datasets for feed-forward SfM methods and different configuration for traditional SfM, using Mip-Splatting (3DGS) and Nerfacto (NeRF). Time are in minutes.}
\centering
\resizebox{0.95\textwidth}{!}{
\begin{tabular}{l|cccc|cccc|cccc}
\toprule
\textbf{Method}             & \multicolumn{4}{c|}{\textbf{Mip-NeRF 360}} & \multicolumn{4}{c|}{\textbf{Tanks \& Temples}} & \multicolumn{4}{c}{\textbf{Zip-NeRF}} \\ 
             & 3DGS     & NeRF    & Time    & Succ.   & 3DGS      & NeRF      & Time      & Succ.   & 3DGS   & NeRF   & Time   & Succ.  \\ \hline

Fast3R & 15.06 & 15.12 & \cellcolor{tabfirst}{0.26} & 9/9 & 12.61 & 13.38 & \cellcolor{tabsecond}{0.33} & 20/21 & OOM & OOM & - & 0/4 \\

MASt3R-SfM & 16.93 & 17.14 & 7.16 & 9/9 & 14.24 & 15.54 & 10.64 & 21/21 & 12.87 & 11.66 & 50.36 & 4/4 \\

VGGT & 2.22 & 1.86 & - & 1/9 & 1.78 & 1.96 & - & 2/21 & OOM & OOM & - & 0/4 \\

VGGT-Long & 20.49 & 18.03 & 1.89 & 9/9 & 18.07 & 16.91 & 2.74 & 21/21 & 17.11 & 16.68 & \cellcolor{tabsecond}{15.37} & 4/4 \\

$\pi^3$ & 16.03 & 14.30 & 0.69 & 9/9 & 11.42 & 9.71 & 0.95 & 17/21 & OOM & OOM & - & 0/4 \\

$\pi^3$-Long & 16.71 & 14.34 & 2.95 & 9/9 & 15.23 & 12.83 & 4.65 & 21/21 & 16.22 & 13.85 & \cellcolor{tabthird}{26.03} & 4/4 \\

DA3 & 17.50 & 15.55 & \cellcolor{tabsecond}{0.34} & 8/9 & 11.93 & 12.28 & \cellcolor{tabfirst}{0.30} & 14/21 & OOM & OOM & - & 0/4 \\

DA3-Streaming & 20.35 & 18.10 & 1.23 & 9/9 & 17.79 & 17.76 & 1.63 & 21/21 & 18.39 & 17.97 & \cellcolor{tabfirst}{8.78} & 4/4 \\

VGGT-$\mathrm{\Omega}$ & 21.71 & 18.47 & \cellcolor{tabthird}{0.48} & 9/9 & 14.60 & 14.61 & \cellcolor{tabthird}{0.60} & 16/21 & OOM & OOM & - & 0/4 \\

VGGT-$\mathrm{\Omega}$-Long & 20.94 & 17.94 & 2.47 & 9/9 & 18.66 & 18.12 & 4.05 & 21/21 & 18.79 & 18.74 & 20.03 & 4/4 \\ \hline

COLMAP-SP-LG (256) & 24.07 & 22.15 & 12.43 & 9/9 & 20.37 & 18.86 & 14.49 & 21/21 & \cellcolor{tabthird}{20.66} & 20.01 & 50.11 & 4/4 \\

COLMAP-SP-LG (512) & \cellcolor{tabthird}{27.24} & \cellcolor{tabthird}{25.71} & 15.82 & 9/9 & \cellcolor{tabthird}{21.42} & \cellcolor{tabthird}{19.87} & 17.21 & 21/21 & \cellcolor{tabfirst}{24.02} & \cellcolor{tabfirst}{23.05} & 105.72 & 4/4 \\

COLMAP-SP-LG (4096) & 26.52 & 24.43 & 23.89 & 9/9 & 21.37 & 19.82 & 34.18 & 21/21 & 17.95 & \cellcolor{tabthird}{22.35} & 300.59 & 4/4 \\

Standard COLMAP (8192) & \cellcolor{tabsecond}{27.67} & \cellcolor{tabfirst}{25.93} & 15.51 & 9/9 & \cellcolor{tabfirst}{22.91} & \cellcolor{tabfirst}{21.32} & 20.05 & 21/21 & \cellcolor{tabsecond}{24.00} & \cellcolor{tabsecond}{22.98} & 213.76 & 4/4 \\

Standard GLOMAP (8192) & \cellcolor{tabfirst}{27.72} & \cellcolor{tabsecond}{25.86} & 10.47 & 9/9 & \cellcolor{tabsecond}{21.57} & \cellcolor{tabsecond}{19.96} & 12.31 & 21/21 & 19.26 & 18.99 & 70.12 & 4/4 \\ \bottomrule
\end{tabular}
}
\label{tab:ffn_psnr}
\end{table*}

Our comparisons clearly show that classical, feature-based SfM approaches are still the state-of-the-art in terms of camera pose accuracy. 
Hence, they are very much relevant in practice. 
Thus, accelerating these methods is still a relevant problem. 
In this work, we show that a simple way of accelerating both COLMAP and GLOMAP is to simply use fewer features.

\subsubsection{Refinement of the feed-forward reconstructions:} considering the bad rendering results provided by the feed-forward SfM models, we explored the idea of refining the provided camera poses using BA. For that, we first use the camera poses to establish image pairs based on proximity. We first do feature extraction using 8192 features. Rather than using exhaustive matching, which is costly, we match each image to its closet 50 views, as long as the viewing ray angle between then is lower than 60\textdegree. For that we use the camera poses provided by the feed-forward model. With these established image pairs, we do triangulation with the model camera poses and refine this reconstruction with BA.

Table~\ref{tab:ffn_refine_Mip-NeRF360} shows the PSNR results with 3DGS and reconstruction times per scene for the Mip-NeRF 360 dataset for the proposed refinement approach using VGGT-$\mathrm{\Omega}$. We compare with standard COLMAP reconstruction for different number of extracted features. As expected, we can see that VGGT-$\mathrm{\Omega}$ provides the fastest reconstruction times; however, the rendering quality is very poor, as shown before. We also show that the refined VGGT-$\mathrm{\Omega}$ model provides the best rendering quality for almost all the scenes without an important compromise in runtime.

\begin{table*}
\caption{Rendering and reconstruction times for the refinement of VGGT-$\mathrm{\Omega}$ camera poses using 8192 features for the Mip-NeRF 360 dataset. We also show PSNR and times for standard COLMAP reconstruction using different number of features. Times are in minutes.}
\centering
\resizebox{0.95\textwidth}{!}{
\begin{tabular}{l|cc|cc|cc|cc|cc|cc|cc}
\toprule
& \multicolumn{10}{c|}{\textbf{Standard COLMAP reconstruction}} & \multicolumn{4}{c}{\textbf{Feed-forward recons}} \\
 & \multicolumn{2}{c|}{512} & \multicolumn{2}{c|}{1024} & \multicolumn{2}{c|}{2048} & \multicolumn{2}{c|}{4096} & \multicolumn{2}{c|}{8192} & \multicolumn{2}{c|}{VGGT-$\mathrm{\Omega}$} & \multicolumn{2}{c}{Refined VGGT-$\mathrm{\Omega}$} \\
\textbf{Scene} & PSNR & Time & PSNR & Time & PSNR & Time & PSNR & Time & PSNR & Time & PSNR & Time & PSNR & Time \\
\midrule
bicycle & 24.02 & \cellcolor{tabsecond}1.13 & 24.73 & \cellcolor{tabthird}1.39 & 25.37 & 2.31 & \cellcolor{tabthird}25.59 & 3.95 & \cellcolor{tabsecond}25.62 & 9.33 & 19.30 & \cellcolor{tabfirst}0.37 & \cellcolor{tabfirst}25.90 & 3.09 \\
bonsai & 31.16 & \cellcolor{tabsecond}1.80 & 31.59 & \cellcolor{tabthird}3.32 & 24.96 & 11.51 & \cellcolor{tabsecond}32.42 & 12.24 & \cellcolor{tabthird}32.40 & 24.91 & 24.94 & \cellcolor{tabfirst}0.77 & \cellcolor{tabfirst}32.93 & 11.85 \\
counter & 28.70 & \cellcolor{tabsecond}1.16 & 28.94 & \cellcolor{tabthird}1.55 & \cellcolor{tabthird}29.09 & 3.77 & 28.89 & 7.32 & \cellcolor{tabsecond}29.16 & 15.21 & 23.92 & \cellcolor{tabfirst}0.53 & \cellcolor{tabfirst}29.21 & 6.44 \\
flowers & 1.12 & \cellcolor{tabsecond}0.61 & 10.30 & \cellcolor{tabthird}1.03 & 21.16 & 1.84 & \cellcolor{tabthird}21.27 & 2.69 & \cellcolor{tabsecond}21.68 & 7.85 & 16.52 & \cellcolor{tabfirst}0.30 & \cellcolor{tabfirst}22.25 & 1.77 \\
garden & 27.02 & \cellcolor{tabsecond}1.50 & 27.36 & \cellcolor{tabthird}2.11 & 27.61 & 3.37 & \cellcolor{tabthird}27.71 & 6.85 & \cellcolor{tabsecond}27.71 & 10.01 & 22.34 & \cellcolor{tabfirst}0.36 & \cellcolor{tabfirst}28.21 & 7.10 \\
kitchen & 31.17 & \cellcolor{tabsecond}1.99 & \cellcolor{tabthird}31.56 & \cellcolor{tabthird}3.58 & \cellcolor{tabsecond}31.56 & 6.61 & 31.36 & 11.93 & 31.37 & 26.78 & 24.05 & \cellcolor{tabfirst}0.71 & \cellcolor{tabfirst}31.70 & 11.87 \\
room & 30.43 & \cellcolor{tabsecond}2.10 & 31.42 & \cellcolor{tabthird}3.43 & 31.40 & 6.09 & \cellcolor{tabfirst}31.65 & 11.45 & \cellcolor{tabthird}31.48 & 21.59 & 25.35 & \cellcolor{tabfirst}0.86 & \cellcolor{tabsecond}31.58 & 11.53 \\
stump & - & 0.42 & 16.02 & \cellcolor{tabthird}0.75 & 26.60 & 0.98 & \cellcolor{tabsecond}26.85 & 1.75 & \cellcolor{tabthird}26.83 & 4.63 & 19.80 & \cellcolor{tabfirst}0.17 & \cellcolor{tabfirst}27.12 & 1.00 \\
treehill & 3.14 & \cellcolor{tabsecond}0.40 & 21.82 & \cellcolor{tabthird}0.82 & 22.15 & 1.27 & \cellcolor{tabthird}22.33 & 2.41 & \cellcolor{tabsecond}22.38 & 6.18 & 19.15 & \cellcolor{tabfirst}0.21 & \cellcolor{tabfirst}24.08 & 1.92 \\
\bottomrule
\end{tabular}
}
\label{tab:ffn_refine_Mip-NeRF360}
\end{table*}

\begin{table*}
\caption{Rendering and time results on FFN refinements using 8192 feats and Tanks \& Temples dataset. Times are in minutes.}
\label{tab:ffn_refine_Tanks_Temples}
\centering
\resizebox{\textwidth}{!}{
\begin{tabular}{l|cc|cc|cc|cc|cc|cc|cc}
\toprule
& \multicolumn{10}{c|}{\textbf{Standard COLMAP reconstruction}} & \multicolumn{4}{c}{\textbf{Feed-forward recons}} \\
 & \multicolumn{2}{c|}{512} & \multicolumn{2}{c|}{1024} & \multicolumn{2}{c|}{2048} & \multicolumn{2}{c|}{4096} & \multicolumn{2}{c|}{8192} & \multicolumn{2}{c|}{$\pi^3$-Long} & \multicolumn{2}{c}{Refined $\pi^3$-Long} \\
\textbf{Scene} & PSNR & Time & PSNR & Time & PSNR & Time & PSNR & Time & PSNR & Time & PSNR & Time & PSNR & Time \\
\midrule
auditorium & 23.03 & \cellcolor{tabsecond}2.67 & \cellcolor{tabthird}23.28 & \cellcolor{tabfirst}2.18 & \cellcolor{tabfirst}23.32 & 4.64 & 23.27 & 8.98 & \cellcolor{tabsecond}23.31 & 6.06 & 16.53 & \cellcolor{tabthird}3.92 & 22.64 & 7.98 \\
ballroom & \cellcolor{tabsecond}21.88 & \cellcolor{tabfirst}2.39 & 19.69 & \cellcolor{tabthird}5.56 & \cellcolor{tabthird}19.78 & 9.46 & \cellcolor{tabfirst}23.11 & 18.60 & 17.13 & 18.28 & 13.44 & \cellcolor{tabsecond}4.41 & 14.34 & 21.23 \\
barn & 27.44 & \cellcolor{tabfirst}2.58 & 27.66 & \cellcolor{tabsecond}4.44 & 27.86 & 8.91 & \cellcolor{tabfirst}27.96 & 13.51 & \cellcolor{tabsecond}27.95 & 13.56 & 18.65 & \cellcolor{tabthird}5.18 & \cellcolor{tabthird}27.87 & 11.31 \\
caterpillar & 23.11 & \cellcolor{tabfirst}2.47 & 23.04 & \cellcolor{tabsecond}3.78 & 23.31 & 7.60 & \cellcolor{tabthird}23.38 & 12.76 & \cellcolor{tabsecond}23.39 & 16.03 & 16.09 & \cellcolor{tabthird}5.21 & \cellcolor{tabfirst}23.44 & 22.12 \\
church & 20.75 & \cellcolor{tabsecond}9.61 & 20.70 & \cellcolor{tabthird}16.47 & \cellcolor{tabthird}21.57 & 29.77 & \cellcolor{tabsecond}21.57 & 38.93 & \cellcolor{tabfirst}21.85 & 68.89 & 14.78 & \cellcolor{tabfirst}7.15 & 17.99 & 29.02 \\
courthouse & 21.06 & \cellcolor{tabsecond}17.72 & \cellcolor{tabthird}21.15 & \cellcolor{tabthird}28.72 & \cellcolor{tabsecond}21.41 & 64.73 & 21.02 & 86.76 & 20.81 & 95.86 & 16.80 & \cellcolor{tabfirst}14.06 & \cellcolor{tabfirst}22.78 & 162.71 \\
courtroom & 22.07 & \cellcolor{tabfirst}1.92 & \cellcolor{tabthird}22.10 & \cellcolor{tabsecond}3.03 & \cellcolor{tabsecond}22.15 & 4.82 & 20.85 & 10.13 & \cellcolor{tabfirst}22.30 & 9.81 & 14.31 & \cellcolor{tabthird}4.19 & 19.82 & 11.62 \\
family & 24.10 & \cellcolor{tabfirst}0.46 & \cellcolor{tabsecond}24.39 & \cellcolor{tabsecond}0.73 & \cellcolor{tabthird}24.38 & \cellcolor{tabthird}1.36 & 24.33 & 3.27 & 24.34 & 3.05 & 15.92 & 2.12 & \cellcolor{tabfirst}24.87 & 5.07 \\
francis & 26.74 & \cellcolor{tabfirst}1.40 & 25.87 & 4.71 & \cellcolor{tabthird}27.12 & \cellcolor{tabthird}3.98 & 26.99 & 6.92 & \cellcolor{tabsecond}27.15 & 6.98 & 17.75 & \cellcolor{tabsecond}3.55 & \cellcolor{tabfirst}27.82 & 13.21 \\
horse & 22.00 & \cellcolor{tabfirst}0.35 & \cellcolor{tabsecond}23.65 & \cellcolor{tabsecond}0.54 & \cellcolor{tabthird}23.56 & \cellcolor{tabthird}1.05 & 23.49 & 1.46 & 23.36 & 1.39 & 13.96 & 1.88 & \cellcolor{tabfirst}24.11 & 3.99 \\
ignatius & 20.30 & \cellcolor{tabfirst}1.13 & \cellcolor{tabsecond}21.16 & \cellcolor{tabsecond}2.09 & \cellcolor{tabthird}21.05 & \cellcolor{tabthird}3.24 & 19.27 & 8.69 & 20.29 & 13.87 & 14.30 & 3.44 & \cellcolor{tabfirst}22.08 & 12.78 \\
lighthouse & 19.10 & \cellcolor{tabfirst}1.03 & 21.31 & \cellcolor{tabsecond}2.54 & \cellcolor{tabthird}21.49 & 4.12 & \cellcolor{tabfirst}21.75 & 6.42 & 21.35 & 13.18 & 13.94 & \cellcolor{tabthird}3.43 & \cellcolor{tabsecond}21.68 & 19.87 \\
m60 & 25.93 & \cellcolor{tabfirst}1.03 & 27.61 & \cellcolor{tabsecond}1.99 & 27.96 & \cellcolor{tabthird}4.08 & \cellcolor{tabsecond}28.10 & 7.21 & \cellcolor{tabthird}28.08 & 6.21 & 15.01 & 4.13 & \cellcolor{tabfirst}28.25 & 11.65 \\
meetingroom & 23.76 & \cellcolor{tabfirst}1.25 & 24.91 & \cellcolor{tabsecond}2.12 & \cellcolor{tabfirst}25.12 & \cellcolor{tabthird}3.54 & \cellcolor{tabthird}25.07 & 3.77 & \cellcolor{tabsecond}25.09 & 3.97 & 14.94 & 4.69 & 24.85 & 10.30 \\
museum & 19.50 & \cellcolor{tabfirst}1.15 & \cellcolor{tabsecond}20.19 & \cellcolor{tabsecond}3.70 & \cellcolor{tabfirst}20.20 & \cellcolor{tabthird}3.81 & 20.02 & 6.79 & \cellcolor{tabthird}20.17 & 7.34 & 12.89 & 4.68 & 18.74 & 14.26 \\
palace & 13.40 & \cellcolor{tabfirst}1.68 & \cellcolor{tabthird}17.37 & \cellcolor{tabsecond}4.03 & \cellcolor{tabfirst}17.98 & 8.34 & \cellcolor{tabsecond}17.64 & 14.73 & 17.33 & 13.24 & 13.64 & \cellcolor{tabthird}6.06 & 13.94 & 22.40 \\
panther & 28.12 & \cellcolor{tabfirst}1.28 & 27.82 & \cellcolor{tabsecond}2.70 & 28.45 & 4.94 & \cellcolor{tabthird}28.59 & 10.79 & \cellcolor{tabsecond}28.61 & 10.65 & 15.91 & \cellcolor{tabthird}4.89 & \cellcolor{tabfirst}28.92 & 17.37 \\
playground & 25.09 & \cellcolor{tabfirst}1.13 & 25.37 & \cellcolor{tabsecond}1.78 & 25.45 & 4.51 & \cellcolor{tabsecond}25.89 & 5.34 & \cellcolor{tabthird}25.53 & 7.69 & 18.08 & \cellcolor{tabthird}4.25 & \cellcolor{tabfirst}25.99 & 10.35 \\
temple & 12.10 & \cellcolor{tabfirst}1.44 & \cellcolor{tabthird}16.02 & \cellcolor{tabsecond}1.55 & \cellcolor{tabfirst}19.01 & \cellcolor{tabthird}2.48 & 15.43 & 4.19 & 15.39 & 3.88 & 13.55 & 3.30 & \cellcolor{tabsecond}18.92 & 9.97 \\
train & 20.37 & \cellcolor{tabfirst}1.16 & \cellcolor{tabthird}21.32 & \cellcolor{tabsecond}1.96 & \cellcolor{tabsecond}21.37 & \cellcolor{tabthird}3.64 & \cellcolor{tabfirst}21.47 & 8.03 & 21.32 & 7.78 & 14.05 & 4.01 & 20.45 & 12.60 \\
truck & 12.35 & \cellcolor{tabsecond}1.65 & 24.14 & \cellcolor{tabfirst}1.39 & 24.21 & \cellcolor{tabthird}2.51 & \cellcolor{tabthird}24.31 & 4.69 & \cellcolor{tabsecond}24.35 & 5.50 & 15.20 & 3.03 & \cellcolor{tabfirst}24.44 & 8.81 \\
\bottomrule
\end{tabular}
}
\end{table*}

Table~\ref{tab:ffn_refine_Tanks_Temples} compares the effectiveness and efficiency of the original $\pi^3$-Long feed-forward approach and its refined version on the Tanks \& Temples dataset. 
As can be seen for the MipNeRF360 dataset, the refined versions of the $\pi^3$-Long poses lead to the best novel view synthesis results. 
At the same time, the reconstruction process remains efficient, with only a few minutes typically required per scene.

%%%%%%%%%%%%%%%%%%%%%%%%%%%%%%%%%%%%%%%%%%%%%%%%%%%%%%%%%%%%%%%%%%%%%%%
%%%%%%%%%%%%%%%%%%%%%%%%%%%%%%%%%%%%%%%%%%%%%%%%%%%%%%%%%%%%%%%%%%%%%%%
%%%%%%%%%%%%%%%%%%%%%%%%%%%%%%%%%%%%%%%%%%%%%%%%%%%%%%%%%%%%%%%%%%%%%%%

\section{Conclusions}
\label{sec:conclusions}
In this paper, we have explored the efficiency-effectiveness trade-off of Structure-from-Motion systems in the context of novel view synthesis. 
In particular, we are interested in computing camera pose estimates for a given set of images as efficiently as possible while ensuring that they are accurate enough for novel view synthesis. 
Using the observation that modern feed-forward approaches to SfM are highly efficient, but do not lead to very accurate pose estimates, we have investigated making classical SfM systems faster simply by using fewer features. 
Detailed experiments show that using modern learned local features and learned feature matchers allow us to significantly accelerate the SfM process at moderate to little costs in pose accuracy. 
Interestingly, as little as 512 features per image can be sufficient for accurate pose estimation. 
We further showed that local feature matching can be used to make feed-forward approaches more effective: 
To this end, we used classical SfM techniques to refine the (rather inaccurate) pose estimates obtained by the feed-forward methods. 
Naturally, this approach comes at the cost of run-time efficiency. 
Still, it often offers a very good trade-off between efficiency and effectiveness. 
Overall, our results show that classical SfM techniques are still relevant, even in the age of transformer-based pose and structure regressors. 
To foster research on the trade-off between efficiency and effectiveness, we will make our benchmark, including all evaluation code and the code for the two strategies, publicly available.

\section*{Acknowledgements}
This work was supported by the Czech Science Foundation (GAČR) EXPRO Grant No. 23-07973X and by the Ministry of Education, Youth and Sports of the Czech Republic through the e-INFRA CZ (ID:90254).

% ---- Bibliography ----
%
% BibTeX users should specify bibliography style 'splncs04'.
% References will then be sorted and formatted in the correct style.
%
\bibliographystyle{splncs04}
\bibliography{ref}

\begin{appendix}

\section{Sparse reconstruction timings per stage}
\label{sec:sparse_times}
For each learned feature extractor, we report the average runtimes for the different stages of the sparse reconstruction pipeline: feature extraction, feature matching, geometric verification, and 3D mapping. The reported timings for the different number of extracted features are for the LightGlue (LG)~\cite{lindenberger2023lightglue} matcher and the COLMAP~\cite{schonberger2016structure} and GLOMAP~\cite{pan2024global} mappers. 

\begin{figure*}%[h]
    \centering
    \includegraphics[width=\textwidth]{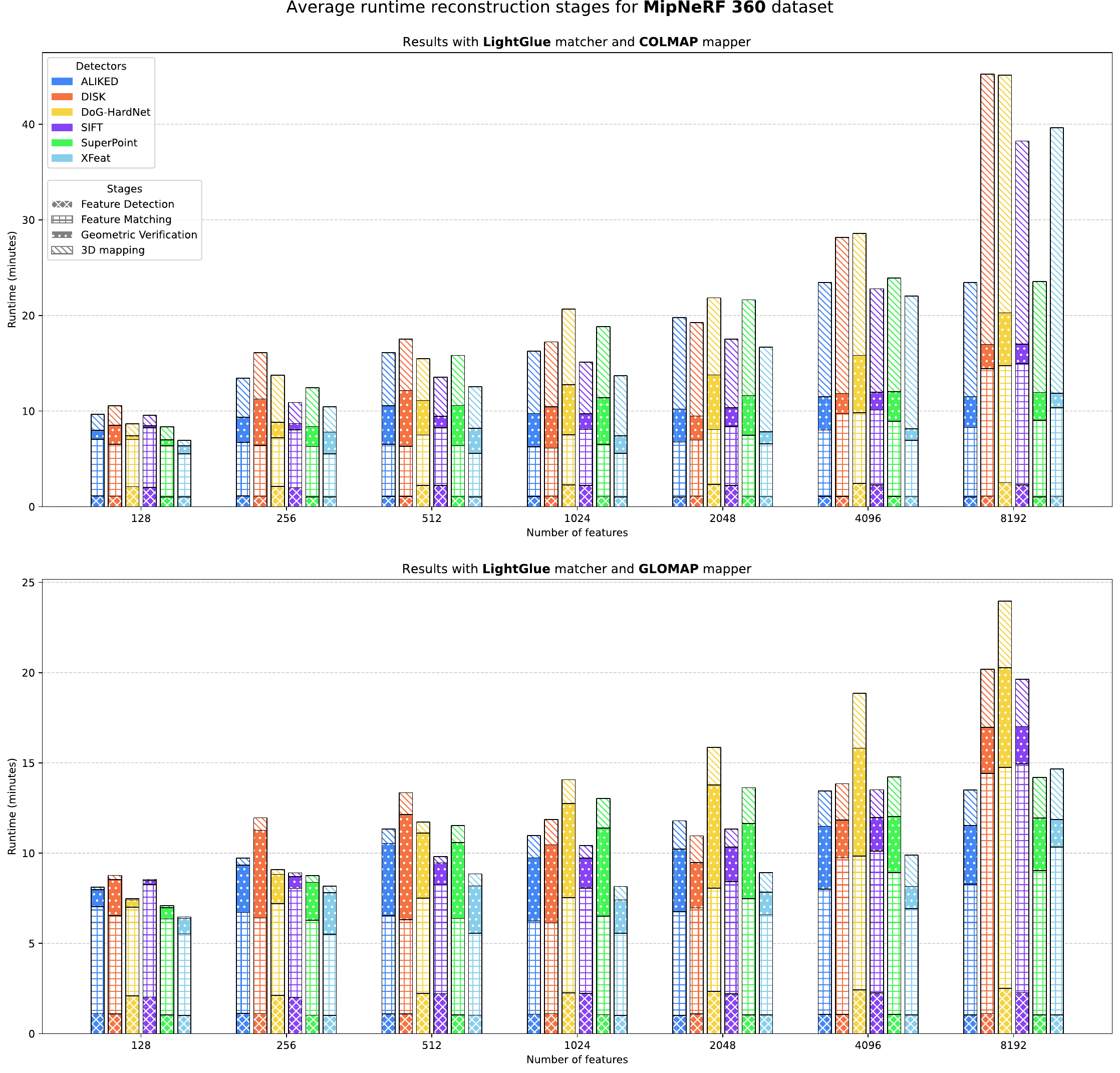}
    \caption{Average runtimes for each stage of the Structure-from-Motion pipeline for the Mip-NeRF 360 dataset.}
    \label{fig:sparse_times-LG-mip360}
\end{figure*}

We can see the average timings for the Mip-NeRF 360 dataset~\cite{barron2022mip} in Figure~\ref{fig:sparse_times-LG-mip360}. 
We note that the feature extraction stage takes almost the same amount of time no matter how many features we ask to detect. 
For a smaller number of keypoints, we can see that feature matching is the stage that takes longer compared to the others. 
When we use COLMAP, we see that with fewer features the feature detection, feature matching, and geometric verification stages usually take longer than the 3D mapping. 
However, with 1024 features or more, the 3D mapping stage slightly becomes the bottleneck of the SfM pipeline, even in some cases taking more time than the other stages combined (as we can see with 8192 features). 
On the other hand, with GLOMAP we see that the mapping stage is always faster than the extraction, matching, and geometric verification stages combined, no matter how many features we use. 
3D mapping times increase with the number of features, but the bottleneck is always the feature matching and geometric verification stages. Finally, as an overall observation, it is interesting to see that both DISK and DoG-HardNet are the learned features that usually lead to the longest average 3D reconstruction times for almost all the different number of extracted features.

\begin{figure*}%[h]
    \centering
    \includegraphics[width=\textwidth]{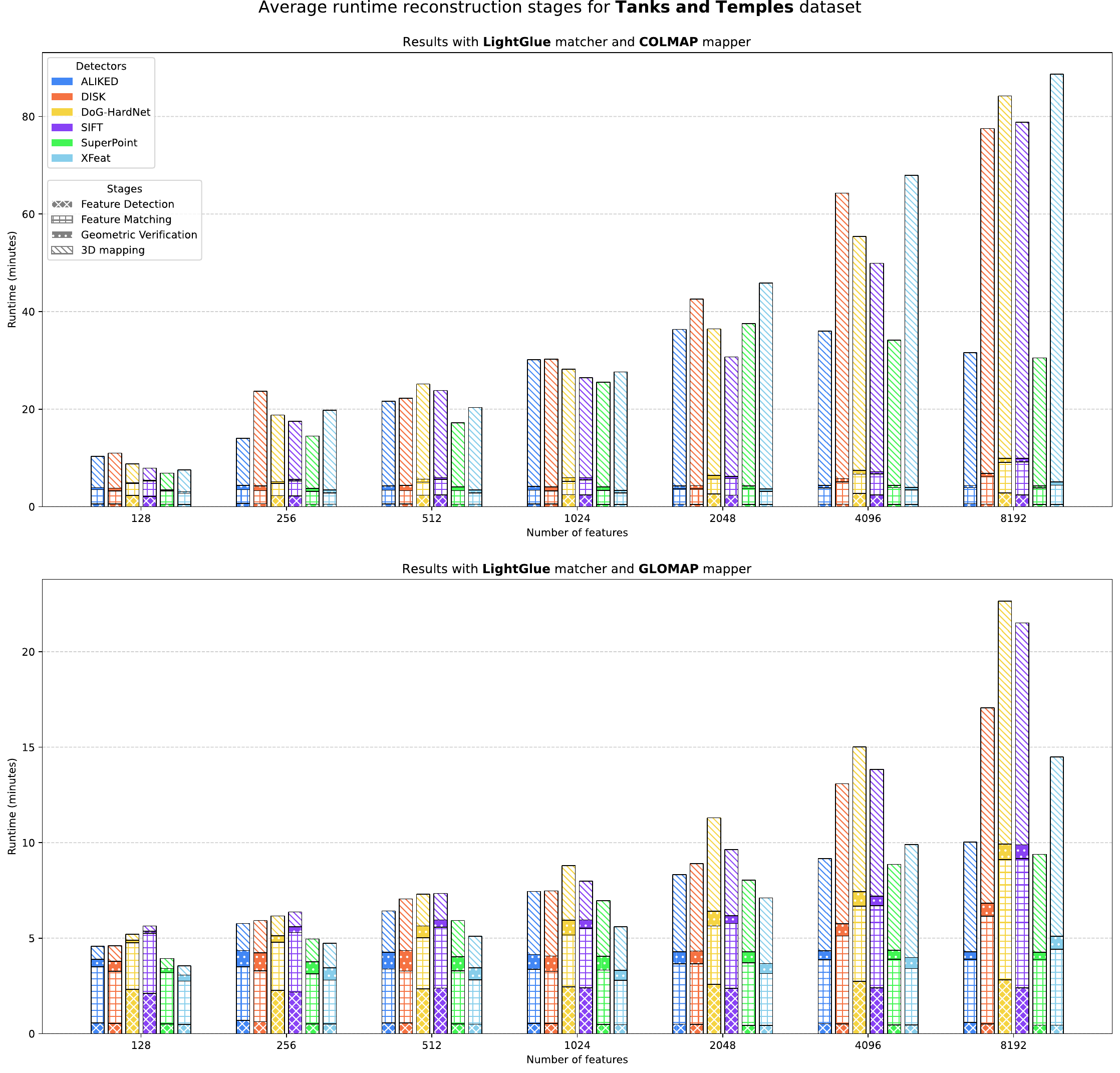}
    \caption{Average runtimes for each stage of the Structure-from-Motion pipeline for the Tanks and Temples dataset.}
    \label{fig:sparse_times-LG-tnt}
\end{figure*}

The average timings per reconstruction stage for the Tanks and Temples dataset are shown in Figure~\ref{fig:sparse_times-LG-tnt}. 
Unlike the Mip-NeRF 360 dataset, with Tanks and Temples we can see that for COLMAP, the 3D mapping is clearly the bottleneck for all the number of extracted features. 
In many cases, the incremental mapping stage represents more than 70\% of the total reconstruction time, and even for some feature extractors with 2048, 4096, and 8192 features, it represents more than 90\% of the total. 
With GLOMAP, on the other hand, the average mapping runtime increases with the number of features, and for 1024 and more it becomes the main bottleneck of the reconstruction process. 
For fewer features, feature matching is the stage that takes longer to run and the 3D mapping is one of the fastest stages of the whole pipeline.

\begin{figure*}%[h]
    \centering
    \includegraphics[width=\textwidth]{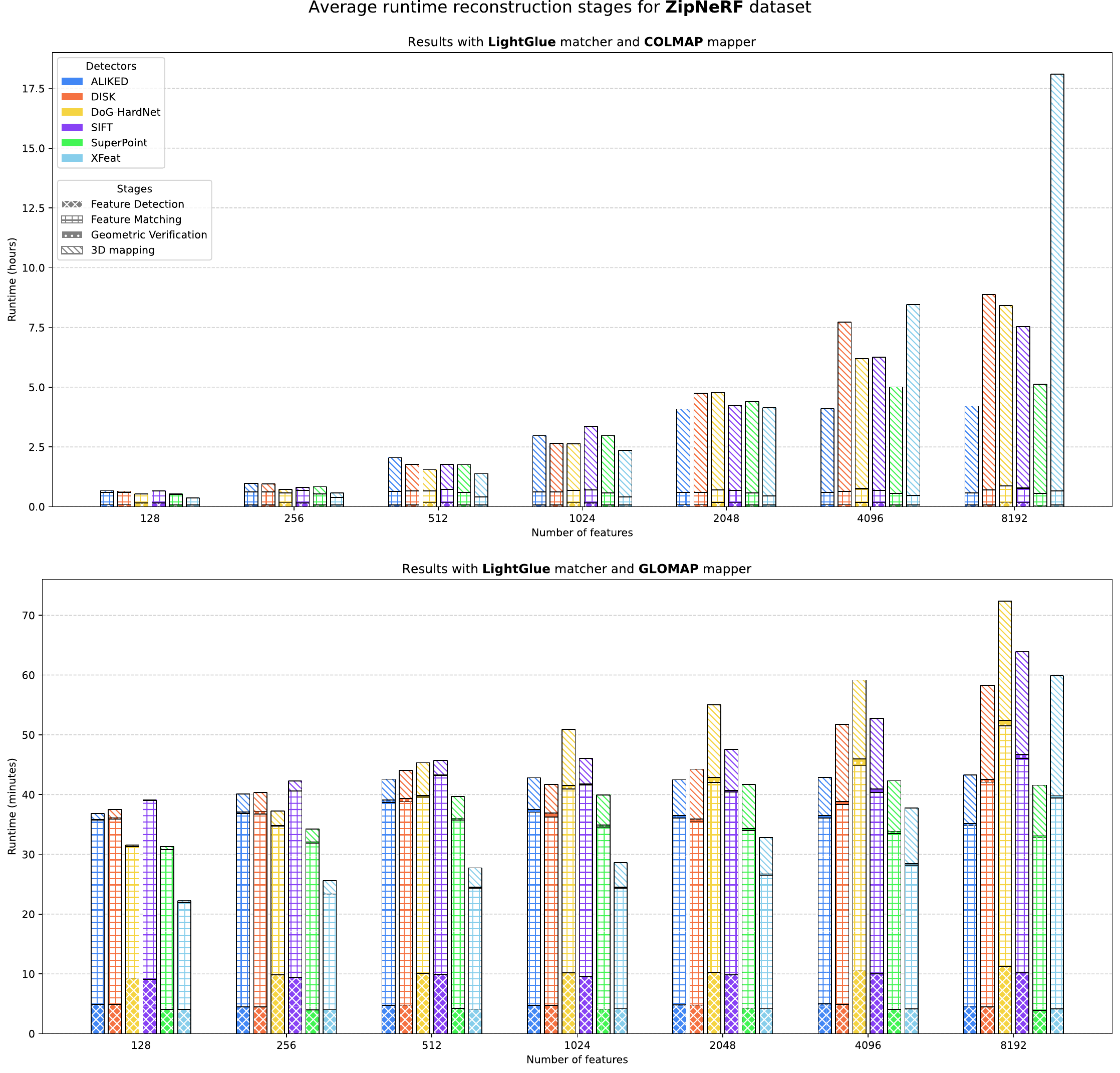}
    \caption{Average runtimes for each stage of the Structure-from-Motion pipeline for the Zip-NeRF dataset.}
    \label{fig:sparse_times-LG-zip}
\end{figure*}

Figure~\ref{fig:sparse_times-LG-zip} shows the average times of the SfM pipeline for the Zip-NeRF dataset. 
For COLMAP, we can see that the timings are dominated by the 3D mapping stage which represents more than 70\% and in some cases more than 90\% of the total reconstruction time. 
Except for 126 and 256 keypoints where feature matching is the stage that takes longer. 
We can also see that for 4096 and 8192 features, ALIKED and SuperPoint provide the best average total runtimes. 
This is because, as discussed in the main paper, for these two feature extractors, the number of extracted features has an upper bound which is around 4000. 
This means that increasing the requested number of features does not result in additional keypoints being extracted. 
That is also why the obtained runtimes for 4096 and 8192 are very similar. 
Note that this also happens with the previous datasets. 
On the other hand, for GLOMAP, feature matching is always the bottleneck for all the number of extracted features. 
The 3D mapping timings are usually very short compared to the extraction, matching, and geometric verification stages combined. 
This behavior is similar to the one observed with the Mip-NeRF 360 dataset.

\section{Additional novel view synthesis results with 3DGS}
\label{sec:add_nvs}
\begin{figure*}
    \centering
    \includegraphics[width=1\textwidth]{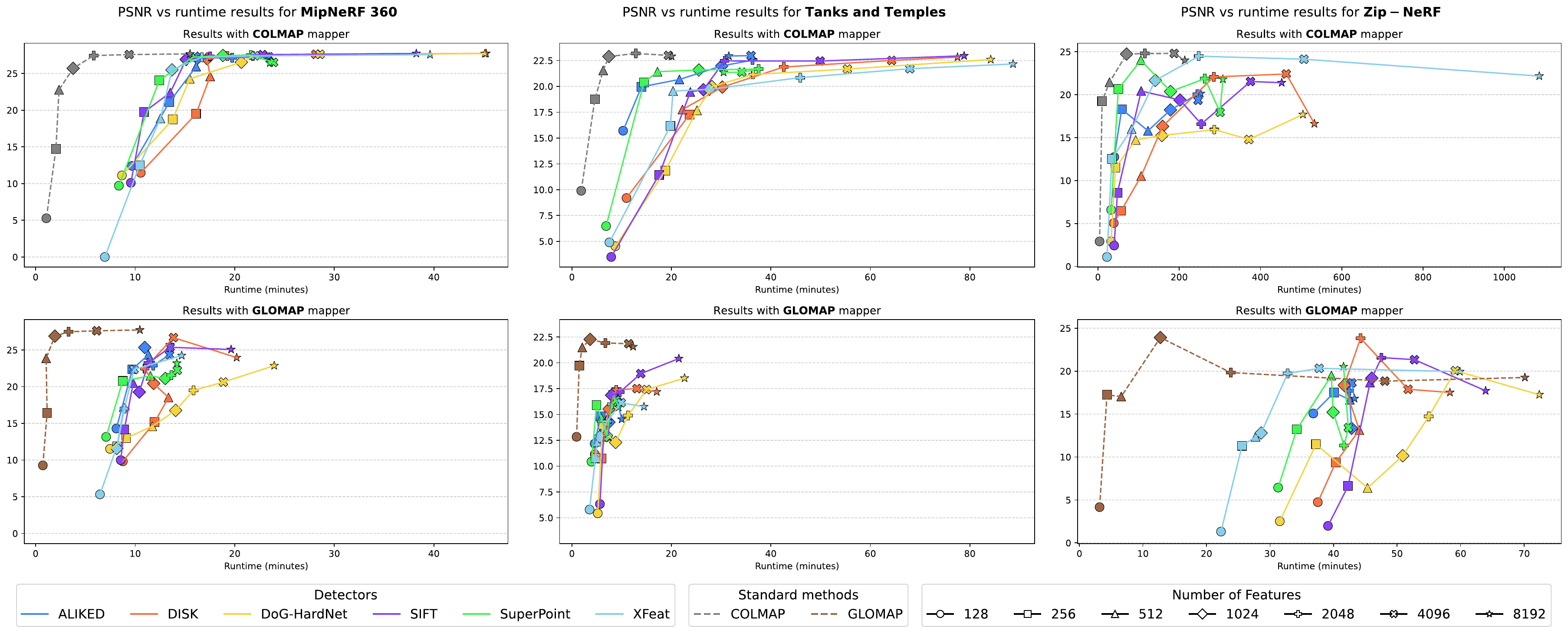}
    \caption{PSNR vs. reconstruction runtime results using the MNN matcher for Mip-NeRF 360, Tanks and Temples, and Zip-NeRF datasets.}
    \label{fig:3data_MNN_psnr_time}
\end{figure*}

In this section, we present the PSNR and reconstruction runtime plots for Mip-NeRF 360, Tank and Temples, and Zip-NeRF datasets using the Mutual Nearest Neighbors (MNN) matcher in Figure~\ref{fig:3data_MNN_psnr_time}. 
These results correspond to the rendering experiments in Section~\ref{subsec:3dgs} from the main paper with Mip-Splatting~\cite{yu2024mip}.

Overall, we can see that the trend of the plots is similar to the ones obtained with LG matcher reported in the main paper. 
For Mip-NeRF 360 dataset, we can see that both SuperPoint and DoG-HardNet tend to be faster than the other feature detectors. 
Note that SuperPoint achieves competitive PSNR results with only 1024 features with COLMAP and 2048 or more features with GLOMAP, while maintaining a short average runtime compared to the other extractors. 
For the Tanks and Temples dataset, with COLMAP, we can see the learned features having a marginal improvement after 2048 or 4096 extracted features. 
This behavior is similar to the one observed on the LG-COLMAP plot from the main paper. 
With GLOMAP, the performance of the learned extractors improves compared to the results with LG-GLOMAP, although the standard GLOMAP is still slightly better and significantly faster. 
Finally for Zip-NeRF dataset, with the COLMAP mapper DoG-HardNet provides the best average PSNR results at 8192 features. We can also see that SuperPoint has better average runtimes for 4096 and 8192 features than the standard COLMAP, while DoG-HardNet and ALIKED provide similar timings for the same number of keypoints. 
On the other hand, with GLOMAP we can see a notable degradation with PSNR metrics significantly below the maximum achieved by the standard GLOMAP. 
Although SuperPoint and DoG-HardNet tend to be better than the other detectors.

\begin{figure*}
    \centering
    \includegraphics[width=\textwidth]{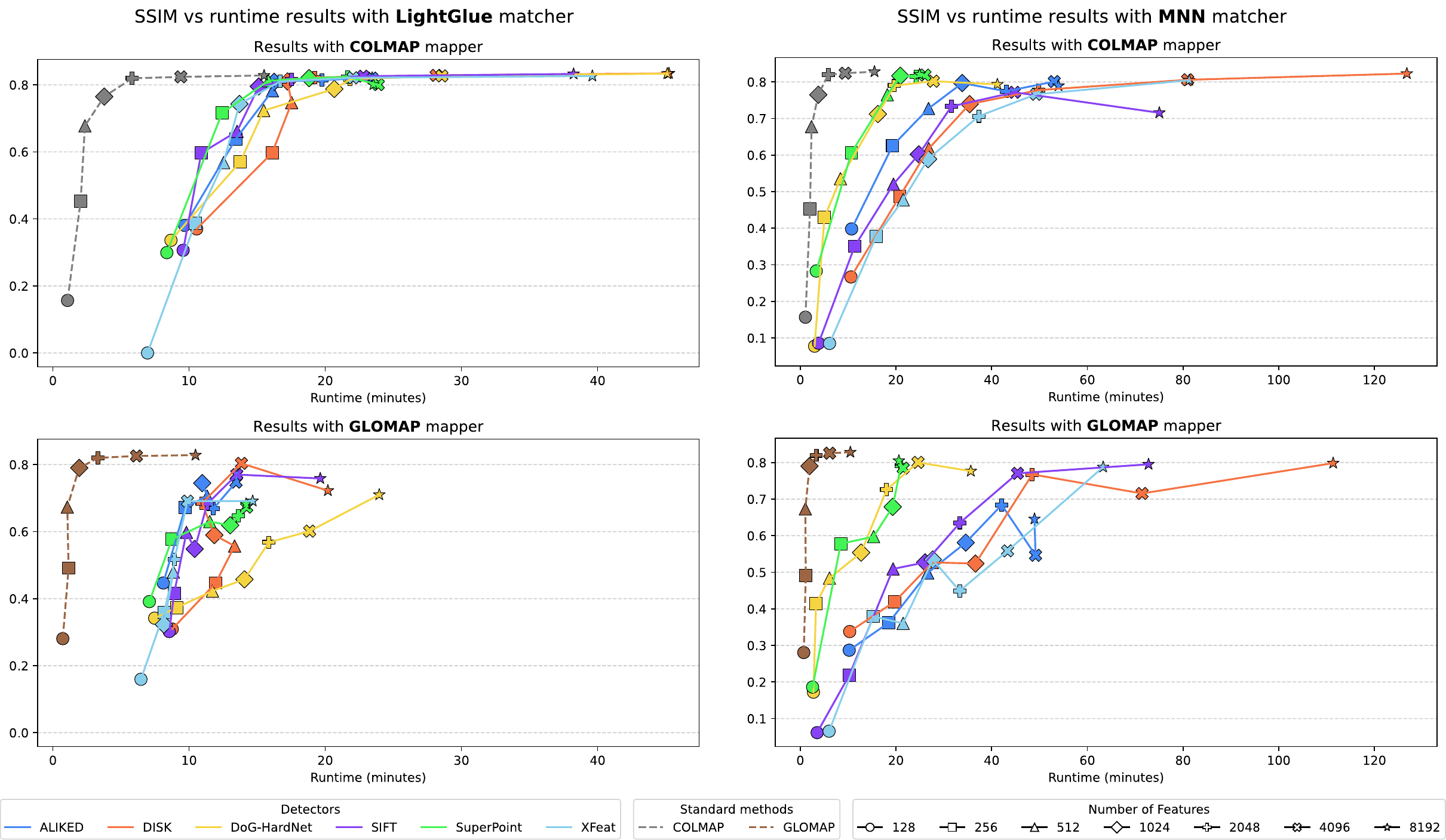}
    \caption{SSIM and runtime results for the Mip-NeRF 360 dataset.}
    \label{fig:mip360_ssim_time}
\end{figure*}

\begin{figure*}
    \centering
    \includegraphics[width=\textwidth]{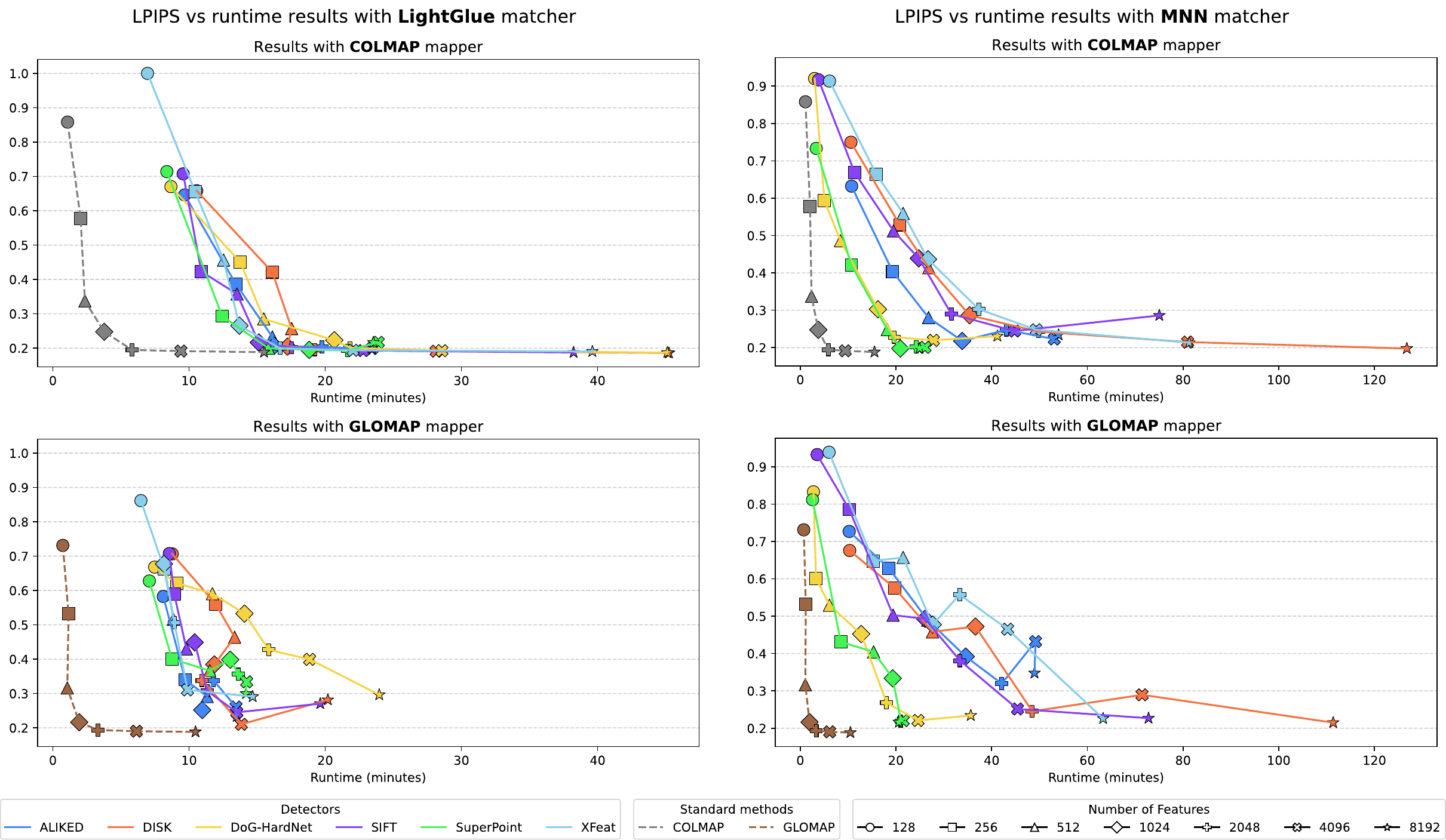}
    \caption{LPIPS and runtime results for the Mip-NeRF 360 dataset.}
    \label{fig:mip360_lpips_time}
\end{figure*}

In this section, we also report the SSIM and LPIPS results corresponding to the rendering experiments with Mip-Splatting from Section~\ref{subsec:3dgs} from the main paper. Figure~\ref{fig:mip360_ssim_time} presents the average SSIM vs average runtimes for the Mip-NeRF 360 dataset. The behavior of the plot is very similar to the one obtained with the PSNR metric. Also, Figure~\ref{fig:mip360_lpips_time} shows the average LPIPS results as a function of the average reconstruction times for the Mip-NeRF 360 dataset.

\begin{figure*}
    \centering
    \includegraphics[width=\textwidth]{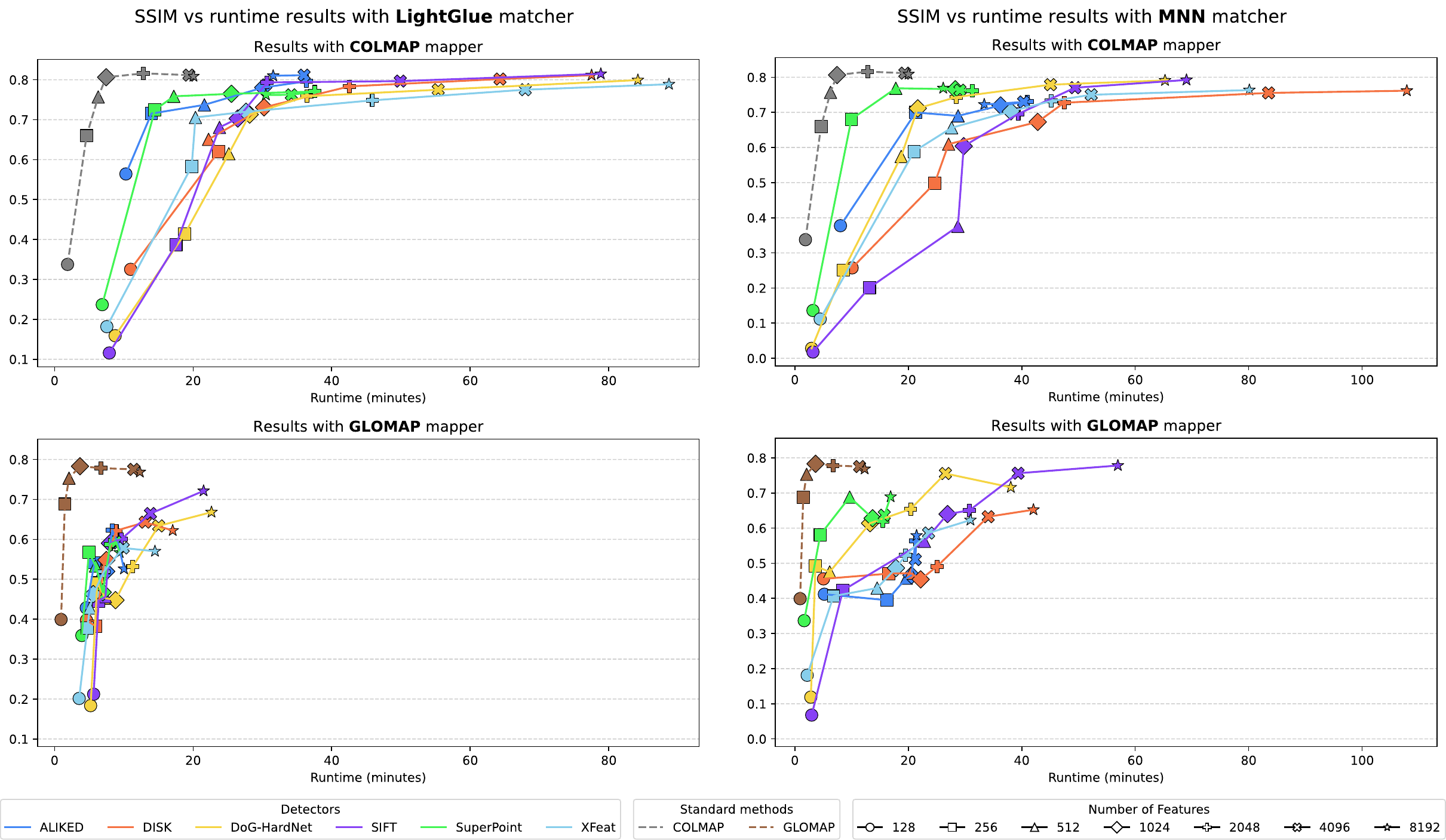}
    \caption{SSIM and runtime results for the Tanks and Temples dataset.}
    \label{fig:tnt_ssim_time}
\end{figure*}

\begin{figure*}
    \centering
    \includegraphics[width=\textwidth]{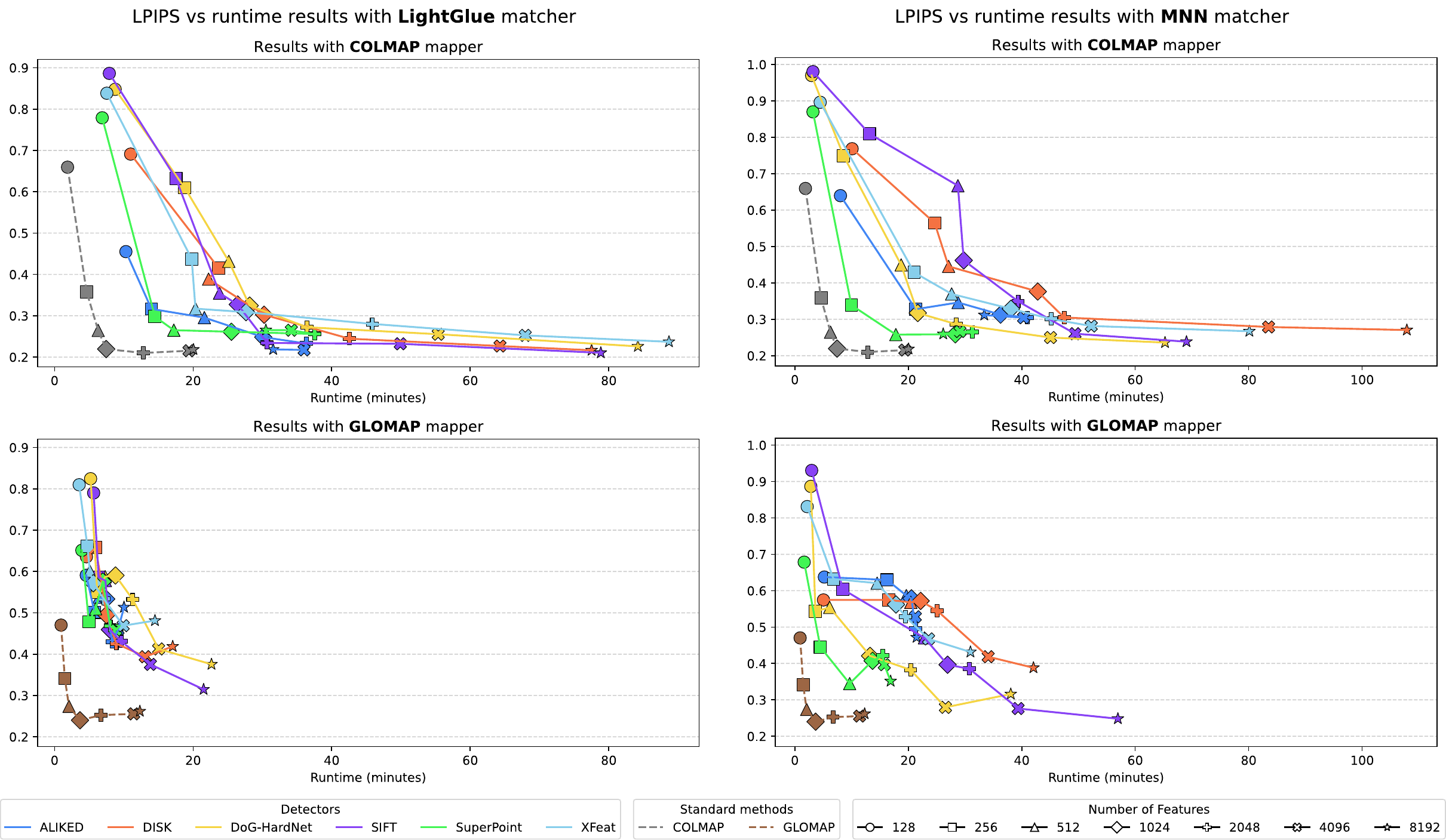}
    \caption{LPIPS and runtime results for the Tanks and Temples dataset.}
    \label{fig:tnt_lpips_time}
\end{figure*}

\begin{figure*}
    \centering
    \includegraphics[width=\textwidth]{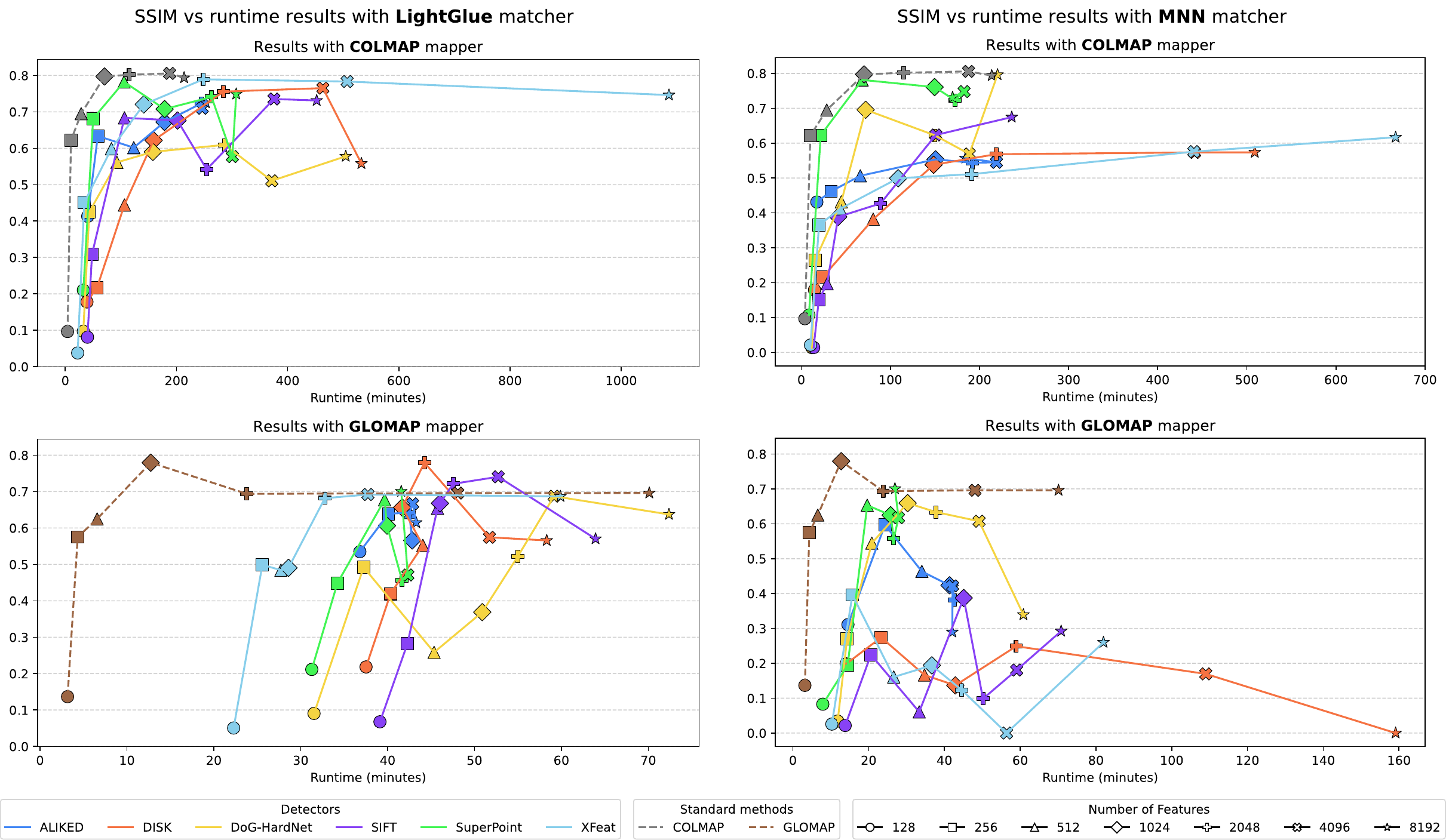}
    \caption{SSIM and runtime results for the Zip-NeRF dataset.}
    \label{fig:zip_ssim_time}
\end{figure*}

\begin{figure*}
    \centering
    \includegraphics[width=\textwidth]{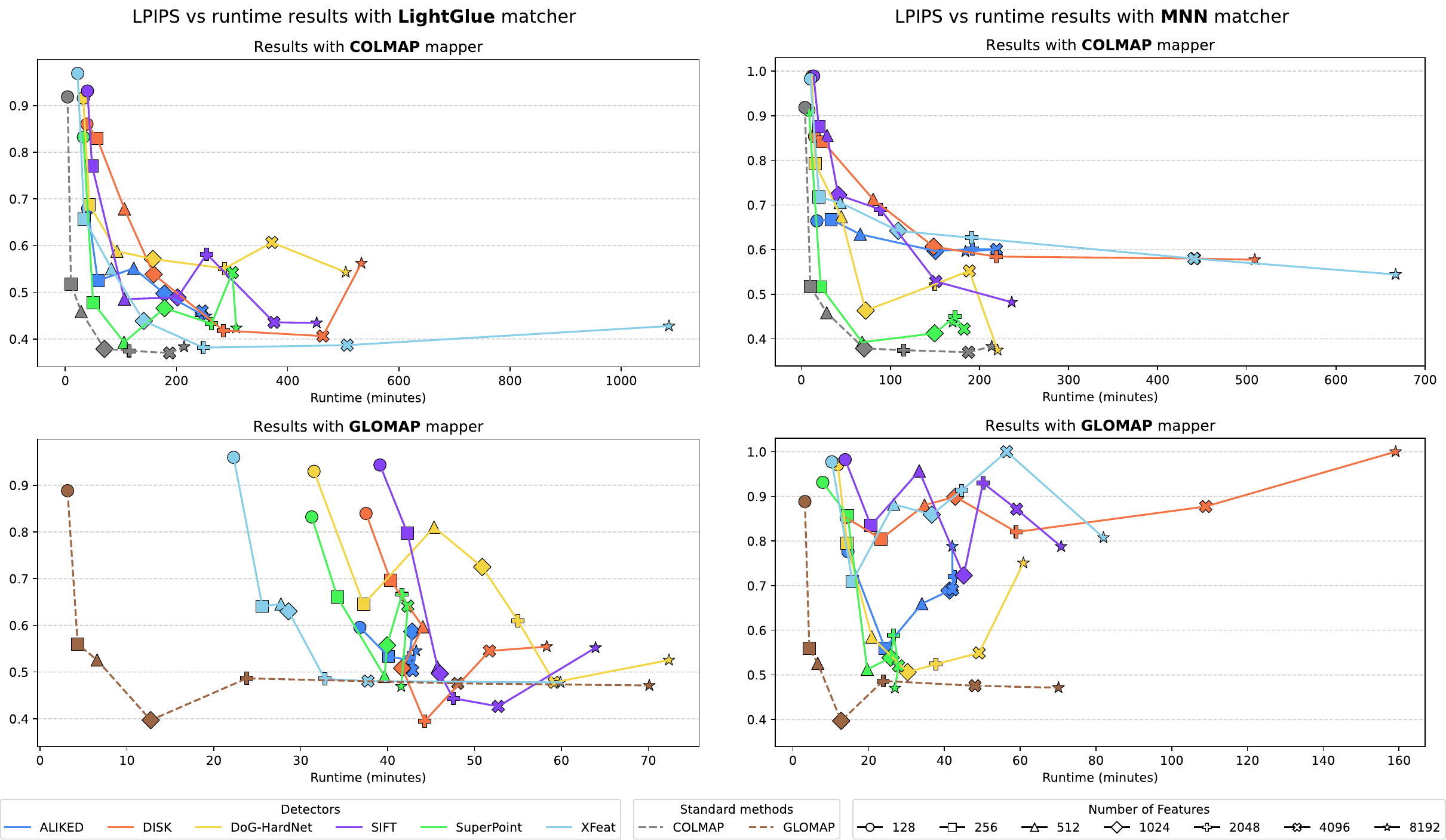}
    \caption{LPIPS and runtime results for the Zip-NeRF dataset.}
    \label{fig:zip_lpips_time}
\end{figure*}

In Figures~\ref{fig:tnt_ssim_time} and~\ref{fig:tnt_lpips_time} we plot the SSIM and LPIPS against the average reconstruction time for the Tanks and Temples datasets. Finally, for the Zip-NeRF dataset, in Figure~\ref{fig:zip_ssim_time} we have SSIM results, and in Figure~\ref{fig:zip_lpips_time} LPIPS plots.

\begin{figure*}
    \centering
    \includegraphics[width=\textwidth]{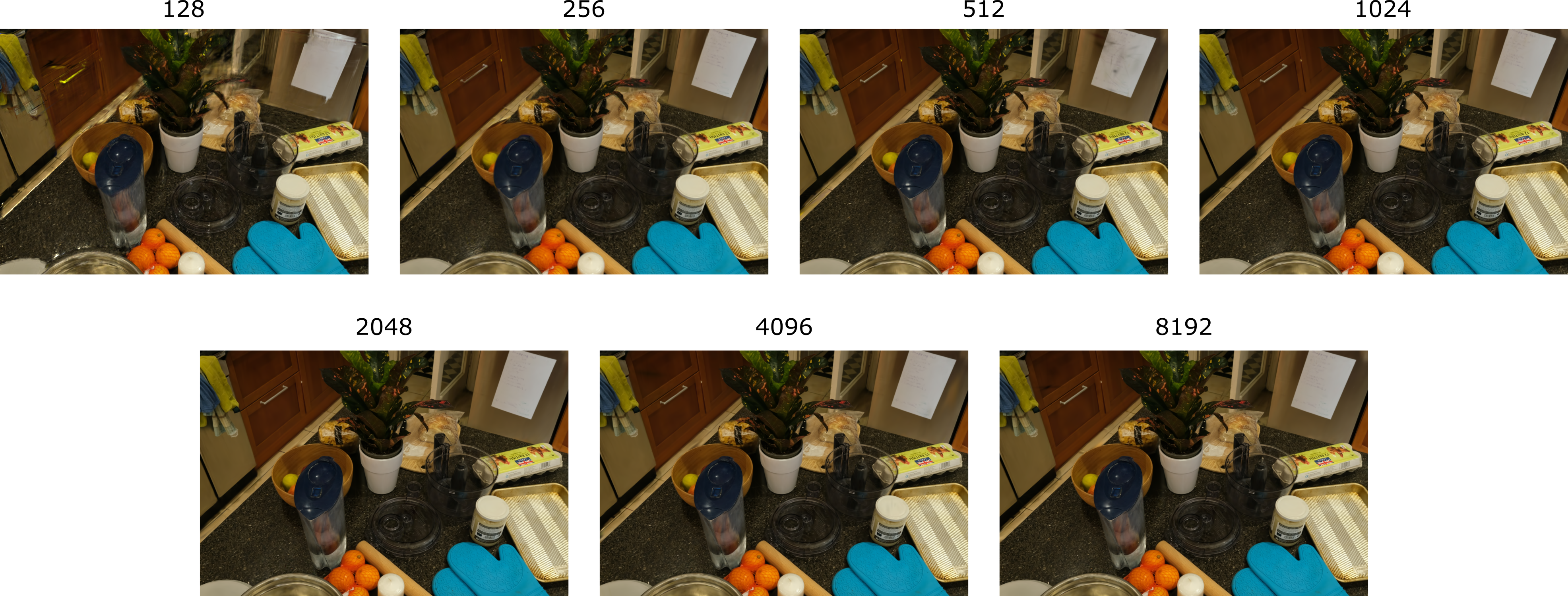}
    \caption{Renderings for different number of keypoints using COLMAP's feature extractor and matcher.}
    \label{fig:renders_room}
\end{figure*}

Finally, in Figure~\ref{fig:renders_room} we show rendering examples for the multiple number of extracted features. The renderings are from the \textit{counter} scene from Mip-NeRF 360 dataset using COLMAP's feature extractor (SIFT) and matcher. With 128 features, we can see many artifacts that are improved with 256 features. The quality of rendering with 512 features is impressive with minimal artifacts. For more than 512 features, we have almost no artifacts in the rendered images. It is important to consider that for 128 and 256 features, we cannot always obtain a proper reconstruction because the mapping stage may fail or because the quality of the camera poses may not be sufficient. In such cases, we are unable to train a 3D Gaussian Splatting or NeRF model.

\section{Per-scene novel view synthesis results}
\label{sec:per_scene_nvs}
We also report the per-scene PSNR results for the different learned features. Table~\ref{tab:psnr_metrics_Mip-NeRF360_colmap_LG} reports the metrics obtained for the different number of extracted features using the LightGlue matcher and COLMAP mapper. 
We can see how SuperPoint performs well for a small number of keypoints. For example, for 512 features, SuperPoint provides the best results in all the scenes. However, see that for 4096 and 8192 features, the top results are obtained with DISK, DoG-HardNet, and SIFT. Also note that for all the scenes, the PSNR differences between the best detectors from 512 and 8192 features are less than 1 dB. This shows that we can get competitive rendering results with few features. In Table~\ref{tab:psnr_metrics_Mip-NeRF360_glomap_LG} we show the results for GLOMAP. For a small number of features ALIKED provides good results compared with the other detectors. For example, for 512 features, the best results are obtained with ALIKED in almost all scenes. For 4096 and 8192 features, SIFT provides the best results in most of the scenes. Also see that unlike COLMAP, the PSNR differences between the best detectors from 512 and 8192 features are less than 1 dB only in 3 scenes. However, between 1024 and 8192 the difference is less than 1 dB for all the scenes except for \textit{bonsai}, \textit{kitchen}, and \textit{treehill}.

\begin{table*}%[h!]
\centering
\resizebox{\textwidth}{!}{
\begin{tabular}{c|l|ccccccccc}
\hline
\textbf{Features} & \textbf{Detector} & bicycle & bonsai & counter & flowers & garden & kitchen & room & stump & treehill \\
\hline
\multirow{6}{*}{\centering 128} & ALIKED & — & \cellcolor{tabfirst}{30.61} & — & — & \cellcolor{tabfirst}{26.41} & \cellcolor{tabthird}{21.39} & \cellcolor{tabfirst}{30.53} & — & \cellcolor{tabfirst}{2.74} \\
 & DISK & \cellcolor{tabsecond}{0.68} & 13.57 & \cellcolor{tabfirst}{27.13} & — & \cellcolor{tabsecond}{26.01} & \cellcolor{tabsecond}{22.41} & \cellcolor{tabthird}{13.40} & — & — \\
 & DoG-HardNet & \cellcolor{tabfirst}{7.03} & 22.71 & \cellcolor{tabsecond}{9.86} & — & 24.58 & 17.36 & \cellcolor{tabsecond}{13.90} & \cellcolor{tabsecond}{4.56} & — \\
 & SIFT & — & \cellcolor{tabthird}{22.82} & \cellcolor{tabthird}{9.49} & — & 24.93 & 17.12 & 11.98 & \cellcolor{tabfirst}{4.61} & — \\
 & SuperPoint & — & \cellcolor{tabsecond}{30.60} & — & — & \cellcolor{tabthird}{25.94} & \cellcolor{tabfirst}{28.78} & 2.08 & — & — \\
 & XFeat & — & — & — & — & — & — & — & — & — \\
\hline
\multirow{6}{*}{\centering 256} & ALIKED & \cellcolor{tabsecond}{24.62} & \cellcolor{tabthird}{29.59} & \cellcolor{tabthird}{28.52} & — & \cellcolor{tabsecond}{26.70} & \cellcolor{tabsecond}{29.78} & 28.72 & — & \cellcolor{tabsecond}{21.77} \\
 & DISK & \cellcolor{tabthird}{22.58} & 20.21 & \cellcolor{tabsecond}{28.61} & \cellcolor{tabsecond}{11.42} & \cellcolor{tabthird}{26.57} & 19.69 & 22.20 & \cellcolor{tabthird}{11.77} & \cellcolor{tabthird}{12.37} \\
 & DoG-HardNet & 18.58 & 23.02 & 27.95 & \cellcolor{tabthird}{1.60} & 26.19 & 20.35 & \cellcolor{tabsecond}{30.58} & \cellcolor{tabsecond}{16.75} & 3.46 \\
 & SIFT & 18.97 & \cellcolor{tabsecond}{30.27} & 27.81 & — & 26.17 & 23.49 & \cellcolor{tabthird}{30.09} & \cellcolor{tabfirst}{17.00} & 3.81 \\
 & SuperPoint & \cellcolor{tabfirst}{24.82} & \cellcolor{tabfirst}{30.30} & \cellcolor{tabfirst}{28.70} & \cellcolor{tabfirst}{15.21} & \cellcolor{tabfirst}{26.94} & \cellcolor{tabfirst}{30.45} & \cellcolor{tabfirst}{30.91} & 7.57 & \cellcolor{tabfirst}{21.78} \\
 & XFeat & — & 28.07 & 27.50 & — & — & \cellcolor{tabthird}{28.03} & 28.71 & — & — \\
\hline
\multirow{6}{*}{\centering 512} & ALIKED & \cellcolor{tabsecond}{24.79} & \cellcolor{tabsecond}{31.55} & \cellcolor{tabthird}{28.64} & \cellcolor{tabsecond}{21.65} & 26.81 & \cellcolor{tabsecond}{29.88} & 23.10 & \cellcolor{tabthird}{25.09} & \cellcolor{tabsecond}{21.94} \\
 & DISK & \cellcolor{tabthird}{24.69} & 12.77 & \cellcolor{tabsecond}{28.79} & \cellcolor{tabthird}{21.65} & \cellcolor{tabsecond}{27.04} & 29.50 & 29.39 & \cellcolor{tabsecond}{25.80} & \cellcolor{tabthird}{21.82} \\
 & DoG-HardNet & 23.06 & 30.29 & 28.34 & 15.47 & \cellcolor{tabthird}{26.98} & 29.18 & \cellcolor{tabsecond}{30.88} & 24.05 & 10.26 \\
 & SIFT & 21.67 & 31.07 & 28.29 & 9.75 & 26.70 & 19.14 & \cellcolor{tabthird}{30.83} & 24.02 & 10.08 \\
 & SuperPoint & \cellcolor{tabfirst}{25.26} & \cellcolor{tabfirst}{31.82} & \cellcolor{tabfirst}{28.83} & \cellcolor{tabfirst}{21.94} & \cellcolor{tabfirst}{27.11} & \cellcolor{tabfirst}{30.65} & \cellcolor{tabfirst}{31.18} & \cellcolor{tabfirst}{26.41} & \cellcolor{tabfirst}{21.98} \\
 & XFeat & 23.73 & \cellcolor{tabthird}{31.37} & 28.39 & — & 26.25 & \cellcolor{tabthird}{29.60} & 30.55 & — & — \\
\hline
\multirow{6}{*}{\centering 1024} & ALIKED & \cellcolor{tabthird}{25.00} & 31.72 & 28.83 & 21.81 & 27.16 & \cellcolor{tabsecond}{30.46} & \cellcolor{tabthird}{30.99} & 25.48 & \cellcolor{tabfirst}{22.26} \\
 & DISK & \cellcolor{tabsecond}{25.20} & \cellcolor{tabfirst}{31.98} & \cellcolor{tabsecond}{28.97} & 21.93 & \cellcolor{tabthird}{27.18} & 29.22 & 30.24 & \cellcolor{tabsecond}{26.39} & \cellcolor{tabthird}{21.87} \\
 & DoG-HardNet & 24.97 & 31.48 & 28.86 & \cellcolor{tabfirst}{22.10} & 27.18 & 29.37 & 29.47 & 25.42 & 19.98 \\
 & SIFT & 24.71 & 31.87 & \cellcolor{tabthird}{28.87} & \cellcolor{tabsecond}{22.04} & \cellcolor{tabsecond}{27.24} & 30.35 & \cellcolor{tabsecond}{31.11} & \cellcolor{tabthird}{25.51} & 20.48 \\
 & SuperPoint & \cellcolor{tabfirst}{25.50} & \cellcolor{tabthird}{31.93} & \cellcolor{tabfirst}{28.99} & \cellcolor{tabthird}{22.03} & \cellcolor{tabfirst}{27.57} & \cellcolor{tabfirst}{31.04} & \cellcolor{tabfirst}{31.28} & \cellcolor{tabfirst}{26.65} & \cellcolor{tabsecond}{22.10} \\
 & XFeat & 24.98 & \cellcolor{tabsecond}{31.95} & 28.83 & 14.02 & 26.97 & \cellcolor{tabthird}{30.38} & 30.96 & 19.50 & 21.82 \\
\hline
\multirow{6}{*}{\centering 2048} & ALIKED & 25.09 & 32.07 & 28.89 & 21.96 & 27.32 & 30.12 & 31.04 & 25.64 & \cellcolor{tabfirst}{22.26} \\
 & DISK & \cellcolor{tabfirst}{25.61} & 32.02 & \cellcolor{tabsecond}{29.04} & 22.09 & \cellcolor{tabthird}{27.43} & 30.11 & \cellcolor{tabsecond}{31.43} & \cellcolor{tabsecond}{26.75} & \cellcolor{tabthird}{22.01} \\
 & DoG-HardNet & \cellcolor{tabthird}{25.55} & 31.77 & \cellcolor{tabthird}{29.03} & \cellcolor{tabfirst}{22.19} & \cellcolor{tabsecond}{27.48} & \cellcolor{tabsecond}{30.76} & \cellcolor{tabfirst}{31.44} & 26.45 & 21.59 \\
 & SIFT & 25.51 & \cellcolor{tabfirst}{32.36} & 29.02 & \cellcolor{tabthird}{22.13} & 27.43 & 30.02 & \cellcolor{tabthird}{31.31} & \cellcolor{tabthird}{26.68} & 21.64 \\
 & SuperPoint & \cellcolor{tabsecond}{25.60} & \cellcolor{tabthird}{32.14} & \cellcolor{tabfirst}{29.17} & \cellcolor{tabsecond}{22.16} & \cellcolor{tabfirst}{27.54} & \cellcolor{tabfirst}{31.02} & 30.78 & \cellcolor{tabfirst}{26.88} & \cellcolor{tabsecond}{22.15} \\
 & XFeat & 25.38 & \cellcolor{tabsecond}{32.29} & 28.93 & 21.66 & 26.86 & \cellcolor{tabthird}{30.60} & 31.22 & 26.09 & 21.94 \\
\hline
\multirow{6}{*}{\centering 4096} & ALIKED & 25.17 & 32.08 & 28.95 & 21.98 & 27.48 & 30.52 & 31.07 & 25.85 & \cellcolor{tabfirst}{22.42} \\
 & DISK & \cellcolor{tabsecond}{25.74} & \cellcolor{tabsecond}{32.57} & 29.08 & \cellcolor{tabsecond}{22.19} & \cellcolor{tabfirst}{27.59} & 30.55 & \cellcolor{tabfirst}{31.72} & \cellcolor{tabthird}{26.92} & \cellcolor{tabsecond}{22.19} \\
 & DoG-HardNet & \cellcolor{tabthird}{25.74} & \cellcolor{tabfirst}{32.62} & 29.14 & \cellcolor{tabfirst}{22.24} & \cellcolor{tabsecond}{27.58} & \cellcolor{tabsecond}{30.79} & \cellcolor{tabsecond}{31.65} & 26.89 & 21.82 \\
 & SIFT & \cellcolor{tabfirst}{25.76} & 32.10 & \cellcolor{tabfirst}{29.17} & 21.89 & 27.44 & \cellcolor{tabfirst}{31.13} & \cellcolor{tabthird}{31.62} & \cellcolor{tabfirst}{27.09} & 21.93 \\
 & SuperPoint & 25.58 & 32.25 & \cellcolor{tabsecond}{29.16} & \cellcolor{tabthird}{22.04} & \cellcolor{tabthird}{27.58} & 21.42 & 31.53 & \cellcolor{tabsecond}{26.96} & \cellcolor{tabthird}{22.15} \\
 & XFeat & 25.72 & \cellcolor{tabthird}{32.36} & \cellcolor{tabthird}{29.14} & 22.02 & 27.29 & \cellcolor{tabthird}{30.73} & 30.74 & 26.67 & 22.01 \\
\hline
\multirow{6}{*}{\centering 8192} & ALIKED & 25.22 & 32.10 & 28.95 & 22.00 & 27.50 & 30.22 & 31.32 & 25.97 & \cellcolor{tabsecond}{22.33} \\
 & DISK & \cellcolor{tabfirst}{25.87} & \cellcolor{tabthird}{32.60} & \cellcolor{tabfirst}{29.27} & \cellcolor{tabthird}{22.09} & \cellcolor{tabfirst}{27.75} & \cellcolor{tabthird}{30.85} & \cellcolor{tabsecond}{31.76} & \cellcolor{tabthird}{27.02} & \cellcolor{tabfirst}{22.36} \\
 & DoG-HardNet & \cellcolor{tabthird}{25.83} & \cellcolor{tabfirst}{32.77} & 28.84 & \cellcolor{tabfirst}{22.36} & \cellcolor{tabsecond}{27.63} & \cellcolor{tabfirst}{31.14} & \cellcolor{tabthird}{31.58} & \cellcolor{tabsecond}{27.14} & 22.02 \\
 & SIFT & \cellcolor{tabsecond}{25.83} & \cellcolor{tabsecond}{32.65} & 29.14 & \cellcolor{tabsecond}{22.22} & \cellcolor{tabthird}{27.62} & \cellcolor{tabsecond}{31.12} & \cellcolor{tabfirst}{31.79} & \cellcolor{tabfirst}{27.16} & 21.95 \\
 & SuperPoint & 25.51 & 32.27 & \cellcolor{tabthird}{29.19} & 21.93 & 27.60 & 21.00 & 31.46 & 26.95 & 21.98 \\
 & XFeat & 25.77 & 32.33 & \cellcolor{tabsecond}{29.22} & 22.08 & 27.51 & 30.81 & 31.32 & 26.84 & \cellcolor{tabthird}{22.20} \\
\hline
\end{tabular}
}
\caption{Per-scene PSNR results on \textbf{Mip-NeRF 360} dataset using \textbf{COLMAP} mapper and \textbf{LG} matcher. Cells with dashes represent cases where the sparse reconstruction process failed. We highlight the \colorbox{tabfirst}{best}, \colorbox{tabsecond}{second}, and \colorbox{tabthird}{third} values for each scene.}
\label{tab:psnr_metrics_Mip-NeRF360_colmap_LG}
\end{table*}

\begin{table*}%[h!]
\centering
\resizebox{\textwidth}{!}{
\begin{tabular}{c|l|ccccccccc}
\hline
\textbf{Features} & \textbf{Detector} & bicycle & bonsai & counter & flowers & garden & kitchen & room & stump & treehill \\
\hline
\multirow{6}{*}{\centering 128} & ALIKED & 1.51 & \cellcolor{tabfirst}{30.61} & \cellcolor{tabfirst}{25.85} & — & \cellcolor{tabfirst}{26.14} & \cellcolor{tabfirst}{25.15} & 14.46 & — & \cellcolor{tabfirst}{4.92} \\
 & DISK & 4.65 & — & \cellcolor{tabsecond}{25.77} & \cellcolor{tabfirst}{2.03} & \cellcolor{tabsecond}{25.44} & 15.21 & 12.75 & \cellcolor{tabthird}{1.16} & \cellcolor{tabthird}{1.33} \\
 & DoG-HardNet & \cellcolor{tabfirst}{6.34} & \cellcolor{tabthird}{22.76} & 7.31 & — & 22.00 & 13.11 & \cellcolor{tabsecond}{25.19} & \cellcolor{tabsecond}{6.18} & 0.79 \\
 & SIFT & \cellcolor{tabsecond}{6.33} & 17.28 & 6.59 & — & \cellcolor{tabthird}{24.54} & 10.42 & 17.70 & \cellcolor{tabfirst}{6.18} & 0.81 \\
 & SuperPoint & \cellcolor{tabthird}{4.81} & \cellcolor{tabsecond}{26.09} & \cellcolor{tabthird}{13.32} & \cellcolor{tabsecond}{0.98} & 24.36 & \cellcolor{tabthird}{15.37} & \cellcolor{tabfirst}{29.92} & 1.05 & \cellcolor{tabsecond}{2.51} \\
 & XFeat & — & 8.43 & 1.32 & — & — & \cellcolor{tabsecond}{19.74} & \cellcolor{tabthird}{18.38} & — & — \\
\hline
\multirow{6}{*}{\centering 256} & ALIKED & \cellcolor{tabfirst}{23.65} & \cellcolor{tabfirst}{25.95} & \cellcolor{tabsecond}{28.30} & \cellcolor{tabsecond}{18.91} & \cellcolor{tabthird}{26.66} & \cellcolor{tabsecond}{15.59} & \cellcolor{tabsecond}{28.36} & 12.05 & \cellcolor{tabfirst}{21.81} \\
 & DISK & \cellcolor{tabsecond}{21.56} & \cellcolor{tabsecond}{19.52} & 23.44 & — & \cellcolor{tabsecond}{26.66} & 15.40 & — & \cellcolor{tabfirst}{17.67} & \cellcolor{tabthird}{12.20} \\
 & DoG-HardNet & 17.29 & 16.64 & 26.05 & \cellcolor{tabthird}{4.08} & 14.80 & — & 14.58 & \cellcolor{tabsecond}{17.52} & 5.91 \\
 & SIFT & 16.47 & 14.87 & \cellcolor{tabthird}{26.76} & 2.31 & 15.11 & \cellcolor{tabthird}{15.58} & \cellcolor{tabthird}{15.81} & 15.36 & 5.15 \\
 & SuperPoint & \cellcolor{tabthird}{18.27} & 16.62 & \cellcolor{tabfirst}{28.64} & \cellcolor{tabfirst}{20.01} & \cellcolor{tabfirst}{26.81} & 15.19 & \cellcolor{tabfirst}{29.18} & \cellcolor{tabthird}{17.13} & \cellcolor{tabsecond}{15.06} \\
 & XFeat & 12.67 & \cellcolor{tabthird}{16.75} & 23.79 & — & 21.51 & \cellcolor{tabfirst}{16.56} & 14.20 & — & 1.42 \\
\hline
\multirow{6}{*}{\centering 512} & ALIKED & \cellcolor{tabfirst}{23.97} & \cellcolor{tabfirst}{25.04} & \cellcolor{tabfirst}{27.70} & \cellcolor{tabsecond}{21.52} & \cellcolor{tabfirst}{26.96} & \cellcolor{tabsecond}{24.32} & \cellcolor{tabfirst}{30.55} & 17.34 & \cellcolor{tabfirst}{21.74} \\
 & DISK & \cellcolor{tabsecond}{23.71} & 13.94 & 13.77 & 12.57 & \cellcolor{tabthird}{26.79} & \cellcolor{tabfirst}{24.86} & \cellcolor{tabsecond}{29.57} & — & \cellcolor{tabsecond}{21.50} \\
 & DoG-HardNet & 15.56 & 15.92 & 23.32 & \cellcolor{tabthird}{20.72} & — & — & 13.80 & \cellcolor{tabsecond}{24.90} & 17.17 \\
 & SIFT & 13.81 & \cellcolor{tabsecond}{24.66} & \cellcolor{tabthird}{25.78} & 18.66 & 22.66 & 15.52 & 23.94 & \cellcolor{tabthird}{24.30} & 15.01 \\
 & SuperPoint & 15.37 & 15.19 & \cellcolor{tabsecond}{26.05} & \cellcolor{tabfirst}{21.73} & \cellcolor{tabsecond}{26.88} & \cellcolor{tabthird}{22.01} & \cellcolor{tabthird}{26.76} & \cellcolor{tabfirst}{26.30} & 12.98 \\
 & XFeat & \cellcolor{tabthird}{21.83} & \cellcolor{tabthird}{16.50} & 24.26 & 5.34 & 24.25 & 19.01 & 14.39 & 11.73 & \cellcolor{tabthird}{17.68} \\
\hline
\multirow{6}{*}{\centering 1024} & ALIKED & \cellcolor{tabsecond}{24.41} & \cellcolor{tabfirst}{26.90} & \cellcolor{tabfirst}{28.22} & \cellcolor{tabsecond}{21.64} & \cellcolor{tabsecond}{27.11} & \cellcolor{tabfirst}{28.61} & \cellcolor{tabsecond}{30.91} & \cellcolor{tabthird}{25.35} & \cellcolor{tabsecond}{14.88} \\
 & DISK & \cellcolor{tabthird}{24.27} & — & \cellcolor{tabsecond}{27.77} & 11.72 & \cellcolor{tabfirst}{27.24} & 19.96 & \cellcolor{tabthird}{30.70} & \cellcolor{tabsecond}{26.33} & \cellcolor{tabfirst}{15.86} \\
 & DoG-HardNet & 14.99 & \cellcolor{tabthird}{18.62} & 25.06 & 11.44 & \cellcolor{tabthird}{27.01} & 15.15 & 21.31 & 17.29 & — \\
 & SIFT & 15.34 & 16.64 & \cellcolor{tabthird}{27.44} & \cellcolor{tabfirst}{21.76} & 19.07 & \cellcolor{tabsecond}{26.86} & 14.70 & 18.03 & \cellcolor{tabthird}{13.55} \\
 & SuperPoint & \cellcolor{tabfirst}{24.88} & \cellcolor{tabsecond}{19.25} & 24.73 & \cellcolor{tabthird}{12.76} & 17.77 & \cellcolor{tabthird}{20.63} & \cellcolor{tabfirst}{30.96} & \cellcolor{tabfirst}{26.70} & 12.30 \\
 & XFeat & — & — & 25.60 & 12.72 & 17.03 & 15.69 & 14.69 & 18.52 & — \\
\hline
\multirow{6}{*}{\centering 2048} & ALIKED & \cellcolor{tabthird}{24.54} & 16.70 & \cellcolor{tabthird}{27.72} & 13.56 & 27.20 & \cellcolor{tabsecond}{23.74} & \cellcolor{tabsecond}{31.03} & 25.58 & \cellcolor{tabfirst}{15.59} \\
 & DISK & \cellcolor{tabsecond}{25.01} & \cellcolor{tabthird}{22.28} & \cellcolor{tabfirst}{28.44} & \cellcolor{tabthird}{13.90} & \cellcolor{tabthird}{27.28} & \cellcolor{tabfirst}{27.51} & \cellcolor{tabfirst}{31.12} & \cellcolor{tabsecond}{26.74} & — \\
 & DoG-HardNet & 24.34 & \cellcolor{tabsecond}{25.15} & 26.27 & 11.57 & — & 21.21 & 28.88 & 26.38 & 11.61 \\
 & SIFT & 15.54 & \cellcolor{tabfirst}{26.94} & \cellcolor{tabsecond}{28.33} & \cellcolor{tabfirst}{22.23} & \cellcolor{tabsecond}{27.30} & 19.66 & 29.97 & \cellcolor{tabthird}{26.47} & \cellcolor{tabsecond}{12.48} \\
 & SuperPoint & \cellcolor{tabfirst}{25.03} & 15.65 & 25.27 & 11.91 & \cellcolor{tabfirst}{27.44} & 19.35 & \cellcolor{tabthird}{30.59} & \cellcolor{tabfirst}{26.80} & \cellcolor{tabthird}{12.22} \\
 & XFeat & — & 18.45 & 26.49 & \cellcolor{tabsecond}{20.79} & 16.15 & \cellcolor{tabthird}{22.25} & 22.91 & 26.20 & — \\
\hline
\multirow{6}{*}{\centering 4096} & ALIKED & 24.70 & 16.38 & \cellcolor{tabthird}{28.53} & \cellcolor{tabsecond}{21.84} & \cellcolor{tabthird}{27.39} & \cellcolor{tabthird}{23.27} & 30.99 & 25.83 & \cellcolor{tabsecond}{20.11} \\
 & DISK & \cellcolor{tabsecond}{25.17} & \cellcolor{tabsecond}{27.98} & \cellcolor{tabsecond}{28.82} & 21.74 & 27.31 & \cellcolor{tabfirst}{30.52} & \cellcolor{tabsecond}{31.14} & \cellcolor{tabsecond}{26.97} & \cellcolor{tabfirst}{20.48} \\
 & DoG-HardNet & — & 25.51 & 27.67 & 11.98 & \cellcolor{tabfirst}{27.51} & 22.37 & 30.55 & 26.79 & 13.23 \\
 & SIFT & \cellcolor{tabfirst}{25.34} & \cellcolor{tabfirst}{31.51} & \cellcolor{tabfirst}{28.86} & \cellcolor{tabfirst}{22.27} & \cellcolor{tabsecond}{27.40} & 21.16 & \cellcolor{tabthird}{31.07} & \cellcolor{tabfirst}{27.01} & 13.63 \\
 & SuperPoint & \cellcolor{tabthird}{24.86} & 25.01 & 27.11 & — & 27.14 & \cellcolor{tabsecond}{25.54} & \cellcolor{tabfirst}{31.25} & \cellcolor{tabthird}{26.83} & 12.36 \\
 & XFeat & 18.59 & \cellcolor{tabthird}{26.83} & 27.49 & \cellcolor{tabthird}{21.78} & 27.26 & 21.27 & 14.57 & 24.15 & \cellcolor{tabthird}{19.05} \\
\hline
\multirow{6}{*}{\centering 8192} & ALIKED & 24.60 & \cellcolor{tabthird}{27.99} & \cellcolor{tabthird}{28.74} & 21.79 & 27.28 & 24.84 & \cellcolor{tabsecond}{31.01} & 25.90 & \cellcolor{tabsecond}{19.25} \\
 & DISK & \cellcolor{tabthird}{25.38} & — & \cellcolor{tabsecond}{28.87} & \cellcolor{tabsecond}{22.19} & \cellcolor{tabsecond}{27.68} & \cellcolor{tabsecond}{31.02} & \cellcolor{tabfirst}{31.60} & \cellcolor{tabthird}{27.05} & \cellcolor{tabfirst}{21.81} \\
 & DoG-HardNet & 25.17 & 16.80 & 27.95 & 21.81 & \cellcolor{tabthird}{27.53} & 20.02 & 22.02 & \cellcolor{tabsecond}{27.12} & \cellcolor{tabthird}{17.31} \\
 & SIFT & \cellcolor{tabfirst}{25.46} & \cellcolor{tabfirst}{31.74} & \cellcolor{tabfirst}{29.06} & \cellcolor{tabfirst}{22.37} & \cellcolor{tabfirst}{27.71} & \cellcolor{tabfirst}{31.53} & 30.55 & \cellcolor{tabfirst}{27.14} & — \\
 & SuperPoint & 24.81 & 19.54 & 26.92 & 12.07 & 27.09 & \cellcolor{tabthird}{28.47} & \cellcolor{tabthird}{30.88} & 26.75 & 12.21 \\
 & XFeat & \cellcolor{tabsecond}{25.41} & \cellcolor{tabsecond}{29.34} & 28.12 & \cellcolor{tabthird}{22.19} & 27.51 & 28.11 & 25.81 & 16.91 & 14.72 \\
\hline
\end{tabular}
}
\caption{Per-scene PSNR results on \textbf{Mip-NeRF 360} dataset using \textbf{GLOMAP} mapper and \textbf{LG} matcher}
\label{tab:psnr_metrics_Mip-NeRF360_glomap_LG}
\end{table*}

In Table~\ref{tab:psnr_metrics_Tanks_Temples_colmap_LG} and Table~\ref{tab:psnr_metrics_Tanks_Temples_glomap_LG} we report the PSNR metrics per-scene for the Tanks and Temples dataset using LG-COLMAP and LG-GLOMAP, respectively. 
Similarly to the Mip-NeRF 360 dataset, we can see that SuperPoint performs well for a small number of keypoints. It is clear to see that we can get competitive renderings with only 512 features and, for some scenes, even with 128 and 256 keypoints. The PSNR difference obtained by using fewer features compared to the maximum number of features (8192) is, for multiple scenes, less than 1~dB. It is also interesting to note that with GLOMAP, for 8192 features, SIFT tends to be the best for almost all the scenes. This dominance is not clear with COLMAP.

\begin{table*}%[h!]
\centering
\resizebox{\textwidth}{!}{
\begin{tabular}{c|l|ccccccccccccccccccccc}
\hline
\textbf{Features} & \textbf{Detector} & auditorium & ballroom & barn & caterpillar & church & courthouse & courtroom & family & francis & horse & ignatius & lighthouse & m60 & meetingroom & museum & palace & panther & playground & temple & train & truck \\
\hline
\multirow{6}{*}{\centering 128} & ALIKED & \cellcolor{tabfirst}{22.17} & \cellcolor{tabfirst}{13.26} & \cellcolor{tabthird}{19.26} & \cellcolor{tabfirst}{22.69} & \cellcolor{tabfirst}{20.36} & \cellcolor{tabfirst}{20.61} & \cellcolor{tabfirst}{14.61} & \cellcolor{tabfirst}{16.68} & 13.83 & \cellcolor{tabfirst}{20.32} & — & \cellcolor{tabfirst}{19.34} & \cellcolor{tabfirst}{26.82} & — & \cellcolor{tabfirst}{11.95} & \cellcolor{tabfirst}{10.54} & \cellcolor{tabfirst}{27.69} & \cellcolor{tabfirst}{13.41} & \cellcolor{tabfirst}{12.11} & \cellcolor{tabfirst}{14.76} & \cellcolor{tabfirst}{9.39} \\
 & DISK & — & \cellcolor{tabsecond}{7.18} & 12.83 & 13.95 & 0.29 & \cellcolor{tabthird}{18.93} & 7.70 & \cellcolor{tabthird}{1.39} & \cellcolor{tabfirst}{17.58} & \cellcolor{tabsecond}{0.73} & \cellcolor{tabsecond}{4.95} & \cellcolor{tabsecond}{14.15} & \cellcolor{tabsecond}{14.35} & \cellcolor{tabsecond}{10.02} & \cellcolor{tabsecond}{9.78} & \cellcolor{tabsecond}{10.30} & \cellcolor{tabsecond}{20.22} & \cellcolor{tabsecond}{4.92} & \cellcolor{tabsecond}{7.95} & \cellcolor{tabsecond}{12.72} & \cellcolor{tabsecond}{3.03} \\
 & DoG-HardNet & — & 0.50 & 2.03 & \cellcolor{tabsecond}{15.21} & \cellcolor{tabthird}{14.48} & 7.39 & \cellcolor{tabthird}{8.92} & — & \cellcolor{tabthird}{16.35} & — & \cellcolor{tabfirst}{8.39} & 5.37 & \cellcolor{tabthird}{0.91} & \cellcolor{tabthird}{6.53} & \cellcolor{tabthird}{4.21} & 2.91 & \cellcolor{tabthird}{0.65} & — & 0.47 & 0.75 & — \\
 & SIFT & — & \cellcolor{tabthird}{0.51} & 1.80 & \cellcolor{tabthird}{14.82} & 14.04 & 4.51 & \cellcolor{tabsecond}{8.97} & 0.62 & \cellcolor{tabsecond}{16.36} & — & \cellcolor{tabthird}{1.62} & — & 0.28 & — & 3.15 & 3.44 & 0.25 & \cellcolor{tabthird}{2.07} & — & \cellcolor{tabthird}{0.75} & — \\
 & SuperPoint & \cellcolor{tabthird}{20.03} & — & \cellcolor{tabfirst}{25.48} & — & 10.76 & \cellcolor{tabsecond}{20.17} & — & \cellcolor{tabsecond}{9.18} & 14.29 & — & — & \cellcolor{tabthird}{11.69} & — & \cellcolor{tabfirst}{13.02} & 3.35 & \cellcolor{tabthird}{6.35} & — & — & 0.97 & 0.72 & — \\
 & XFeat & \cellcolor{tabsecond}{20.40} & — & \cellcolor{tabsecond}{24.09} & — & \cellcolor{tabsecond}{14.60} & 15.56 & — & — & 13.43 & — & — & 11.00 & — & — & — & — & — & — & \cellcolor{tabthird}{3.80} & — & — \\
\hline
\multirow{6}{*}{\centering 256} & ALIKED & \cellcolor{tabsecond}{23.39} & \cellcolor{tabfirst}{17.63} & \cellcolor{tabthird}{18.39} & \cellcolor{tabsecond}{22.84} & \cellcolor{tabfirst}{20.37} & \cellcolor{tabsecond}{16.12} & \cellcolor{tabthird}{15.00} & \cellcolor{tabsecond}{23.86} & \cellcolor{tabfirst}{23.73} & \cellcolor{tabfirst}{23.08} & \cellcolor{tabfirst}{21.04} & 14.51 & \cellcolor{tabfirst}{27.47} & \cellcolor{tabfirst}{24.57} & \cellcolor{tabthird}{12.68} & 9.93 & \cellcolor{tabfirst}{28.27} & \cellcolor{tabfirst}{24.88} & \cellcolor{tabthird}{12.08} & \cellcolor{tabsecond}{15.63} & \cellcolor{tabsecond}{23.54} \\
 & DISK & 21.39 & 11.36 & 16.50 & \cellcolor{tabthird}{22.21} & \cellcolor{tabthird}{18.25} & 15.64 & 12.44 & \cellcolor{tabthird}{22.22} & \cellcolor{tabthird}{19.17} & 15.26 & \cellcolor{tabthird}{19.36} & 16.13 & \cellcolor{tabthird}{26.84} & 11.75 & \cellcolor{tabsecond}{12.83} & 10.02 & \cellcolor{tabthird}{27.98} & \cellcolor{tabthird}{22.69} & 7.43 & 11.41 & \cellcolor{tabthird}{21.98} \\
 & DoG-HardNet & 20.33 & \cellcolor{tabthird}{14.29} & 18.03 & 21.64 & 15.17 & 14.94 & 10.11 & 14.41 & — & 21.37 & 5.82 & \cellcolor{tabsecond}{16.71} & 26.43 & 11.93 & 10.28 & \cellcolor{tabthird}{11.29} & 0.46 & — & — & 1.58 & 14.11 \\
 & SIFT & 19.16 & 12.65 & 8.16 & 21.88 & 15.21 & \cellcolor{tabthird}{16.05} & — & 2.42 & \cellcolor{tabsecond}{19.95} & \cellcolor{tabthird}{21.98} & 17.40 & — & 26.09 & 0.78 & 9.19 & 11.06 & 0.44 & 18.25 & — & 1.39 & 17.89 \\
 & SuperPoint & \cellcolor{tabfirst}{23.55} & 11.19 & \cellcolor{tabfirst}{26.51} & \cellcolor{tabfirst}{22.97} & \cellcolor{tabsecond}{19.27} & \cellcolor{tabfirst}{21.07} & \cellcolor{tabfirst}{20.10} & \cellcolor{tabfirst}{24.10} & 13.96 & \cellcolor{tabsecond}{22.99} & \cellcolor{tabsecond}{20.88} & \cellcolor{tabfirst}{20.71} & \cellcolor{tabsecond}{27.46} & \cellcolor{tabsecond}{18.24} & \cellcolor{tabfirst}{14.18} & \cellcolor{tabfirst}{13.34} & \cellcolor{tabsecond}{28.15} & \cellcolor{tabsecond}{24.77} & \cellcolor{tabfirst}{15.27} & \cellcolor{tabthird}{15.40} & \cellcolor{tabfirst}{23.68} \\
 & XFeat & \cellcolor{tabthird}{21.93} & \cellcolor{tabsecond}{14.99} & \cellcolor{tabsecond}{25.76} & 20.95 & 16.07 & 11.55 & \cellcolor{tabsecond}{19.36} & — & 12.90 & 19.88 & 18.31 & \cellcolor{tabthird}{16.14} & 24.62 & \cellcolor{tabthird}{12.45} & 11.80 & \cellcolor{tabsecond}{11.81} & 17.44 & 16.60 & \cellcolor{tabsecond}{12.72} & \cellcolor{tabfirst}{17.00} & 17.00 \\
\hline
\multirow{6}{*}{\centering 512} & ALIKED & \cellcolor{tabfirst}{23.73} & \cellcolor{tabthird}{14.14} & 20.12 & \cellcolor{tabsecond}{23.01} & \cellcolor{tabfirst}{20.95} & \cellcolor{tabthird}{15.54} & 16.52 & \cellcolor{tabsecond}{24.29} & \cellcolor{tabfirst}{26.98} & \cellcolor{tabsecond}{23.15} & \cellcolor{tabsecond}{21.28} & 19.29 & \cellcolor{tabfirst}{27.84} & \cellcolor{tabsecond}{24.69} & \cellcolor{tabfirst}{14.39} & 10.87 & 27.68 & \cellcolor{tabfirst}{24.98} & \cellcolor{tabsecond}{14.51} & \cellcolor{tabthird}{16.29} & \cellcolor{tabthird}{23.80} \\
 & DISK & 22.29 & 13.67 & 13.08 & \cellcolor{tabfirst}{23.01} & 16.06 & 9.72 & 14.43 & 16.98 & 17.87 & 13.89 & \cellcolor{tabfirst}{21.35} & 16.34 & \cellcolor{tabthird}{27.69} & \cellcolor{tabthird}{24.18} & \cellcolor{tabthird}{13.95} & \cellcolor{tabthird}{11.31} & \cellcolor{tabsecond}{28.26} & 20.32 & 7.88 & \cellcolor{tabsecond}{16.74} & \cellcolor{tabfirst}{23.89} \\
 & DoG-HardNet & \cellcolor{tabthird}{22.92} & 13.63 & 20.26 & 16.66 & 17.25 & 10.30 & \cellcolor{tabthird}{18.43} & 19.78 & 20.45 & 21.94 & 20.63 & 20.29 & 27.16 & 21.11 & 11.27 & 10.86 & \cellcolor{tabthird}{28.11} & 19.09 & — & 13.38 & 17.75 \\
 & SIFT & 22.67 & \cellcolor{tabsecond}{14.58} & \cellcolor{tabthird}{25.63} & 22.73 & 16.72 & \cellcolor{tabsecond}{17.85} & 11.18 & \cellcolor{tabthird}{22.46} & \cellcolor{tabthird}{23.16} & \cellcolor{tabthird}{21.99} & 20.94 & \cellcolor{tabsecond}{20.94} & 27.04 & 17.95 & 10.58 & 11.23 & 28.06 & \cellcolor{tabthird}{24.18} & 9.64 & \cellcolor{tabfirst}{20.32} & 18.84 \\
 & SuperPoint & \cellcolor{tabsecond}{23.72} & 13.51 & \cellcolor{tabfirst}{27.10} & \cellcolor{tabthird}{22.91} & \cellcolor{tabsecond}{19.75} & \cellcolor{tabfirst}{19.25} & \cellcolor{tabfirst}{21.60} & \cellcolor{tabfirst}{24.37} & \cellcolor{tabsecond}{24.23} & \cellcolor{tabfirst}{23.25} & \cellcolor{tabthird}{21.24} & \cellcolor{tabfirst}{21.27} & \cellcolor{tabsecond}{27.80} & \cellcolor{tabfirst}{24.77} & \cellcolor{tabsecond}{14.25} & \cellcolor{tabfirst}{14.08} & \cellcolor{tabfirst}{28.28} & \cellcolor{tabsecond}{24.68} & \cellcolor{tabfirst}{14.99} & 14.87 & \cellcolor{tabsecond}{23.88} \\
 & XFeat & 22.84 & \cellcolor{tabfirst}{20.89} & \cellcolor{tabsecond}{26.63} & 22.30 & \cellcolor{tabthird}{19.70} & 14.80 & \cellcolor{tabsecond}{19.43} & 22.41 & 12.95 & 21.69 & 20.15 & \cellcolor{tabthird}{20.93} & 26.59 & 23.92 & 12.85 & \cellcolor{tabsecond}{13.70} & 21.35 & 23.04 & \cellcolor{tabthird}{13.26} & 14.84 & 16.23 \\
\hline
\multirow{6}{*}{\centering 1024} & ALIKED & \cellcolor{tabsecond}{23.92} & \cellcolor{tabsecond}{19.09} & \cellcolor{tabfirst}{27.42} & \cellcolor{tabthird}{23.11} & \cellcolor{tabfirst}{21.12} & 16.78 & \cellcolor{tabsecond}{21.87} & \cellcolor{tabsecond}{24.37} & \cellcolor{tabsecond}{27.38} & \cellcolor{tabfirst}{23.29} & \cellcolor{tabsecond}{21.33} & 15.86 & 27.77 & \cellcolor{tabsecond}{24.96} & \cellcolor{tabfirst}{18.62} & \cellcolor{tabsecond}{13.85} & \cellcolor{tabthird}{28.28} & \cellcolor{tabfirst}{25.28} & \cellcolor{tabfirst}{15.99} & \cellcolor{tabsecond}{17.20} & \cellcolor{tabfirst}{24.08} \\
 & DISK & 23.42 & 16.98 & 18.87 & 23.07 & 18.75 & \cellcolor{tabthird}{20.72} & 16.40 & \cellcolor{tabthird}{24.05} & 21.78 & \cellcolor{tabthird}{22.81} & 21.20 & 15.61 & \cellcolor{tabfirst}{28.03} & \cellcolor{tabthird}{24.86} & \cellcolor{tabsecond}{17.19} & 11.22 & 19.11 & \cellcolor{tabthird}{24.97} & 9.02 & \cellcolor{tabthird}{16.51} & \cellcolor{tabsecond}{24.07} \\
 & DoG-HardNet & \cellcolor{tabthird}{23.81} & 17.03 & 19.23 & 23.05 & 18.02 & 12.62 & \cellcolor{tabthird}{20.82} & 22.99 & 22.39 & 22.02 & 20.95 & \cellcolor{tabthird}{20.91} & \cellcolor{tabthird}{27.83} & 23.98 & 12.70 & 12.84 & 28.26 & 24.40 & 8.19 & 14.09 & 23.29 \\
 & SIFT & 23.36 & \cellcolor{tabthird}{18.09} & 18.94 & \cellcolor{tabsecond}{23.16} & \cellcolor{tabthird}{19.18} & \cellcolor{tabsecond}{21.01} & \cellcolor{tabfirst}{21.88} & 23.38 & \cellcolor{tabthird}{23.55} & 22.01 & \cellcolor{tabthird}{21.26} & 20.85 & 27.69 & 24.02 & 11.84 & 11.53 & \cellcolor{tabsecond}{28.36} & — & 11.65 & \cellcolor{tabfirst}{20.87} & 20.56 \\
 & SuperPoint & \cellcolor{tabfirst}{23.96} & 15.41 & \cellcolor{tabthird}{24.74} & \cellcolor{tabfirst}{23.20} & \cellcolor{tabsecond}{20.33} & \cellcolor{tabfirst}{21.25} & 17.62 & \cellcolor{tabfirst}{24.37} & \cellcolor{tabfirst}{27.54} & \cellcolor{tabsecond}{23.03} & \cellcolor{tabfirst}{21.52} & \cellcolor{tabfirst}{21.24} & \cellcolor{tabsecond}{28.01} & \cellcolor{tabfirst}{25.14} & \cellcolor{tabthird}{14.97} & \cellcolor{tabthird}{13.42} & \cellcolor{tabfirst}{28.38} & \cellcolor{tabsecond}{25.28} & \cellcolor{tabsecond}{14.80} & 15.08 & \cellcolor{tabthird}{24.00} \\
 & XFeat & 22.77 & \cellcolor{tabfirst}{22.26} & \cellcolor{tabsecond}{26.99} & 22.77 & 18.16 & 16.32 & 20.50 & 23.59 & 13.28 & 22.09 & 20.61 & \cellcolor{tabsecond}{20.94} & 27.07 & 23.52 & 13.23 & \cellcolor{tabfirst}{14.28} & 17.26 & 24.40 & \cellcolor{tabthird}{13.72} & 14.57 & 15.61 \\
\hline
\multirow{6}{*}{\centering 2048} & ALIKED & \cellcolor{tabsecond}{23.86} & \cellcolor{tabsecond}{21.03} & \cellcolor{tabfirst}{27.78} & \cellcolor{tabfirst}{23.25} & \cellcolor{tabfirst}{21.56} & 19.07 & \cellcolor{tabsecond}{22.19} & \cellcolor{tabfirst}{24.61} & \cellcolor{tabthird}{26.95} & \cellcolor{tabfirst}{23.27} & \cellcolor{tabthird}{21.67} & \cellcolor{tabfirst}{21.41} & 27.98 & \cellcolor{tabsecond}{24.86} & 14.84 & \cellcolor{tabsecond}{14.00} & \cellcolor{tabsecond}{28.49} & \cellcolor{tabthird}{24.98} & \cellcolor{tabfirst}{15.92} & \cellcolor{tabfirst}{21.31} & \cellcolor{tabfirst}{24.31} \\
 & DISK & 23.62 & 16.95 & \cellcolor{tabthird}{27.57} & \cellcolor{tabsecond}{23.23} & \cellcolor{tabsecond}{21.15} & \cellcolor{tabsecond}{21.37} & 18.66 & \cellcolor{tabthird}{24.20} & 21.46 & \cellcolor{tabthird}{22.93} & \cellcolor{tabfirst}{21.87} & 17.37 & \cellcolor{tabfirst}{28.11} & \cellcolor{tabfirst}{24.90} & \cellcolor{tabfirst}{18.03} & 13.24 & \cellcolor{tabfirst}{28.56} & \cellcolor{tabfirst}{25.40} & \cellcolor{tabthird}{15.29} & \cellcolor{tabsecond}{21.21} & \cellcolor{tabsecond}{24.17} \\
 & DoG-HardNet & \cellcolor{tabthird}{23.68} & 16.12 & 19.34 & 23.10 & 19.94 & 16.82 & 15.18 & 23.67 & \cellcolor{tabfirst}{27.07} & 22.80 & 20.96 & \cellcolor{tabthird}{20.69} & \cellcolor{tabsecond}{28.05} & 24.70 & \cellcolor{tabthird}{16.64} & 13.06 & \cellcolor{tabthird}{28.42} & 24.83 & 14.50 & 20.82 & 23.78 \\
 & SIFT & 23.62 & \cellcolor{tabthird}{20.78} & \cellcolor{tabsecond}{27.71} & 23.12 & 19.41 & \cellcolor{tabfirst}{21.40} & \cellcolor{tabfirst}{22.34} & 23.71 & 26.68 & 22.68 & 21.26 & \cellcolor{tabsecond}{21.12} & \cellcolor{tabthird}{28.03} & 24.53 & \cellcolor{tabsecond}{17.76} & \cellcolor{tabfirst}{15.79} & 28.33 & 24.85 & 14.24 & \cellcolor{tabthird}{20.96} & 23.54 \\
 & SuperPoint & \cellcolor{tabfirst}{24.22} & 15.03 & 25.99 & \cellcolor{tabthird}{23.13} & \cellcolor{tabthird}{20.78} & \cellcolor{tabthird}{21.18} & 18.12 & \cellcolor{tabsecond}{24.36} & \cellcolor{tabsecond}{27.02} & \cellcolor{tabsecond}{23.14} & \cellcolor{tabsecond}{21.76} & 19.52 & 27.89 & \cellcolor{tabthird}{24.74} & 16.50 & \cellcolor{tabthird}{13.98} & 28.30 & \cellcolor{tabsecond}{25.07} & \cellcolor{tabsecond}{15.52} & 14.97 & \cellcolor{tabthird}{24.00} \\
 & XFeat & 23.21 & \cellcolor{tabfirst}{22.93} & 27.18 & 22.91 & 17.85 & 16.74 & \cellcolor{tabthird}{19.02} & 23.95 & 16.61 & 22.64 & 21.10 & 20.55 & 27.41 & 24.02 & 12.46 & 13.91 & 28.13 & 24.70 & 14.41 & 14.44 & 23.62 \\
\hline
\multirow{6}{*}{\centering 4096} & ALIKED & \cellcolor{tabthird}{23.93} & \cellcolor{tabfirst}{23.63} & \cellcolor{tabthird}{27.70} & 23.20 & 20.93 & 20.50 & \cellcolor{tabsecond}{21.83} & \cellcolor{tabfirst}{24.55} & \cellcolor{tabsecond}{27.65} & \cellcolor{tabsecond}{23.25} & \cellcolor{tabsecond}{21.61} & \cellcolor{tabfirst}{21.54} & 27.97 & \cellcolor{tabthird}{24.87} & \cellcolor{tabfirst}{19.48} & \cellcolor{tabsecond}{15.66} & \cellcolor{tabthird}{28.45} & 24.00 & \cellcolor{tabfirst}{15.97} & \cellcolor{tabfirst}{21.50} & \cellcolor{tabsecond}{24.30} \\
 & DISK & 23.46 & 20.69 & \cellcolor{tabfirst}{27.89} & \cellcolor{tabfirst}{23.32} & \cellcolor{tabfirst}{21.55} & \cellcolor{tabsecond}{21.17} & 18.22 & \cellcolor{tabsecond}{24.46} & 25.33 & \cellcolor{tabfirst}{23.34} & \cellcolor{tabfirst}{21.68} & 19.90 & \cellcolor{tabfirst}{28.16} & \cellcolor{tabfirst}{25.13} & \cellcolor{tabsecond}{19.18} & 13.76 & \cellcolor{tabfirst}{28.59} & \cellcolor{tabsecond}{25.16} & \cellcolor{tabsecond}{15.90} & \cellcolor{tabsecond}{21.48} & \cellcolor{tabfirst}{24.35} \\
 & DoG-HardNet & 23.87 & 19.97 & 26.30 & \cellcolor{tabthird}{23.25} & 20.63 & 16.62 & \cellcolor{tabthird}{18.68} & 24.17 & \cellcolor{tabfirst}{27.79} & \cellcolor{tabthird}{23.06} & 21.12 & 14.93 & \cellcolor{tabsecond}{28.15} & \cellcolor{tabsecond}{24.92} & 14.92 & \cellcolor{tabthird}{14.77} & \cellcolor{tabsecond}{28.50} & \cellcolor{tabfirst}{25.55} & \cellcolor{tabthird}{15.86} & 17.95 & \cellcolor{tabthird}{24.14} \\
 & SIFT & \cellcolor{tabsecond}{24.12} & \cellcolor{tabsecond}{23.62} & \cellcolor{tabsecond}{27.83} & \cellcolor{tabsecond}{23.26} & \cellcolor{tabthird}{20.98} & \cellcolor{tabthird}{21.10} & 17.20 & 23.99 & 26.31 & 22.93 & 21.32 & \cellcolor{tabsecond}{21.04} & \cellcolor{tabthird}{28.00} & 24.43 & \cellcolor{tabthird}{16.12} & \cellcolor{tabfirst}{15.74} & 28.43 & 24.11 & 15.70 & \cellcolor{tabthird}{21.30} & 23.95 \\
 & SuperPoint & \cellcolor{tabfirst}{24.40} & 15.17 & 23.53 & 23.13 & \cellcolor{tabsecond}{21.01} & 20.88 & 16.45 & \cellcolor{tabthird}{24.38} & \cellcolor{tabthird}{27.61} & 23.01 & \cellcolor{tabthird}{21.51} & 19.05 & 27.89 & 24.67 & 14.10 & 13.93 & 28.42 & \cellcolor{tabthird}{25.09} & 15.61 & 14.96 & 23.98 \\
 & XFeat & 23.22 & \cellcolor{tabthird}{23.20} & 27.15 & 23.09 & 20.96 & \cellcolor{tabfirst}{21.21} & \cellcolor{tabfirst}{21.87} & 24.00 & 23.07 & 22.74 & 21.03 & \cellcolor{tabthird}{20.50} & 27.52 & 24.61 & 12.37 & 13.85 & 28.20 & 23.14 & 14.75 & 15.69 & 23.80 \\
\hline
\multirow{6}{*}{\centering 8192} & ALIKED & \cellcolor{tabthird}{23.64} & \cellcolor{tabthird}{22.65} & 27.66 & \cellcolor{tabthird}{23.26} & \cellcolor{tabsecond}{21.39} & 20.87 & \cellcolor{tabsecond}{21.85} & 24.39 & \cellcolor{tabsecond}{27.44} & \cellcolor{tabfirst}{23.35} & \cellcolor{tabsecond}{21.83} & \cellcolor{tabsecond}{21.38} & \cellcolor{tabsecond}{28.06} & \cellcolor{tabsecond}{24.82} & \cellcolor{tabsecond}{19.62} & 15.45 & \cellcolor{tabsecond}{28.39} & 23.96 & 15.78 & \cellcolor{tabsecond}{21.49} & \cellcolor{tabsecond}{24.34} \\
 & DISK & \cellcolor{tabfirst}{23.92} & 22.56 & \cellcolor{tabthird}{27.79} & 23.23 & \cellcolor{tabfirst}{21.68} & \cellcolor{tabthird}{20.99} & 20.23 & \cellcolor{tabthird}{24.41} & 26.02 & \cellcolor{tabsecond}{23.26} & 21.32 & 19.90 & \cellcolor{tabfirst}{28.23} & \cellcolor{tabfirst}{25.06} & \cellcolor{tabthird}{19.58} & \cellcolor{tabthird}{15.52} & \cellcolor{tabfirst}{28.60} & \cellcolor{tabsecond}{24.80} & \cellcolor{tabthird}{16.29} & \cellcolor{tabfirst}{21.62} & \cellcolor{tabfirst}{24.39} \\
 & DoG-HardNet & 23.14 & 18.50 & \cellcolor{tabfirst}{27.85} & \cellcolor{tabfirst}{23.35} & \cellcolor{tabthird}{21.25} & 20.93 & 19.95 & \cellcolor{tabfirst}{24.54} & 27.40 & 23.16 & 21.41 & \cellcolor{tabfirst}{21.60} & \cellcolor{tabthird}{27.95} & 24.59 & 13.87 & \cellcolor{tabfirst}{16.73} & \cellcolor{tabthird}{28.28} & \cellcolor{tabfirst}{25.81} & \cellcolor{tabfirst}{19.03} & 21.04 & \cellcolor{tabthird}{24.12} \\
 & SIFT & 23.27 & \cellcolor{tabfirst}{23.55} & \cellcolor{tabsecond}{27.81} & \cellcolor{tabsecond}{23.29} & 20.91 & \cellcolor{tabsecond}{21.03} & \cellcolor{tabfirst}{22.45} & 24.17 & \cellcolor{tabfirst}{27.59} & 22.93 & \cellcolor{tabthird}{21.55} & \cellcolor{tabthird}{21.04} & 27.93 & \cellcolor{tabthird}{24.68} & \cellcolor{tabfirst}{20.02} & \cellcolor{tabsecond}{15.83} & 28.28 & 21.67 & \cellcolor{tabsecond}{18.99} & \cellcolor{tabthird}{21.19} & 23.97 \\
 & SuperPoint & \cellcolor{tabsecond}{23.75} & 15.13 & 23.58 & 23.20 & 21.15 & \cellcolor{tabfirst}{21.07} & 15.96 & \cellcolor{tabsecond}{24.44} & 27.40 & \cellcolor{tabthird}{23.21} & \cellcolor{tabfirst}{21.95} & 20.16 & 27.93 & 24.65 & 13.58 & 13.78 & 28.25 & \cellcolor{tabthird}{24.80} & 15.44 & 15.18 & 23.96 \\
 & XFeat & 22.24 & \cellcolor{tabsecond}{23.10} & 27.08 & 23.07 & 20.87 & 19.03 & \cellcolor{tabthird}{21.83} & 24.24 & \cellcolor{tabthird}{27.43} & 22.95 & 21.49 & 20.80 & 27.49 & 24.16 & 12.88 & 13.36 & 28.03 & 24.72 & 15.89 & 21.02 & 23.82 \\
\hline
\end{tabular}
}
\caption{Per-scene PSNR results on Tanks and Temples dataset using \textbf{COLMAP} mapper and \textbf{LG} matcher}
\label{tab:psnr_metrics_Tanks_Temples_colmap_LG}
\end{table*}

\begin{table*}%[h!]
\centering
\resizebox{\textwidth}{!}{
\begin{tabular}{c|l|ccccccccccccccccccccc}
\hline
\textbf{Features} & \textbf{Detector} & auditorium & ballroom & barn & caterpillar & church & courthouse & courtroom & family & francis & horse & ignatius & lighthouse & m60 & meetingroom & museum & palace & panther & playground & temple & train & truck \\
\hline
\multirow{6}{*}{\centering 128} & ALIKED & — & \cellcolor{tabfirst}{20.45} & \cellcolor{tabfirst}{25.90} & \cellcolor{tabthird}{12.38} & — & \cellcolor{tabsecond}{14.52} & \cellcolor{tabfirst}{15.57} & — & \cellcolor{tabthird}{14.53} & \cellcolor{tabfirst}{21.85} & \cellcolor{tabsecond}{13.53} & \cellcolor{tabfirst}{17.86} & \cellcolor{tabfirst}{26.83} & \cellcolor{tabsecond}{11.90} & \cellcolor{tabfirst}{13.06} & \cellcolor{tabfirst}{10.69} & \cellcolor{tabfirst}{27.65} & — & \cellcolor{tabsecond}{8.84} & — & — \\
 & DISK & — & \cellcolor{tabsecond}{11.40} & 12.37 & \cellcolor{tabsecond}{13.77} & — & \cellcolor{tabthird}{13.82} & \cellcolor{tabsecond}{14.77} & \cellcolor{tabsecond}{14.15} & 13.13 & \cellcolor{tabsecond}{17.48} & \cellcolor{tabfirst}{15.22} & \cellcolor{tabthird}{12.77} & \cellcolor{tabsecond}{18.59} & — & \cellcolor{tabsecond}{12.25} & — & \cellcolor{tabthird}{23.41} & — & \cellcolor{tabfirst}{9.35} & \cellcolor{tabfirst}{14.29} & \cellcolor{tabfirst}{17.50} \\
 & DoG-HardNet & \cellcolor{tabthird}{7.47} & — & 3.69 & — & — & 6.67 & — & 4.77 & \cellcolor{tabfirst}{24.27} & 6.48 & — & 6.64 & 6.67 & 7.38 & 3.30 & 2.73 & 17.79 & \cellcolor{tabsecond}{6.38} & 0.80 & 6.31 & \cellcolor{tabthird}{2.81} \\
 & SIFT & 5.68 & 8.69 & 5.52 & 9.71 & \cellcolor{tabfirst}{10.50} & 11.52 & \cellcolor{tabthird}{10.88} & 2.92 & \cellcolor{tabsecond}{16.69} & 3.62 & 11.48 & — & 4.67 & 5.09 & 3.49 & 3.01 & 6.00 & — & 0.98 & \cellcolor{tabthird}{8.12} & \cellcolor{tabsecond}{4.37} \\
 & SuperPoint & \cellcolor{tabfirst}{16.45} & 3.87 & \cellcolor{tabsecond}{15.30} & \cellcolor{tabfirst}{20.76} & \cellcolor{tabsecond}{8.16} & \cellcolor{tabfirst}{19.76} & 4.47 & \cellcolor{tabfirst}{18.27} & — & — & \cellcolor{tabthird}{12.58} & \cellcolor{tabsecond}{13.61} & \cellcolor{tabthird}{8.00} & \cellcolor{tabfirst}{15.59} & \cellcolor{tabthird}{4.46} & \cellcolor{tabthird}{4.64} & \cellcolor{tabsecond}{24.14} & \cellcolor{tabfirst}{11.14} & 3.83 & \cellcolor{tabsecond}{13.99} & — \\
 & XFeat & \cellcolor{tabsecond}{10.30} & \cellcolor{tabthird}{10.21} & \cellcolor{tabthird}{13.69} & 6.59 & — & — & 3.55 & \cellcolor{tabthird}{6.11} & 12.34 & \cellcolor{tabthird}{8.95} & — & 11.96 & 1.87 & \cellcolor{tabthird}{8.00} & 3.81 & \cellcolor{tabsecond}{6.48} & 7.34 & \cellcolor{tabthird}{1.36} & \cellcolor{tabthird}{5.36} & 3.86 & — \\
\hline
\multirow{6}{*}{\centering 256} & ALIKED & \cellcolor{tabfirst}{21.45} & — & \cellcolor{tabfirst}{24.58} & \cellcolor{tabfirst}{21.64} & — & \cellcolor{tabthird}{11.96} & 11.37 & \cellcolor{tabsecond}{16.37} & \cellcolor{tabthird}{13.58} & \cellcolor{tabsecond}{21.00} & \cellcolor{tabthird}{19.60} & \cellcolor{tabfirst}{21.16} & \cellcolor{tabfirst}{27.32} & \cellcolor{tabfirst}{13.60} & \cellcolor{tabfirst}{13.43} & \cellcolor{tabfirst}{12.39} & 11.77 & — & \cellcolor{tabfirst}{13.60} & \cellcolor{tabsecond}{18.10} & \cellcolor{tabfirst}{18.47} \\
 & DISK & 15.40 & 11.64 & \cellcolor{tabthird}{19.43} & 11.82 & \cellcolor{tabthird}{13.70} & \cellcolor{tabsecond}{12.05} & 10.92 & \cellcolor{tabthird}{13.90} & — & — & 13.53 & 13.49 & 25.18 & 12.38 & 12.07 & — & — & — & \cellcolor{tabthird}{9.95} & 12.27 & \cellcolor{tabsecond}{17.82} \\
 & DoG-HardNet & \cellcolor{tabthird}{17.95} & \cellcolor{tabfirst}{16.73} & \cellcolor{tabsecond}{20.94} & — & \cellcolor{tabsecond}{13.71} & — & \cellcolor{tabthird}{13.21} & 13.54 & \cellcolor{tabsecond}{14.43} & 15.08 & 13.78 & 17.73 & 23.46 & 12.43 & \cellcolor{tabsecond}{13.30} & \cellcolor{tabthird}{11.80} & \cellcolor{tabthird}{23.84} & \cellcolor{tabsecond}{14.96} & 9.16 & 15.15 & 12.65 \\
 & SIFT & — & — & 13.11 & 11.89 & \cellcolor{tabfirst}{14.23} & — & \cellcolor{tabfirst}{17.94} & \cellcolor{tabfirst}{18.14} & \cellcolor{tabfirst}{16.06} & \cellcolor{tabfirst}{21.74} & \cellcolor{tabsecond}{20.29} & \cellcolor{tabthird}{19.42} & \cellcolor{tabthird}{26.19} & \cellcolor{tabthird}{12.44} & 12.14 & 6.99 & \cellcolor{tabsecond}{26.68} & — & — & \cellcolor{tabfirst}{19.06} & 13.41 \\
 & SuperPoint & \cellcolor{tabsecond}{19.23} & \cellcolor{tabthird}{12.11} & 13.10 & \cellcolor{tabsecond}{15.63} & 13.04 & \cellcolor{tabfirst}{19.65} & \cellcolor{tabsecond}{14.60} & — & 13.06 & \cellcolor{tabthird}{20.32} & \cellcolor{tabfirst}{21.47} & \cellcolor{tabsecond}{20.43} & \cellcolor{tabsecond}{27.11} & \cellcolor{tabsecond}{12.48} & 11.97 & \cellcolor{tabsecond}{12.19} & \cellcolor{tabfirst}{28.07} & \cellcolor{tabfirst}{15.62} & \cellcolor{tabsecond}{10.51} & \cellcolor{tabthird}{17.86} & 15.49 \\
 & XFeat & 16.67 & \cellcolor{tabsecond}{12.18} & 17.16 & \cellcolor{tabthird}{13.86} & 13.61 & — & 11.39 & — & 13.03 & — & 18.60 & 17.09 & 23.65 & — & \cellcolor{tabthird}{13.06} & — & 20.57 & — & 8.97 & 10.85 & \cellcolor{tabthird}{15.50} \\
\hline
\multirow{6}{*}{\centering 512} & ALIKED & \cellcolor{tabsecond}{19.47} & 11.94 & \cellcolor{tabfirst}{25.01} & \cellcolor{tabfirst}{22.23} & — & 12.30 & \cellcolor{tabfirst}{20.78} & \cellcolor{tabsecond}{14.60} & — & 11.92 & \cellcolor{tabsecond}{21.23} & \cellcolor{tabsecond}{21.43} & 14.85 & — & 12.87 & \cellcolor{tabthird}{12.38} & 21.24 & 13.58 & 10.05 & \cellcolor{tabsecond}{12.38} & \cellcolor{tabthird}{13.11} \\
 & DISK & 16.47 & 12.12 & \cellcolor{tabsecond}{20.91} & 14.65 & \cellcolor{tabfirst}{14.55} & 12.00 & \cellcolor{tabsecond}{20.21} & 13.69 & \cellcolor{tabsecond}{14.01} & 13.86 & 11.65 & 18.41 & \cellcolor{tabsecond}{26.98} & \cellcolor{tabsecond}{13.40} & 12.59 & 11.04 & \cellcolor{tabthird}{21.91} & 13.66 & \cellcolor{tabthird}{11.64} & — & 11.81 \\
 & DoG-HardNet & \cellcolor{tabthird}{18.34} & \cellcolor{tabfirst}{13.22} & 14.94 & 12.50 & — & \cellcolor{tabthird}{14.05} & 12.67 & \cellcolor{tabthird}{13.71} & 13.26 & \cellcolor{tabthird}{14.77} & — & \cellcolor{tabfirst}{21.48} & 24.48 & — & \cellcolor{tabthird}{13.02} & \cellcolor{tabsecond}{12.70} & 20.72 & \cellcolor{tabsecond}{14.36} & 11.11 & \cellcolor{tabthird}{12.32} & \cellcolor{tabsecond}{14.33} \\
 & SIFT & \cellcolor{tabfirst}{20.98} & 12.18 & — & 12.27 & — & \cellcolor{tabfirst}{20.06} & 10.85 & — & \cellcolor{tabfirst}{15.05} & \cellcolor{tabsecond}{14.78} & \cellcolor{tabfirst}{21.37} & \cellcolor{tabthird}{21.37} & \cellcolor{tabthird}{26.81} & \cellcolor{tabfirst}{13.75} & 12.64 & — & \cellcolor{tabsecond}{26.88} & — & \cellcolor{tabfirst}{14.18} & \cellcolor{tabfirst}{17.68} & 12.38 \\
 & SuperPoint & 15.33 & \cellcolor{tabthird}{12.25} & — & \cellcolor{tabsecond}{21.86} & — & \cellcolor{tabsecond}{14.15} & \cellcolor{tabthird}{19.03} & 11.24 & \cellcolor{tabthird}{13.90} & 11.08 & \cellcolor{tabthird}{20.79} & 20.05 & \cellcolor{tabfirst}{27.75} & \cellcolor{tabthird}{13.24} & \cellcolor{tabfirst}{14.02} & \cellcolor{tabfirst}{14.00} & \cellcolor{tabfirst}{27.67} & \cellcolor{tabfirst}{20.02} & — & 11.01 & \cellcolor{tabfirst}{20.35} \\
 & XFeat & — & \cellcolor{tabsecond}{12.73} & \cellcolor{tabthird}{20.58} & \cellcolor{tabthird}{19.44} & — & — & 13.76 & \cellcolor{tabfirst}{14.70} & — & \cellcolor{tabfirst}{18.34} & 13.40 & 17.22 & 24.09 & 12.68 & \cellcolor{tabsecond}{13.14} & 11.17 & 18.58 & \cellcolor{tabthird}{14.07} & \cellcolor{tabsecond}{11.64} & 10.98 & 12.98 \\
\hline
\multirow{6}{*}{\centering 1024} & ALIKED & 16.27 & — & \cellcolor{tabsecond}{26.14} & 13.66 & \cellcolor{tabthird}{13.69} & \cellcolor{tabsecond}{14.41} & \cellcolor{tabfirst}{19.80} & \cellcolor{tabthird}{12.40} & \cellcolor{tabthird}{14.25} & \cellcolor{tabsecond}{11.43} & \cellcolor{tabsecond}{21.30} & \cellcolor{tabsecond}{21.46} & \cellcolor{tabsecond}{27.74} & — & — & — & \cellcolor{tabfirst}{28.39} & \cellcolor{tabsecond}{18.25} & \cellcolor{tabsecond}{12.92} & \cellcolor{tabthird}{15.45} & 11.85 \\
 & DISK & \cellcolor{tabthird}{20.12} & \cellcolor{tabthird}{12.38} & 24.92 & 12.59 & — & \cellcolor{tabthird}{13.84} & \cellcolor{tabthird}{12.00} & \cellcolor{tabfirst}{13.45} & \cellcolor{tabfirst}{18.78} & 10.90 & \cellcolor{tabfirst}{21.60} & \cellcolor{tabthird}{21.31} & 26.76 & — & \cellcolor{tabthird}{11.90} & \cellcolor{tabsecond}{11.35} & 27.78 & \cellcolor{tabfirst}{19.46} & 10.29 & 15.45 & \cellcolor{tabsecond}{21.06} \\
 & DoG-HardNet & \cellcolor{tabsecond}{21.06} & 12.00 & 12.26 & 15.37 & \cellcolor{tabsecond}{14.08} & 12.03 & — & 10.02 & — & 10.48 & 19.43 & 20.81 & 15.81 & \cellcolor{tabfirst}{13.88} & 11.09 & — & 17.32 & 11.85 & 10.79 & \cellcolor{tabsecond}{17.71} & 12.21 \\
 & SIFT & \cellcolor{tabfirst}{22.79} & — & \cellcolor{tabfirst}{26.85} & \cellcolor{tabfirst}{21.47} & \cellcolor{tabfirst}{14.38} & \cellcolor{tabfirst}{16.09} & \cellcolor{tabsecond}{12.61} & \cellcolor{tabsecond}{13.34} & \cellcolor{tabsecond}{14.72} & \cellcolor{tabfirst}{14.83} & 13.67 & \cellcolor{tabfirst}{21.66} & \cellcolor{tabthird}{27.29} & \cellcolor{tabsecond}{13.48} & \cellcolor{tabfirst}{13.08} & \cellcolor{tabfirst}{15.81} & \cellcolor{tabthird}{27.85} & \cellcolor{tabthird}{14.83} & \cellcolor{tabfirst}{16.28} & 11.71 & \cellcolor{tabfirst}{21.47} \\
 & SuperPoint & 16.73 & \cellcolor{tabfirst}{13.11} & \cellcolor{tabthird}{25.32} & \cellcolor{tabthird}{16.78} & — & 12.96 & 11.01 & — & 13.21 & — & \cellcolor{tabthird}{20.22} & 16.12 & \cellcolor{tabfirst}{27.84} & \cellcolor{tabthird}{12.55} & 11.43 & \cellcolor{tabthird}{11.24} & \cellcolor{tabsecond}{28.07} & — & — & \cellcolor{tabfirst}{21.15} & \cellcolor{tabthird}{14.38} \\
 & XFeat & 16.71 & \cellcolor{tabsecond}{12.50} & 22.20 & \cellcolor{tabsecond}{19.09} & — & 11.53 & 11.29 & 11.11 & 13.78 & \cellcolor{tabthird}{11.39} & 14.50 & 18.74 & 24.17 & — & \cellcolor{tabsecond}{11.96} & 10.84 & 21.84 & 14.30 & \cellcolor{tabthird}{12.31} & 11.31 & — \\
\hline
\multirow{6}{*}{\centering 2048} & ALIKED & 16.53 & \cellcolor{tabfirst}{13.86} & \cellcolor{tabfirst}{27.14} & \cellcolor{tabsecond}{22.95} & — & \cellcolor{tabsecond}{13.44} & \cellcolor{tabsecond}{13.36} & \cellcolor{tabfirst}{15.20} & 14.77 & 10.90 & 21.30 & \cellcolor{tabthird}{21.65} & \cellcolor{tabsecond}{27.85} & 11.76 & 12.58 & \cellcolor{tabthird}{12.10} & \cellcolor{tabsecond}{28.36} & \cellcolor{tabsecond}{19.21} & \cellcolor{tabsecond}{15.64} & \cellcolor{tabfirst}{21.19} & \cellcolor{tabsecond}{20.45} \\
 & DISK & \cellcolor{tabthird}{21.87} & 11.92 & \cellcolor{tabsecond}{26.11} & \cellcolor{tabfirst}{23.05} & \cellcolor{tabfirst}{17.31} & \cellcolor{tabthird}{12.85} & \cellcolor{tabthird}{12.65} & 11.37 & \cellcolor{tabthird}{15.90} & 11.37 & 21.03 & \cellcolor{tabfirst}{22.35} & \cellcolor{tabfirst}{28.01} & \cellcolor{tabfirst}{14.93} & \cellcolor{tabfirst}{14.37} & 11.02 & \cellcolor{tabfirst}{28.44} & \cellcolor{tabthird}{19.18} & \cellcolor{tabthird}{14.94} & 11.80 & \cellcolor{tabthird}{14.62} \\
 & DoG-HardNet & 20.97 & — & 24.36 & — & \cellcolor{tabthird}{14.25} & 12.79 & 12.19 & 11.11 & \cellcolor{tabsecond}{20.37} & \cellcolor{tabthird}{11.91} & 20.45 & \cellcolor{tabsecond}{22.07} & 26.83 & 12.42 & 11.22 & 11.27 & 27.33 & 16.35 & 9.56 & 14.68 & 12.75 \\
 & SIFT & \cellcolor{tabsecond}{22.08} & \cellcolor{tabthird}{12.18} & \cellcolor{tabthird}{25.90} & \cellcolor{tabthird}{22.87} & \cellcolor{tabsecond}{15.02} & \cellcolor{tabfirst}{14.84} & \cellcolor{tabfirst}{18.90} & \cellcolor{tabthird}{13.54} & \cellcolor{tabfirst}{26.63} & \cellcolor{tabfirst}{13.60} & \cellcolor{tabfirst}{21.80} & 21.01 & \cellcolor{tabthird}{27.77} & — & \cellcolor{tabthird}{13.12} & — & \cellcolor{tabthird}{27.98} & 12.01 & \cellcolor{tabfirst}{18.42} & \cellcolor{tabsecond}{20.90} & 11.72 \\
 & SuperPoint & \cellcolor{tabfirst}{22.76} & \cellcolor{tabsecond}{12.69} & 25.81 & — & 14.12 & 11.33 & 11.80 & \cellcolor{tabsecond}{13.62} & 13.83 & 9.77 & \cellcolor{tabthird}{21.34} & 12.30 & 27.70 & \cellcolor{tabsecond}{13.51} & \cellcolor{tabsecond}{13.31} & \cellcolor{tabfirst}{12.74} & 27.78 & \cellcolor{tabfirst}{22.88} & 11.15 & \cellcolor{tabthird}{15.98} & \cellcolor{tabfirst}{21.47} \\
 & XFeat & 17.24 & 11.92 & 24.40 & — & 13.34 & — & 10.86 & 12.57 & 14.66 & \cellcolor{tabsecond}{12.48} & \cellcolor{tabsecond}{21.39} & 20.99 & 25.51 & \cellcolor{tabthird}{12.64} & 11.98 & \cellcolor{tabsecond}{12.31} & 26.38 & 17.93 & 10.97 & 10.55 & 12.86 \\
\hline
\multirow{6}{*}{\centering 4096} & ALIKED & \cellcolor{tabsecond}{22.09} & 12.04 & \cellcolor{tabthird}{27.16} & \cellcolor{tabthird}{22.73} & — & \cellcolor{tabsecond}{14.93} & \cellcolor{tabsecond}{15.00} & 10.94 & 13.74 & \cellcolor{tabthird}{11.75} & \cellcolor{tabfirst}{21.25} & \cellcolor{tabthird}{21.77} & \cellcolor{tabthird}{27.75} & — & 12.48 & 10.19 & \cellcolor{tabfirst}{28.27} & 19.89 & \cellcolor{tabfirst}{15.60} & 20.64 & 12.65 \\
 & DISK & 19.06 & \cellcolor{tabsecond}{13.45} & \cellcolor{tabfirst}{27.30} & \cellcolor{tabsecond}{23.23} & — & \cellcolor{tabthird}{14.92} & 11.56 & 13.23 & \cellcolor{tabthird}{17.31} & 8.75 & \cellcolor{tabsecond}{21.12} & 20.92 & \cellcolor{tabfirst}{28.06} & \cellcolor{tabsecond}{18.95} & \cellcolor{tabfirst}{15.33} & \cellcolor{tabthird}{11.21} & 21.37 & \cellcolor{tabfirst}{23.86} & \cellcolor{tabthird}{13.31} & \cellcolor{tabfirst}{21.08} & \cellcolor{tabfirst}{23.53} \\
 & DoG-HardNet & 16.11 & \cellcolor{tabfirst}{14.64} & \cellcolor{tabsecond}{27.29} & 22.49 & \cellcolor{tabthird}{13.77} & 14.18 & \cellcolor{tabthird}{12.74} & \cellcolor{tabfirst}{15.34} & 16.09 & \cellcolor{tabfirst}{19.18} & 20.88 & \cellcolor{tabfirst}{22.43} & 18.72 & 13.14 & \cellcolor{tabthird}{13.22} & \cellcolor{tabsecond}{11.24} & 27.38 & 20.87 & 10.41 & \cellcolor{tabthird}{20.70} & 14.79 \\
 & SIFT & \cellcolor{tabfirst}{22.40} & 11.86 & 26.97 & \cellcolor{tabfirst}{23.26} & \cellcolor{tabfirst}{16.38} & \cellcolor{tabfirst}{18.15} & 12.19 & 10.66 & \cellcolor{tabfirst}{26.02} & \cellcolor{tabsecond}{15.94} & \cellcolor{tabthird}{21.11} & \cellcolor{tabsecond}{22.41} & \cellcolor{tabsecond}{27.89} & \cellcolor{tabfirst}{22.55} & \cellcolor{tabsecond}{14.18} & \cellcolor{tabfirst}{14.41} & \cellcolor{tabsecond}{28.02} & 14.21 & \cellcolor{tabsecond}{14.15} & 12.13 & \cellcolor{tabsecond}{23.10} \\
 & SuperPoint & \cellcolor{tabthird}{20.64} & \cellcolor{tabthird}{12.14} & 25.04 & 14.76 & \cellcolor{tabsecond}{13.89} & 11.92 & \cellcolor{tabfirst}{16.38} & \cellcolor{tabsecond}{15.02} & 13.66 & 10.54 & 20.98 & 15.74 & 27.52 & \cellcolor{tabthird}{13.59} & 12.61 & — & \cellcolor{tabthird}{27.57} & \cellcolor{tabthird}{21.73} & 10.04 & \cellcolor{tabsecond}{21.07} & \cellcolor{tabthird}{19.27} \\
 & XFeat & 18.21 & 11.71 & 23.46 & 21.95 & 13.48 & 10.40 & 12.23 & \cellcolor{tabthird}{13.62} & \cellcolor{tabsecond}{22.31} & 10.97 & 20.85 & 20.98 & 25.03 & — & 12.57 & 10.96 & 26.03 & \cellcolor{tabsecond}{22.49} & 11.50 & 16.81 & 13.24 \\
\hline
\multirow{6}{*}{\centering 8192} & ALIKED & — & 11.98 & 23.55 & 22.72 & 13.24 & \cellcolor{tabthird}{13.31} & — & — & — & 9.49 & 20.81 & 20.16 & 27.60 & \cellcolor{tabsecond}{14.15} & \cellcolor{tabsecond}{13.88} & \cellcolor{tabsecond}{11.18} & \cellcolor{tabfirst}{28.45} & 19.69 & 11.38 & \cellcolor{tabfirst}{21.34} & \cellcolor{tabthird}{22.45} \\
 & DISK & \cellcolor{tabsecond}{21.72} & 12.08 & \cellcolor{tabfirst}{27.01} & \cellcolor{tabsecond}{23.09} & 13.33 & \cellcolor{tabsecond}{14.77} & \cellcolor{tabthird}{13.77} & 10.56 & 18.82 & 9.14 & \cellcolor{tabsecond}{21.01} & \cellcolor{tabsecond}{21.06} & \cellcolor{tabfirst}{27.86} & — & \cellcolor{tabfirst}{18.33} & — & \cellcolor{tabsecond}{28.13} & \cellcolor{tabfirst}{23.55} & 11.35 & \cellcolor{tabsecond}{21.32} & \cellcolor{tabfirst}{24.11} \\
 & DoG-HardNet & \cellcolor{tabfirst}{21.73} & \cellcolor{tabthird}{12.47} & \cellcolor{tabthird}{26.79} & \cellcolor{tabthird}{22.93} & \cellcolor{tabthird}{14.66} & 11.35 & 12.26 & \cellcolor{tabsecond}{14.67} & \cellcolor{tabsecond}{24.46} & 10.28 & 20.26 & \cellcolor{tabthird}{20.89} & \cellcolor{tabthird}{27.63} & 11.93 & 11.48 & 11.10 & 27.40 & \cellcolor{tabthird}{21.99} & \cellcolor{tabfirst}{19.73} & \cellcolor{tabthird}{21.27} & \cellcolor{tabsecond}{23.59} \\
 & SIFT & \cellcolor{tabthird}{21.62} & \cellcolor{tabfirst}{14.55} & \cellcolor{tabsecond}{26.97} & \cellcolor{tabfirst}{23.23} & \cellcolor{tabfirst}{17.10} & \cellcolor{tabfirst}{15.57} & \cellcolor{tabfirst}{21.64} & \cellcolor{tabfirst}{19.29} & \cellcolor{tabfirst}{27.29} & \cellcolor{tabfirst}{14.26} & \cellcolor{tabfirst}{21.22} & \cellcolor{tabfirst}{22.23} & \cellcolor{tabsecond}{27.67} & \cellcolor{tabfirst}{23.26} & 12.97 & 11.08 & \cellcolor{tabthird}{27.81} & 18.56 & \cellcolor{tabsecond}{19.20} & 21.04 & 22.12 \\
 & SuperPoint & 21.60 & \cellcolor{tabsecond}{14.09} & 26.27 & — & 14.28 & 11.38 & \cellcolor{tabsecond}{21.34} & \cellcolor{tabthird}{13.82} & — & \cellcolor{tabsecond}{11.42} & \cellcolor{tabthird}{20.99} & 12.55 & 27.56 & 11.96 & \cellcolor{tabthird}{13.28} & \cellcolor{tabfirst}{11.37} & 27.49 & \cellcolor{tabsecond}{22.30} & \cellcolor{tabthird}{13.91} & 20.78 & 13.08 \\
 & XFeat & 16.12 & 12.13 & 24.18 & 13.63 & \cellcolor{tabsecond}{14.69} & 12.41 & 13.14 & 10.87 & \cellcolor{tabthird}{21.83} & \cellcolor{tabthird}{11.34} & 20.17 & 19.17 & 25.80 & \cellcolor{tabthird}{13.04} & 12.36 & \cellcolor{tabthird}{11.11} & 26.14 & 14.45 & 10.19 & 12.00 & 16.15 \\
\hline
\end{tabular}
}
\caption{Per-scene PSNR results on Tanks and Temples dataset using \textbf{GLOMAP} mapper and \textbf{LG} matcher}
\label{tab:psnr_metrics_Tanks_Temples_glomap_LG}
\end{table*}

Finally, we show the per-scene rendering results of the Zip-NeRF dataset for COLMAP and LG
in Table~\ref{tab:psnr_metrics_Zip-NeRF_colmap_LG} and for GLOMAP and LG in Table~\ref{tab:psnr_metrics_Zip-NeRF_glomap_LG}. Again, the comments for the previous datasets are also valid for this one: SuperPoint tends to be the best for fewer features (512 and less), and for the same feature extractor, more features improve the rendering metric, but the improvement is marginal. Also see that for \textit{berlin} scene, GLOMAP struggles to provide competitive PSNR results compared to COLMAP.

\begin{table}%[h]
\centering
\begin{minipage}{0.47\textwidth}
    \centering
    \resizebox{\linewidth}{!}{
    \begin{tabular}{c|l|cccc}
    \hline
    \textbf{Features} & \textbf{Detector} & alameda & berlin & london & nyc \\
    \hline
    \multirow{6}{*}{\centering 128} & ALIKED & \cellcolor{tabfirst}{3.67} & \cellcolor{tabfirst}{24.90} & — & \cellcolor{tabfirst}{22.38} \\
     & DISK & \cellcolor{tabthird}{2.70} & \cellcolor{tabthird}{8.29} & \cellcolor{tabsecond}{1.99} & \cellcolor{tabsecond}{7.20} \\
     & DoG-HardNet & 2.16 & 2.56 & \cellcolor{tabthird}{1.96} & 5.00 \\
     & SIFT & 1.46 & 3.00 & 1.82 & 3.44 \\
     & SuperPoint & \cellcolor{tabsecond}{3.33} & \cellcolor{tabsecond}{10.20} & \cellcolor{tabfirst}{5.69} & \cellcolor{tabthird}{7.14} \\
     & XFeat & 0.11 & 4.29 & — & — \\
    \hline
    \multirow{6}{*}{\centering 256} & ALIKED & \cellcolor{tabthird}{9.74} & \cellcolor{tabfirst}{26.35} & \cellcolor{tabfirst}{13.32} & \cellcolor{tabsecond}{23.83} \\
     & DISK & — & \cellcolor{tabthird}{21.39} & 4.50 & — \\
     & DoG-HardNet & \cellcolor{tabsecond}{10.94} & 16.86 & 7.68 & 10.57 \\
     & SIFT & 8.54 & 16.94 & 5.16 & 3.71 \\
     & SuperPoint & \cellcolor{tabfirst}{17.97} & \cellcolor{tabsecond}{26.23} & \cellcolor{tabsecond}{13.21} & \cellcolor{tabfirst}{25.21} \\
     & XFeat & 2.95 & 16.28 & \cellcolor{tabthird}{7.76} & \cellcolor{tabthird}{23.12} \\
    \hline
    \multirow{6}{*}{\centering 512} & ALIKED & 11.69 & 24.23 & \cellcolor{tabthird}{14.60} & \cellcolor{tabthird}{12.65} \\
     & DISK & \cellcolor{tabthird}{12.48} & 15.97 & 7.66 & 5.97 \\
     & DoG-HardNet & \cellcolor{tabsecond}{13.40} & 25.15 & 8.56 & 11.77 \\
     & SIFT & 11.29 & \cellcolor{tabsecond}{26.52} & \cellcolor{tabsecond}{19.27} & \cellcolor{tabsecond}{24.65} \\
     & SuperPoint & \cellcolor{tabfirst}{20.73} & \cellcolor{tabfirst}{26.59} & \cellcolor{tabfirst}{23.00} & \cellcolor{tabfirst}{25.74} \\
     & XFeat & 11.87 & \cellcolor{tabthird}{25.79} & 14.39 & 11.91 \\
    \hline
    \multirow{6}{*}{\centering 1024} & ALIKED & 19.44 & 18.08 & \cellcolor{tabsecond}{15.90} & \cellcolor{tabthird}{19.47} \\
     & DISK & 12.02 & 17.71 & 14.43 & \cellcolor{tabsecond}{21.05} \\
     & DoG-HardNet & 11.66 & 22.19 & 14.53 & 12.62 \\
     & SIFT & \cellcolor{tabthird}{19.64} & \cellcolor{tabsecond}{26.63} & 14.78 & 16.45 \\
     & SuperPoint & \cellcolor{tabfirst}{20.60} & \cellcolor{tabfirst}{26.72} & \cellcolor{tabfirst}{20.47} & 13.68 \\
     & XFeat & \cellcolor{tabsecond}{19.90} & \cellcolor{tabthird}{26.50} & \cellcolor{tabthird}{14.96} & \cellcolor{tabfirst}{25.28} \\
    \hline
    \multirow{6}{*}{\centering 2048} & ALIKED & \cellcolor{tabfirst}{20.79} & 18.29 & 21.06 & 20.13 \\
     & DISK & \cellcolor{tabsecond}{20.70} & 19.72 & 24.74 & \cellcolor{tabsecond}{23.10} \\
     & DoG-HardNet & 13.26 & \cellcolor{tabthird}{23.01} & 15.30 & 12.16 \\
     & SIFT & 20.55 & — & \cellcolor{tabsecond}{25.18} & \cellcolor{tabthird}{20.60} \\
     & SuperPoint & 20.50 & \cellcolor{tabfirst}{26.76} & \cellcolor{tabfirst}{25.35} & 14.84 \\
     & XFeat & \cellcolor{tabthird}{20.69} & \cellcolor{tabsecond}{26.60} & \cellcolor{tabthird}{25.07} & \cellcolor{tabfirst}{25.57} \\
    \hline
    \multirow{6}{*}{\centering 4096} & ALIKED & 20.84 & 14.83 & 21.61 & 20.09 \\
     & DISK & \cellcolor{tabsecond}{20.91} & \cellcolor{tabthird}{20.27} & \cellcolor{tabfirst}{25.50} & 22.95 \\
     & DoG-HardNet & 11.97 & — & 22.97 & \cellcolor{tabsecond}{24.19} \\
     & SIFT & \cellcolor{tabfirst}{20.96} & \cellcolor{tabsecond}{26.69} & \cellcolor{tabsecond}{25.43} & 13.07 \\
     & SuperPoint & 20.44 & — & \cellcolor{tabthird}{25.41} & \cellcolor{tabfirst}{25.95} \\
     & XFeat & \cellcolor{tabthird}{20.86} & \cellcolor{tabfirst}{26.72} & 25.14 & \cellcolor{tabthird}{23.81} \\
    \hline
    \multirow{6}{*}{\centering 8192} & ALIKED & \cellcolor{tabthird}{20.85} & 18.01 & 21.85 & 19.85 \\
     & DISK & 20.69 & — & \cellcolor{tabfirst}{25.55} & \cellcolor{tabthird}{20.24} \\
     & DoG-HardNet & \cellcolor{tabsecond}{20.97} & — & \cellcolor{tabthird}{25.50} & \cellcolor{tabsecond}{24.29} \\
     & SIFT & \cellcolor{tabfirst}{21.04} & \cellcolor{tabfirst}{26.88} & \cellcolor{tabsecond}{25.54} & 12.10 \\
     & SuperPoint & 18.85 & \cellcolor{tabthird}{19.76} & 22.67 & \cellcolor{tabfirst}{25.92} \\
     & XFeat & 20.37 & \cellcolor{tabsecond}{26.70} & 25.29 & 16.31 \\
    \hline
    \end{tabular}
    }
    \caption{Per-scene PSNR results on \textbf{Zip-NeRF} dataset using \textbf{COLMAP} mapper and \textbf{LG} matcher}
    \label{tab:psnr_metrics_Zip-NeRF_colmap_LG}
\end{minipage}
\hfill
\begin{minipage}{0.47\textwidth}
    \centering
    \resizebox{\linewidth}{!}{
    \begin{tabular}{c|l|cccc}
    \hline
    \textbf{Features} & \textbf{Detector} & alameda & berlin & london & nyc \\
    \hline
    \multirow{6}{*}{\centering 128} & ALIKED & \cellcolor{tabfirst}{13.38} & \cellcolor{tabfirst}{22.02} & \cellcolor{tabfirst}{13.36} & \cellcolor{tabfirst}{11.51} \\
     & DISK & — & \cellcolor{tabsecond}{10.09} & — & \cellcolor{tabthird}{8.87} \\
     & DoG-HardNet & \cellcolor{tabthird}{3.49} & 3.30 & \cellcolor{tabthird}{0.96} & 2.32 \\
     & SIFT & 1.75 & 2.31 & 0.81 & 3.06 \\
     & SuperPoint & \cellcolor{tabsecond}{7.29} & \cellcolor{tabthird}{4.01} & \cellcolor{tabsecond}{3.34} & \cellcolor{tabsecond}{11.12} \\
     & XFeat & 1.30 & 3.43 & — & 0.54 \\
    \hline
    \multirow{6}{*}{\centering 256} & ALIKED & \cellcolor{tabfirst}{13.76} & \cellcolor{tabfirst}{24.23} & \cellcolor{tabfirst}{21.01} & 11.09 \\
     & DISK & \cellcolor{tabthird}{12.41} & \cellcolor{tabthird}{13.47} & — & 11.58 \\
     & DoG-HardNet & 10.71 & 11.77 & \cellcolor{tabthird}{10.90} & \cellcolor{tabsecond}{12.52} \\
     & SIFT & 8.90 & — & 5.96 & 11.63 \\
     & SuperPoint & \cellcolor{tabsecond}{12.43} & — & \cellcolor{tabsecond}{15.18} & \cellcolor{tabfirst}{25.34} \\
     & XFeat & 11.29 & \cellcolor{tabsecond}{13.55} & 8.60 & \cellcolor{tabthird}{11.82} \\
    \hline
    \multirow{6}{*}{\centering 512} & ALIKED & 11.48 & 13.16 & \cellcolor{tabfirst}{24.11} & 17.92 \\
     & DISK & \cellcolor{tabthird}{12.46} & 14.62 & 14.13 & 11.34 \\
     & DoG-HardNet & — & — & 14.03 & 11.59 \\
     & SIFT & \cellcolor{tabsecond}{13.14} & \cellcolor{tabsecond}{23.24} & 14.28 & \cellcolor{tabfirst}{24.04} \\
     & SuperPoint & \cellcolor{tabfirst}{18.46} & \cellcolor{tabfirst}{24.63} & \cellcolor{tabthird}{15.21} & \cellcolor{tabsecond}{19.63} \\
     & XFeat & — & \cellcolor{tabthird}{15.46} & \cellcolor{tabsecond}{15.56} & \cellcolor{tabthird}{18.39} \\
    \hline
    \multirow{6}{*}{\centering 1024} & ALIKED & 11.74 & 13.38 & \cellcolor{tabthird}{15.32} & 12.92 \\
     & DISK & \cellcolor{tabthird}{12.59} & \cellcolor{tabsecond}{22.74} & \cellcolor{tabsecond}{15.33} & \cellcolor{tabthird}{22.76} \\
     & DoG-HardNet & — & \cellcolor{tabthird}{21.00} & — & 19.58 \\
     & SIFT & 12.50 & \cellcolor{tabfirst}{24.61} & 14.34 & \cellcolor{tabfirst}{25.30} \\
     & SuperPoint & \cellcolor{tabfirst}{17.29} & 13.09 & \cellcolor{tabfirst}{17.37} & 13.12 \\
     & XFeat & \cellcolor{tabsecond}{13.05} & 14.12 & — & \cellcolor{tabsecond}{24.00} \\
    \hline
    \multirow{6}{*}{\centering 2048} & ALIKED & 15.21 & 23.15 & \cellcolor{tabthird}{15.09} & 17.47 \\
     & DISK & \cellcolor{tabfirst}{20.64} & \cellcolor{tabthird}{24.44} & \cellcolor{tabfirst}{25.00} & \cellcolor{tabsecond}{25.22} \\
     & DoG-HardNet & 12.51 & \cellcolor{tabsecond}{24.75} & — & 21.65 \\
     & SIFT & \cellcolor{tabsecond}{20.31} & \cellcolor{tabfirst}{25.55} & 14.82 & \cellcolor{tabfirst}{25.62} \\
     & SuperPoint & 11.46 & 20.18 & — & 13.67 \\
     & XFeat & \cellcolor{tabthird}{15.35} & 23.64 & \cellcolor{tabsecond}{16.32} & \cellcolor{tabthird}{23.62} \\
    \hline
    \multirow{6}{*}{\centering 4096} & ALIKED & 15.98 & \cellcolor{tabthird}{22.37} & 15.39 & 20.77 \\
     & DISK & \cellcolor{tabsecond}{20.56} & — & \cellcolor{tabfirst}{25.31} & \cellcolor{tabsecond}{25.67} \\
     & DoG-HardNet & 15.98 & \cellcolor{tabfirst}{25.05} & 14.55 & 24.63 \\
     & SIFT & \cellcolor{tabfirst}{20.68} & 13.82 & \cellcolor{tabsecond}{24.97} & \cellcolor{tabfirst}{25.86} \\
     & SuperPoint & 11.47 & — & \cellcolor{tabthird}{17.12} & \cellcolor{tabthird}{25.06} \\
     & XFeat & \cellcolor{tabthird}{16.14} & \cellcolor{tabsecond}{24.41} & 16.76 & 23.97 \\
    \hline
    \multirow{6}{*}{\centering 8192} & ALIKED & 14.97 & \cellcolor{tabfirst}{25.04} & 14.25 & 12.99 \\
     & DISK & \cellcolor{tabsecond}{19.09} & — & \cellcolor{tabfirst}{25.23} & \cellcolor{tabsecond}{25.74} \\
     & DoG-HardNet & 15.61 & \cellcolor{tabsecond}{24.81} & \cellcolor{tabthird}{16.38} & 12.24 \\
     & SIFT & \cellcolor{tabfirst}{20.76} & — & \cellcolor{tabsecond}{24.31} & \cellcolor{tabfirst}{25.79} \\
     & SuperPoint & 16.58 & \cellcolor{tabthird}{24.27} & 16.09 & \cellcolor{tabthird}{25.18} \\
     & XFeat & \cellcolor{tabthird}{17.50} & 23.09 & 15.00 & 24.19 \\
    \hline
    \end{tabular}
    }
    \caption{Per-scene PSNR results on \textbf{Zip-NeRF} dataset using \textbf{GLOMAP} mapper and \textbf{LG} matcher}
    \label{tab:psnr_metrics_Zip-NeRF_glomap_LG}
\end{minipage}
\end{table}

\section{Additional feed-forward refinement results}
\label{sec:ffn_refine}
In this section, we present the PSNR renderings and reconstruction time results of the feed-forward reconstruction refinement experiments for all the feed-forward models used in the main paper: Depth-Anything-3 (DA3)~\cite{lin2025depth}, $\pi^3$~\cite{wang2025pi}, VGGT~\cite{wang2025vggt}, and Fast3R~\cite{yang2025fast3r} using different number of features during the refinement process. In Tables~\ref{tab:ffn_refine_Mip-NeRF360_8192feats}, \ref{tab:ffn_refine_Mip-NeRF360_4096feats}, and \ref{tab:ffn_refine_Mip-NeRF360_2048feats} we show the refinement results for 8192, 4096, and 2048 features, respectively, on the Mip-NeRF 360 dataset. We can see that the best performance is obtained when we extract 8192 keypoints for the Bundle Adjustment (BA) refinement process. However, as expected, we also get slower timings when we use fewer features. Furthermore, note that for Fast3R and VGGT we are able to improve the PSNR metrics with our refinement approach, however; the performance is still poor compared with the other refined feed-forward models, and standard COLMAP. This is because the initial quality of the camera poses is not good compared to DA3 and $\pi^3$.

\begin{table*}%[]
\centering
\resizebox{\textwidth}{!}{
\begin{tabular}{lcc|cc|cc|cc|cc|cc|cc|cc|cc|cc|cc|cc|cc}
\toprule
& \multicolumn{10}{c|}{\textbf{Standard COLMAP reconstruction}} & \multicolumn{16}{c}{\textbf{Feed-forward recons}} \\
 & \multicolumn{2}{c|}{512} & \multicolumn{2}{c|}{1024} & \multicolumn{2}{c|}{2048} & \multicolumn{2}{c|}{4096} & \multicolumn{2}{c|}{8192} & \multicolumn{2}{c|}{DA3} & \multicolumn{2}{c|}{Refined DA3} & \multicolumn{2}{c|}{$\pi^3$} & \multicolumn{2}{c|}{Refined $\pi^3$} & \multicolumn{2}{c|}{Fast3R} & \multicolumn{2}{c|}{Refined Fast3R} & \multicolumn{2}{c|}{VGGT} & \multicolumn{2}{c}{Refined VGGT} \\
\textbf{Scene} & PSNR & Time & PSNR & Time & PSNR & Time & PSNR & Time & PSNR & Time & PSNR & Time & PSNR & Time & PSNR & Time & PSNR & Time & PSNR & Time & PSNR & Time & PSNR & Time & PSNR & Time \\
\midrule
bicycle & 24.02 & 1.13 & 24.73 & 1.39 & 25.37 & 2.31 & \cellcolor{tabthird}25.59 & 3.95 & \cellcolor{tabsecond}25.62 & 9.33 & 18.54 & \cellcolor{tabsecond}0.42 & \cellcolor{tabfirst}25.84 & 2.83 & 14.96 & \cellcolor{tabthird}0.60 & - & 5.71 & 14.85 & \cellcolor{tabfirst}0.28 & 14.11 & 5.86 & OOM & OOM & OOM & OOM \\
bonsai & 31.16 & 1.80 & 31.59 & 3.32 & 24.96 & 11.51 & \cellcolor{tabsecond}32.42 & 12.24 & \cellcolor{tabthird}32.40 & 24.91 & 21.10 & \cellcolor{tabsecond}0.69 & - & 14.45 & 18.66 & \cellcolor{tabthird}1.11 & \cellcolor{tabfirst}32.76 & 12.75 & 15.54 & \cellcolor{tabfirst}0.42 & 20.68 & 22.23 & OOM & OOM & OOM & OOM \\
counter & 28.70 & 1.16 & 28.94 & 1.55 & 29.09 & 3.77 & 28.89 & 7.32 & \cellcolor{tabthird}29.16 & 15.21 & 22.89 & \cellcolor{tabsecond}0.48 & \cellcolor{tabfirst}29.21 & 5.86 & 16.98 & \cellcolor{tabthird}0.77 & \cellcolor{tabsecond}29.20 & 8.41 & 16.08 & \cellcolor{tabfirst}0.29 & 20.82 & 10.94 & OOM & OOM & OOM & OOM \\
flowers & 1.12 & 0.61 & 10.30 & 1.03 & 21.16 & 1.84 & \cellcolor{tabthird}21.27 & 2.69 & \cellcolor{tabsecond}21.68 & 7.85 & 16.86 & \cellcolor{tabsecond}0.28 & \cellcolor{tabfirst}22.25 & 1.63 & 11.51 & \cellcolor{tabthird}0.43 & 18.28 & 3.57 & 10.09 & \cellcolor{tabfirst}0.16 & 11.93 & 1.31 & OOM & OOM & OOM & OOM \\
garden & 27.02 & 1.50 & 27.36 & 2.11 & 27.61 & 3.37 & 27.71 & 6.85 & \cellcolor{tabthird}27.71 & 10.01 & 19.45 & \cellcolor{tabsecond}0.29 & \cellcolor{tabfirst}28.18 & 4.43 & 17.32 & \cellcolor{tabthird}0.48 & \cellcolor{tabsecond}28.10 & 5.93 & 15.80 & \cellcolor{tabfirst}0.17 & 16.65 & 6.95 & OOM & OOM & OOM & OOM \\
kitchen & 31.17 & 1.99 & 31.56 & 3.58 & \cellcolor{tabthird}31.56 & 6.61 & 31.36 & 11.93 & 31.37 & 26.78 & 21.52 & \cellcolor{tabsecond}0.63 & \cellcolor{tabfirst}31.81 & 9.27 & 18.52 & \cellcolor{tabthird}1.02 & \cellcolor{tabsecond}31.74 & 14.93 & 17.65 & \cellcolor{tabfirst}0.38 & 27.10 & 16.31 & OOM & OOM & OOM & OOM \\
room & 30.43 & \cellcolor{tabthird}2.10 & \cellcolor{tabthird}31.42 & 3.43 & 31.40 & 6.09 & \cellcolor{tabfirst}31.65 & 11.45 & \cellcolor{tabsecond}31.48 & 21.59 & OOM & OOM & OOM & OOM & 17.42 & \cellcolor{tabsecond}1.25 & 30.60 & 21.25 & 17.46 & \cellcolor{tabfirst}0.47 & 21.80 & 20.22 & OOM & OOM & OOM & OOM \\
stump & - & 0.42 & 16.02 & 0.75 & 26.60 & 0.98 & 26.85 & 1.75 & 26.83 & 4.63 & 19.06 & \cellcolor{tabsecond}0.16 & \cellcolor{tabthird}27.01 & 1.23 & 17.17 & 0.24 & \cellcolor{tabfirst}27.11 & 1.08 & 15.68 & \cellcolor{tabfirst}0.09 & - & 0.09 & 19.98 & \cellcolor{tabthird}0.20 & \cellcolor{tabsecond}27.07 & 0.56 \\
treehill & 3.14 & 0.40 & 21.82 & 0.82 & 22.15 & 1.27 & \cellcolor{tabthird}22.33 & 2.41 & \cellcolor{tabsecond}22.38 & 6.18 & 18.08 & \cellcolor{tabsecond}0.19 & - & 3.35 & 11.69 & \cellcolor{tabthird}0.30 & \cellcolor{tabfirst}24.08 & 2.92 & 12.39 & \cellcolor{tabfirst}0.11 & 15.79 & 5.37 & OOM & OOM & OOM & OOM \\
\bottomrule
\end{tabular}
}
\caption{Rendering and time results of the feed-forward refinements using \textbf{8192} feats on the MipNeRF 360 dataset. Times are in minutes.}
\label{tab:ffn_refine_Mip-NeRF360_8192feats}
\end{table*}

\begin{table*}%[]
\centering
\resizebox{\textwidth}{!}{
\begin{tabular}{lcc|cc|cc|cc|cc|cc|cc|cc|cc|cc|cc|cc|cc}
\toprule
& \multicolumn{10}{c|}{\textbf{Standard COLMAP reconstruction}} & \multicolumn{16}{c}{\textbf{Feed-forward recons}} \\
 & \multicolumn{2}{c|}{512} & \multicolumn{2}{c|}{1024} & \multicolumn{2}{c|}{2048} & \multicolumn{2}{c|}{4096} & \multicolumn{2}{c|}{8192} & \multicolumn{2}{c|}{DA3} & \multicolumn{2}{c|}{Refined DA3} & \multicolumn{2}{c|}{$\pi^3$} & \multicolumn{2}{c|}{Refined $\pi^3$} & \multicolumn{2}{c|}{Fast3R} & \multicolumn{2}{c|}{Refined Fast3R} & \multicolumn{2}{c|}{VGGT} & \multicolumn{2}{c}{Refined VGGT} \\
\textbf{Scene} & PSNR & Time & PSNR & Time & PSNR & Time & PSNR & Time & PSNR & Time & PSNR & Time & PSNR & Time & PSNR & Time & PSNR & Time & PSNR & Time & PSNR & Time & PSNR & Time & PSNR & Time \\
\midrule
bicycle & 24.02 & 1.13 & 24.73 & 1.39 & 25.37 & 2.31 & \cellcolor{tabthird}25.59 & 3.95 & \cellcolor{tabfirst}25.62 & 9.33 & 18.54 & \cellcolor{tabsecond}0.42 & \cellcolor{tabsecond}25.62 & 2.57 & 14.96 & \cellcolor{tabthird}0.60 & 18.38 & 3.54 & 14.85 & \cellcolor{tabfirst}0.28 & 13.00 & 3.47 & OOM & OOM & OOM & OOM \\
bonsai & 31.16 & 1.80 & 31.59 & 3.32 & 24.96 & 11.51 & \cellcolor{tabsecond}32.42 & 12.24 & \cellcolor{tabthird}32.40 & 24.91 & 21.10 & \cellcolor{tabsecond}0.69 & - & 8.41 & 18.66 & \cellcolor{tabthird}1.11 & \cellcolor{tabfirst}32.73 & 8.36 & 15.54 & \cellcolor{tabfirst}0.42 & 16.73 & 15.87 & OOM & OOM & OOM & OOM \\
counter & 28.70 & 1.16 & 28.94 & 1.55 & 29.09 & 3.77 & 28.89 & 7.32 & \cellcolor{tabthird}29.16 & 15.21 & 22.89 & \cellcolor{tabsecond}0.48 & \cellcolor{tabsecond}29.20 & 3.82 & 16.98 & \cellcolor{tabthird}0.77 & \cellcolor{tabfirst}29.20 & 4.68 & 16.08 & \cellcolor{tabfirst}0.29 & - & 0.29 & OOM & OOM & OOM & OOM \\
flowers & 1.12 & 0.61 & 10.30 & 1.03 & 21.16 & 1.84 & \cellcolor{tabthird}21.27 & 2.69 & \cellcolor{tabsecond}21.68 & 7.85 & 16.86 & \cellcolor{tabsecond}0.28 & \cellcolor{tabfirst}22.07 & 1.40 & 11.51 & \cellcolor{tabthird}0.43 & 16.93 & 2.26 & 10.09 & \cellcolor{tabfirst}0.16 & - & 0.16 & OOM & OOM & OOM & OOM \\
garden & 27.02 & 1.50 & 27.36 & 2.11 & 27.61 & 3.37 & 27.71 & 6.85 & \cellcolor{tabthird}27.71 & 10.01 & 19.45 & \cellcolor{tabsecond}0.29 & \cellcolor{tabfirst}28.11 & 3.13 & 17.32 & \cellcolor{tabthird}0.48 & \cellcolor{tabsecond}28.06 & 4.09 & 15.80 & \cellcolor{tabfirst}0.17 & 16.60 & 3.66 & OOM & OOM & OOM & OOM \\
kitchen & 31.17 & 1.99 & \cellcolor{tabthird}31.56 & 3.58 & \cellcolor{tabsecond}31.56 & 6.61 & 31.36 & 11.93 & 31.37 & 26.78 & 21.52 & \cellcolor{tabsecond}0.63 & \cellcolor{tabfirst}31.65 & 6.41 & 18.52 & \cellcolor{tabthird}1.02 & 31.50 & 8.36 & 17.65 & \cellcolor{tabfirst}0.38 & 27.00 & 14.81 & OOM & OOM & OOM & OOM \\
room & 30.43 & \cellcolor{tabthird}2.10 & \cellcolor{tabthird}31.42 & 3.43 & 31.40 & 6.09 & \cellcolor{tabfirst}31.65 & 11.45 & \cellcolor{tabsecond}31.48 & 21.59 & OOM & OOM & OOM & OOM & 17.42 & \cellcolor{tabsecond}1.25 & 28.15 & 14.30 & 17.46 & \cellcolor{tabfirst}0.47 & 21.32 & 17.51 & OOM & OOM & OOM & OOM \\
stump & - & 0.42 & 16.02 & 0.75 & 26.60 & 0.98 & 26.85 & 1.75 & 26.83 & 4.63 & 19.06 & \cellcolor{tabsecond}0.16 & \cellcolor{tabsecond}26.99 & 0.88 & 17.17 & 0.24 & \cellcolor{tabfirst}27.02 & 1.17 & 15.68 & \cellcolor{tabfirst}0.09 & - & 0.09 & 19.98 & \cellcolor{tabthird}0.20 & \cellcolor{tabthird}26.93 & 0.66 \\
treehill & 3.14 & 0.40 & 21.82 & 0.82 & 22.15 & 1.27 & \cellcolor{tabthird}22.33 & 2.41 & \cellcolor{tabsecond}22.38 & 6.18 & 18.08 & \cellcolor{tabsecond}0.19 & - & 1.94 & 11.69 & \cellcolor{tabthird}0.30 & \cellcolor{tabfirst}23.91 & 1.85 & 12.39 & \cellcolor{tabfirst}0.11 & 15.74 & 2.87 & OOM & OOM & OOM & OOM \\
\bottomrule
\end{tabular}
}
\caption{Rendering and time results of the feed-forward refinements using \textbf{4096} feats on the MipNeRF 360 dataset. Times are in minutes.}
\label{tab:ffn_refine_Mip-NeRF360_4096feats}
\end{table*}

\begin{table*}%[]
\centering
\resizebox{\textwidth}{!}{
\begin{tabular}{lcc|cc|cc|cc|cc|cc|cc|cc|cc|cc|cc|cc|cc}
\toprule
& \multicolumn{10}{c|}{\textbf{Standard COLMAP reconstruction}} & \multicolumn{16}{c}{\textbf{Feed-forward recons}} \\
 & \multicolumn{2}{c|}{512} & \multicolumn{2}{c|}{1024} & \multicolumn{2}{c|}{2048} & \multicolumn{2}{c|}{4096} & \multicolumn{2}{c|}{8192} & \multicolumn{2}{c|}{DA3} & \multicolumn{2}{c|}{Refined DA3} & \multicolumn{2}{c|}{$\pi^3$} & \multicolumn{2}{c|}{Refined $\pi^3$} & \multicolumn{2}{c|}{Fast3R} & \multicolumn{2}{c|}{Refined Fast3R} & \multicolumn{2}{c|}{VGGT} & \multicolumn{2}{c}{Refined VGGT} \\
\textbf{Scene} & PSNR & Time & PSNR & Time & PSNR & Time & PSNR & Time & PSNR & Time & PSNR & Time & PSNR & Time & PSNR & Time & PSNR & Time & PSNR & Time & PSNR & Time & PSNR & Time & PSNR & Time \\
\midrule
bicycle & 24.02 & 1.13 & 24.73 & 1.39 & \cellcolor{tabthird}25.37 & 2.31 & \cellcolor{tabsecond}25.59 & 3.95 & \cellcolor{tabfirst}25.62 & 9.33 & 18.54 & \cellcolor{tabsecond}0.42 & 25.24 & 1.80 & 14.96 & \cellcolor{tabthird}0.60 & 24.82 & 2.63 & 14.85 & \cellcolor{tabfirst}0.28 & 14.87 & 2.52 & OOM & OOM & OOM & OOM \\
bonsai & 31.16 & 1.80 & 31.59 & 3.32 & 24.96 & 11.51 & \cellcolor{tabfirst}32.42 & 12.24 & \cellcolor{tabsecond}32.40 & 24.91 & 21.10 & \cellcolor{tabsecond}0.69 & - & 5.87 & 18.66 & \cellcolor{tabthird}1.11 & \cellcolor{tabthird}32.15 & 5.29 & 15.54 & \cellcolor{tabfirst}0.42 & 18.20 & 6.85 & OOM & OOM & OOM & OOM \\
counter & 28.70 & 1.16 & 28.94 & 1.55 & \cellcolor{tabsecond}29.09 & 3.77 & 28.89 & 7.32 & \cellcolor{tabfirst}29.16 & 15.21 & 22.89 & \cellcolor{tabsecond}0.48 & 29.03 & 2.36 & 16.98 & \cellcolor{tabthird}0.77 & \cellcolor{tabthird}29.04 & 3.83 & 16.08 & \cellcolor{tabfirst}0.29 & - & 0.29 & OOM & OOM & OOM & OOM \\
flowers & 1.12 & 0.61 & 10.30 & 1.03 & 21.16 & 1.84 & \cellcolor{tabthird}21.27 & 2.69 & \cellcolor{tabsecond}21.68 & 7.85 & 16.86 & \cellcolor{tabsecond}0.28 & \cellcolor{tabfirst}21.71 & 1.04 & 11.51 & \cellcolor{tabthird}0.43 & 17.45 & 1.63 & 10.09 & \cellcolor{tabfirst}0.16 & - & 0.16 & OOM & OOM & OOM & OOM \\
garden & 27.02 & 1.50 & 27.36 & 2.11 & 27.61 & 3.37 & 27.71 & 6.85 & \cellcolor{tabthird}27.71 & 10.01 & 19.45 & \cellcolor{tabsecond}0.29 & \cellcolor{tabfirst}27.92 & 1.72 & 17.32 & \cellcolor{tabthird}0.48 & \cellcolor{tabsecond}27.79 & 2.18 & 15.80 & \cellcolor{tabfirst}0.17 & 16.62 & 1.72 & OOM & OOM & OOM & OOM \\
kitchen & 31.17 & 1.99 & \cellcolor{tabthird}31.56 & 3.58 & \cellcolor{tabsecond}31.56 & 6.61 & 31.36 & 11.93 & 31.37 & 26.78 & 21.52 & \cellcolor{tabsecond}0.63 & \cellcolor{tabfirst}31.65 & 4.28 & 18.52 & \cellcolor{tabthird}1.02 & 31.52 & 4.44 & 17.65 & \cellcolor{tabfirst}0.38 & 26.99 & 8.09 & OOM & OOM & OOM & OOM \\
room & 30.43 & \cellcolor{tabthird}2.10 & \cellcolor{tabthird}31.42 & 3.43 & 31.40 & 6.09 & \cellcolor{tabfirst}31.65 & 11.45 & \cellcolor{tabsecond}31.48 & 21.59 & OOM & OOM & OOM & OOM & 17.42 & \cellcolor{tabsecond}1.25 & 29.24 & 7.78 & 17.46 & \cellcolor{tabfirst}0.47 & 21.17 & 7.32 & OOM & OOM & OOM & OOM \\
stump & - & 0.42 & 16.02 & 0.75 & 26.60 & 0.98 & \cellcolor{tabfirst}26.85 & 1.75 & \cellcolor{tabsecond}26.83 & 4.63 & 19.06 & \cellcolor{tabsecond}0.16 & \cellcolor{tabthird}26.69 & 0.76 & 17.17 & 0.24 & 26.61 & 0.87 & 15.68 & \cellcolor{tabfirst}0.09 & - & 0.09 & 19.98 & \cellcolor{tabthird}0.20 & 26.62 & 0.47 \\
treehill & 3.14 & 0.40 & 21.82 & 0.82 & 22.15 & 1.27 & \cellcolor{tabthird}22.33 & 2.41 & \cellcolor{tabsecond}22.38 & 6.18 & 18.08 & \cellcolor{tabsecond}0.19 & - & 1.38 & 11.69 & \cellcolor{tabthird}0.30 & \cellcolor{tabfirst}23.85 & 1.53 & 12.39 & \cellcolor{tabfirst}0.11 & 16.46 & 1.66 & OOM & OOM & OOM & OOM \\
\bottomrule
\end{tabular}
}
\caption{Rendering and time results of the feed-forward refinements using \textbf{2048} feats on the MipNeRF 360 dataset. Times are in minutes.}
\label{tab:ffn_refine_Mip-NeRF360_2048feats}
\end{table*}

In Tables~\ref{tab:ffn_refine_Tanks_Temples_8192feats}, \ref{tab:ffn_refine_Tanks_Temples_4096feats}, \ref{tab:ffn_refine_Tanks_Temples_2048feats} we report the PSNR and reconstruction time results for 8192, 4096, and 2048 features, respectively, on the Tanks and Temples dataset. As discussed in the main paper, all of the scenes run out of memory (OOM) for VGGT except for 2 of them. In the same way as with the Mip-NeRF 360 dataset, we can see that the PSNR metrics of the refined 3D models improve with the number of features, but timings are also more affected than in the Mip-NeRF 360 dataset case, where for 8192 features the final times are comparable or longer than the ones obtained with Standard COLMAP with 8192 keypoints. However, for 2048 features (Table~\ref{tab:ffn_refine_Tanks_Temples_2048feats}), we can see that the timings are better for many scenes, but the performance is not as good as with 8192 features for the refinement.

\begin{table*}%[]
\centering
\resizebox{\textwidth}{!}{
\begin{tabular}{lcc|cc|cc|cc|cc|cc|cc|cc|cc|cc|cc|cc|cc}
\toprule
& \multicolumn{10}{c|}{\textbf{Standard COLMAP reconstruction}} & \multicolumn{16}{c}{\textbf{Feed-forward recons}} \\
 & \multicolumn{2}{c|}{512} & \multicolumn{2}{c|}{1024} & \multicolumn{2}{c|}{2048} & \multicolumn{2}{c|}{4096} & \multicolumn{2}{c|}{8192} & \multicolumn{2}{c|}{DA3} & \multicolumn{2}{c|}{Refined DA3} & \multicolumn{2}{c|}{$\pi^3$} & \multicolumn{2}{c|}{Refined $\pi^3$} & \multicolumn{2}{c|}{Fast3R} & \multicolumn{2}{c|}{Refined Fast3R} & \multicolumn{2}{c|}{VGGT} & \multicolumn{2}{c}{Refined VGGT} \\
\textbf{Scene} & PSNR & Time & PSNR & Time & PSNR & Time & PSNR & Time & PSNR & Time & PSNR & Time & PSNR & Time & PSNR & Time & PSNR & Time & PSNR & Time & PSNR & Time & PSNR & Time & PSNR & Time \\
\midrule
auditorium & 23.03 & 2.67 & \cellcolor{tabthird}23.28 & 2.18 & \cellcolor{tabfirst}23.32 & 4.64 & 23.27 & 8.98 & \cellcolor{tabsecond}23.31 & 6.06 & 21.47 & \cellcolor{tabsecond}0.52 & - & 7.14 & 16.55 & \cellcolor{tabthird}1.21 & 22.65 & 5.54 & 17.90 & \cellcolor{tabfirst}0.30 & 16.61 & 6.53 & OOM & OOM & OOM & OOM \\
ballroom & \cellcolor{tabsecond}21.88 & 2.39 & 19.69 & 5.56 & \cellcolor{tabthird}19.78 & 9.46 & \cellcolor{tabfirst}23.11 & 18.60 & 17.13 & 18.28 & 13.08 & \cellcolor{tabsecond}0.60 & 15.67 & 17.26 & 12.43 & \cellcolor{tabthird}1.40 & 16.15 & 19.36 & 11.97 & \cellcolor{tabfirst}0.34 & 13.17 & 17.85 & OOM & OOM & OOM & OOM \\
barn & 27.44 & \cellcolor{tabsecond}2.58 & 27.66 & \cellcolor{tabthird}4.44 & \cellcolor{tabthird}27.86 & 8.91 & \cellcolor{tabfirst}27.96 & 13.51 & \cellcolor{tabsecond}27.95 & 13.56 & OOM & OOM & OOM & OOM & OOM & OOM & OOM & OOM & 10.28 & \cellcolor{tabfirst}0.53 & - & 0.53 & OOM & OOM & OOM & OOM \\
caterpillar & 23.11 & \cellcolor{tabthird}2.47 & 23.04 & 3.78 & 23.31 & 7.60 & \cellcolor{tabthird}23.38 & 12.76 & \cellcolor{tabsecond}23.39 & 16.03 & OOM & OOM & OOM & OOM & 15.04 & \cellcolor{tabsecond}1.91 & \cellcolor{tabfirst}23.54 & 13.20 & 12.29 & \cellcolor{tabfirst}0.47 & 13.04 & 18.20 & OOM & OOM & OOM & OOM \\
church & 20.75 & \cellcolor{tabsecond}9.61 & 20.70 & \cellcolor{tabthird}16.47 & \cellcolor{tabthird}21.57 & 29.77 & \cellcolor{tabsecond}21.57 & 38.93 & \cellcolor{tabfirst}21.85 & 68.89 & OOM & OOM & OOM & OOM & OOM & OOM & OOM & OOM & 12.52 & \cellcolor{tabfirst}0.86 & 13.54 & 17.26 & OOM & OOM & OOM & OOM \\
courthouse & \cellcolor{tabthird}21.06 & \cellcolor{tabfirst}17.72 & \cellcolor{tabsecond}21.15 & \cellcolor{tabsecond}28.72 & \cellcolor{tabfirst}21.41 & \cellcolor{tabthird}64.73 & 21.02 & 86.76 & 20.81 & 95.86 & OOM & OOM & OOM & OOM & OOM & OOM & OOM & OOM & OOM & OOM & OOM & OOM & OOM & OOM & OOM & OOM \\
courtroom & 22.07 & 1.92 & \cellcolor{tabthird}22.10 & 3.03 & \cellcolor{tabsecond}22.15 & 4.82 & 20.85 & 10.13 & \cellcolor{tabfirst}22.30 & 9.81 & 17.53 & \cellcolor{tabsecond}0.52 & 19.82 & 11.34 & 13.94 & \cellcolor{tabthird}1.22 & 20.51 & 12.05 & 14.80 & \cellcolor{tabfirst}0.30 & 16.06 & 9.27 & OOM & OOM & OOM & OOM \\
family & 24.10 & 0.46 & \cellcolor{tabsecond}24.39 & 0.73 & \cellcolor{tabthird}24.38 & 1.36 & 24.33 & 3.27 & 24.34 & 3.05 & 19.25 & \cellcolor{tabsecond}0.17 & 23.43 & 3.61 & 13.98 & 0.35 & \cellcolor{tabfirst}24.87 & 3.69 & 13.51 & \cellcolor{tabfirst}0.09 & - & 8.14 & 19.95 & \cellcolor{tabthird}0.19 & 23.41 & 2.25 \\
francis & 26.74 & 1.40 & 25.87 & 4.71 & \cellcolor{tabthird}27.12 & 3.98 & 26.99 & 6.92 & \cellcolor{tabsecond}27.15 & 6.98 & 22.34 & \cellcolor{tabsecond}0.52 & 26.85 & 13.72 & 17.38 & \cellcolor{tabthird}1.22 & \cellcolor{tabfirst}27.80 & 11.05 & 14.72 & \cellcolor{tabfirst}0.30 & 16.04 & 12.61 & OOM & OOM & OOM & OOM \\
horse & 22.00 & 0.35 & \cellcolor{tabsecond}23.65 & 0.54 & \cellcolor{tabthird}23.56 & 1.05 & 23.49 & 1.46 & 23.36 & 1.39 & 16.49 & \cellcolor{tabsecond}0.16 & 17.98 & 2.39 & 13.44 & 0.35 & \cellcolor{tabfirst}24.25 & 2.14 & 12.85 & \cellcolor{tabfirst}0.09 & 14.87 & 2.34 & 17.45 & \cellcolor{tabthird}0.19 & 19.54 & 1.35 \\
ignatius & 20.30 & 1.13 & \cellcolor{tabsecond}21.16 & 2.09 & \cellcolor{tabthird}21.05 & 3.24 & 19.27 & 8.69 & 20.29 & 13.87 & 17.20 & \cellcolor{tabsecond}0.41 & 20.38 & 7.35 & 12.95 & \cellcolor{tabthird}0.95 & \cellcolor{tabfirst}22.01 & 14.58 & 13.20 & \cellcolor{tabfirst}0.23 & 14.12 & 12.13 & OOM & OOM & OOM & OOM \\
lighthouse & 19.10 & \cellcolor{tabthird}1.03 & 21.31 & 2.54 & \cellcolor{tabthird}21.49 & 4.12 & \cellcolor{tabfirst}21.75 & 6.42 & 21.35 & 13.18 & 17.73 & \cellcolor{tabsecond}0.45 & - & 16.32 & 13.06 & 1.24 & \cellcolor{tabsecond}21.65 & 14.94 & 14.55 & \cellcolor{tabfirst}0.28 & 16.18 & 14.83 & OOM & OOM & OOM & OOM \\
m60 & 25.93 & \cellcolor{tabthird}1.03 & 27.61 & 1.99 & 27.96 & 4.08 & \cellcolor{tabsecond}28.10 & 7.21 & \cellcolor{tabthird}28.08 & 6.21 & 17.97 & \cellcolor{tabsecond}0.46 & \cellcolor{tabfirst}28.49 & 8.72 & 13.28 & 1.27 & - & 17.48 & 12.96 & \cellcolor{tabfirst}0.29 & 14.49 & 13.48 & OOM & OOM & OOM & OOM \\
meetingroom & 23.76 & \cellcolor{tabsecond}1.25 & 24.91 & 2.12 & \cellcolor{tabfirst}25.12 & 3.54 & \cellcolor{tabthird}25.07 & 3.77 & \cellcolor{tabsecond}25.09 & 3.97 & OOM & OOM & OOM & OOM & 14.28 & \cellcolor{tabthird}1.80 & 24.40 & 7.54 & 13.78 & \cellcolor{tabfirst}0.44 & 11.64 & 8.66 & OOM & OOM & OOM & OOM \\
museum & 19.50 & \cellcolor{tabsecond}1.15 & \cellcolor{tabsecond}20.19 & 3.70 & \cellcolor{tabfirst}20.20 & 3.81 & 20.02 & 6.79 & \cellcolor{tabthird}20.17 & 7.34 & OOM & OOM & OOM & OOM & 12.14 & \cellcolor{tabthird}1.21 & 18.46 & 10.37 & 13.38 & \cellcolor{tabfirst}0.30 & 14.00 & 14.30 & OOM & OOM & OOM & OOM \\
palace & 13.40 & \cellcolor{tabsecond}1.68 & \cellcolor{tabthird}17.37 & \cellcolor{tabthird}4.03 & \cellcolor{tabfirst}17.98 & 8.34 & \cellcolor{tabsecond}17.64 & 14.73 & 17.33 & 13.24 & OOM & OOM & OOM & OOM & OOM & OOM & OOM & OOM & 10.48 & \cellcolor{tabfirst}0.85 & - & 0.85 & OOM & OOM & OOM & OOM \\
panther & 28.12 & 1.28 & 27.82 & 2.70 & 28.45 & 4.94 & 28.59 & 10.79 & \cellcolor{tabthird}28.61 & 10.65 & 19.43 & \cellcolor{tabsecond}0.46 & \cellcolor{tabfirst}28.99 & 8.86 & 14.48 & \cellcolor{tabthird}1.28 & \cellcolor{tabsecond}28.98 & 10.17 & 15.62 & \cellcolor{tabfirst}0.29 & 17.70 & 10.45 & OOM & OOM & OOM & OOM \\
playground & 25.09 & \cellcolor{tabthird}1.13 & 25.37 & 1.78 & 25.45 & 4.51 & \cellcolor{tabsecond}25.89 & 5.34 & 25.53 & 7.69 & 19.90 & \cellcolor{tabsecond}0.53 & \cellcolor{tabthird}25.57 & 6.67 & 16.02 & 1.22 & \cellcolor{tabfirst}26.07 & 7.62 & 14.69 & \cellcolor{tabfirst}0.31 & 17.26 & 9.52 & OOM & OOM & OOM & OOM \\
temple & 12.10 & 1.44 & \cellcolor{tabthird}16.02 & 1.55 & \cellcolor{tabfirst}19.01 & 2.48 & 15.43 & 4.19 & 15.39 & 3.88 & 15.21 & \cellcolor{tabsecond}0.52 & - & 7.58 & 12.94 & \cellcolor{tabthird}1.22 & \cellcolor{tabsecond}18.47 & 7.17 & 12.88 & \cellcolor{tabfirst}0.30 & 13.84 & 8.09 & OOM & OOM & OOM & OOM \\
train & 20.37 & \cellcolor{tabthird}1.16 & \cellcolor{tabthird}21.32 & 1.96 & \cellcolor{tabsecond}21.37 & 3.64 & \cellcolor{tabfirst}21.47 & 8.03 & 21.32 & 7.78 & 16.46 & \cellcolor{tabsecond}0.52 & 21.01 & 6.06 & 13.45 & 1.21 & 21.18 & 7.95 & 11.05 & \cellcolor{tabfirst}0.30 & 11.81 & 11.29 & OOM & OOM & OOM & OOM \\
truck & 12.35 & 1.65 & 24.14 & 1.39 & 24.21 & 2.51 & 24.31 & 4.69 & \cellcolor{tabthird}24.35 & 5.50 & 16.55 & \cellcolor{tabsecond}0.44 & \cellcolor{tabfirst}24.44 & 5.50 & 14.56 & \cellcolor{tabthird}0.86 & \cellcolor{tabsecond}24.42 & 4.43 & 11.45 & \cellcolor{tabfirst}0.22 & 12.05 & 7.98 & OOM & OOM & OOM & OOM \\
\bottomrule
\end{tabular}
}
\caption{Rendering and time results of the feed-forward refinements using \textbf{8192} feats on the Tanks \& Temples. Times are in minutes.}
\label{tab:ffn_refine_Tanks_Temples_8192feats}
\end{table*}

\begin{table*}%[]
\centering
\resizebox{\textwidth}{!}{
\begin{tabular}{lcc|cc|cc|cc|cc|cc|cc|cc|cc|cc|cc|cc|cc}
\toprule
& \multicolumn{10}{c|}{\textbf{Standard COLMAP reconstruction}} & \multicolumn{16}{c}{\textbf{Feed-forward recons}} \\
 & \multicolumn{2}{c|}{512} & \multicolumn{2}{c|}{1024} & \multicolumn{2}{c|}{2048} & \multicolumn{2}{c|}{4096} & \multicolumn{2}{c|}{8192} & \multicolumn{2}{c|}{DA3} & \multicolumn{2}{c|}{Refined DA3} & \multicolumn{2}{c|}{$\pi^3$} & \multicolumn{2}{c|}{Refined $\pi^3$} & \multicolumn{2}{c|}{Fast3R} & \multicolumn{2}{c|}{Refined Fast3R} & \multicolumn{2}{c|}{VGGT} & \multicolumn{2}{c}{Refined VGGT} \\
\textbf{Scene} & PSNR & Time & PSNR & Time & PSNR & Time & PSNR & Time & PSNR & Time & PSNR & Time & PSNR & Time & PSNR & Time & PSNR & Time & PSNR & Time & PSNR & Time & PSNR & Time & PSNR & Time \\
\midrule
auditorium & 23.03 & 2.67 & \cellcolor{tabthird}23.28 & 2.18 & \cellcolor{tabfirst}23.32 & 4.64 & 23.27 & 8.98 & \cellcolor{tabsecond}23.31 & 6.06 & 21.47 & \cellcolor{tabsecond}0.52 & - & 6.84 & 16.55 & \cellcolor{tabthird}1.21 & 22.16 & 6.25 & 17.90 & \cellcolor{tabfirst}0.30 & 14.52 & 6.60 & OOM & OOM & OOM & OOM \\
ballroom & \cellcolor{tabsecond}21.88 & 2.39 & 19.69 & 5.56 & \cellcolor{tabthird}19.78 & 9.46 & \cellcolor{tabfirst}23.11 & 18.60 & 17.13 & 18.28 & 13.08 & \cellcolor{tabsecond}0.60 & 15.60 & 15.12 & 12.43 & \cellcolor{tabthird}1.40 & 16.14 & 15.48 & 11.97 & \cellcolor{tabfirst}0.34 & 13.39 & 15.19 & OOM & OOM & OOM & OOM \\
barn & 27.44 & \cellcolor{tabsecond}2.58 & 27.66 & \cellcolor{tabthird}4.44 & \cellcolor{tabthird}27.86 & 8.91 & \cellcolor{tabfirst}27.96 & 13.51 & \cellcolor{tabsecond}27.95 & 13.56 & OOM & OOM & OOM & OOM & OOM & OOM & OOM & OOM & 10.28 & \cellcolor{tabfirst}0.53 & - & 0.53 & OOM & OOM & OOM & OOM \\
caterpillar & 23.11 & \cellcolor{tabthird}2.47 & 23.04 & 3.78 & 23.31 & 7.60 & \cellcolor{tabsecond}23.38 & 12.76 & \cellcolor{tabfirst}23.39 & 16.03 & OOM & OOM & OOM & OOM & 15.04 & \cellcolor{tabsecond}1.91 & \cellcolor{tabthird}23.37 & 18.20 & 12.29 & \cellcolor{tabfirst}0.47 & 13.15 & 13.35 & OOM & OOM & OOM & OOM \\
church & 20.75 & \cellcolor{tabsecond}9.61 & 20.70 & \cellcolor{tabthird}16.47 & \cellcolor{tabthird}21.57 & 29.77 & \cellcolor{tabsecond}21.57 & 38.93 & \cellcolor{tabfirst}21.85 & 68.89 & OOM & OOM & OOM & OOM & OOM & OOM & OOM & OOM & 12.52 & \cellcolor{tabfirst}0.86 & 13.34 & 20.25 & OOM & OOM & OOM & OOM \\
courthouse & \cellcolor{tabthird}21.06 & \cellcolor{tabfirst}17.72 & \cellcolor{tabsecond}21.15 & \cellcolor{tabsecond}28.72 & \cellcolor{tabfirst}21.41 & \cellcolor{tabthird}64.73 & 21.02 & 86.76 & 20.81 & 95.86 & OOM & OOM & OOM & OOM & OOM & OOM & OOM & OOM & OOM & OOM & OOM & OOM & OOM & OOM & OOM & OOM \\
courtroom & 22.07 & 1.92 & \cellcolor{tabthird}22.10 & 3.03 & \cellcolor{tabsecond}22.15 & 4.82 & 20.85 & 10.13 & \cellcolor{tabfirst}22.30 & 9.81 & 17.53 & \cellcolor{tabsecond}0.52 & 19.87 & 11.64 & 13.94 & \cellcolor{tabthird}1.22 & 20.75 & 8.82 & 14.80 & \cellcolor{tabfirst}0.30 & 16.15 & 9.44 & OOM & OOM & OOM & OOM \\
family & 24.10 & 0.46 & \cellcolor{tabsecond}24.39 & 0.73 & \cellcolor{tabthird}24.38 & 1.36 & 24.33 & 3.27 & 24.34 & 3.05 & 19.25 & \cellcolor{tabsecond}0.17 & 23.42 & 3.15 & 13.98 & 0.35 & \cellcolor{tabfirst}24.81 & 5.64 & 13.51 & \cellcolor{tabfirst}0.09 & 24.04 & 4.62 & 19.95 & \cellcolor{tabthird}0.19 & 23.42 & 2.07 \\
francis & 26.74 & 1.40 & 25.87 & 4.71 & \cellcolor{tabthird}27.12 & 3.98 & 26.99 & 6.92 & \cellcolor{tabsecond}27.15 & 6.98 & 22.34 & \cellcolor{tabsecond}0.52 & 27.04 & 13.61 & 17.38 & \cellcolor{tabthird}1.22 & \cellcolor{tabfirst}28.21 & 11.05 & 14.72 & \cellcolor{tabfirst}0.30 & 15.90 & 11.12 & OOM & OOM & OOM & OOM \\
horse & 22.00 & 0.35 & \cellcolor{tabsecond}23.65 & 0.54 & \cellcolor{tabthird}23.56 & 1.05 & 23.49 & 1.46 & 23.36 & 1.39 & 16.49 & \cellcolor{tabsecond}0.16 & 18.03 & 2.27 & 13.44 & 0.35 & \cellcolor{tabfirst}24.36 & 1.85 & 12.85 & \cellcolor{tabfirst}0.09 & 14.87 & 2.56 & 17.45 & \cellcolor{tabthird}0.19 & 19.49 & 1.42 \\
ignatius & 20.30 & 1.13 & \cellcolor{tabsecond}21.16 & 2.09 & \cellcolor{tabthird}21.05 & 3.24 & 19.27 & 8.69 & 20.29 & 13.87 & 17.20 & \cellcolor{tabsecond}0.41 & 20.41 & 5.46 & 12.95 & \cellcolor{tabthird}0.95 & \cellcolor{tabfirst}21.96 & 8.31 & 13.20 & \cellcolor{tabfirst}0.23 & 14.15 & 7.20 & OOM & OOM & OOM & OOM \\
lighthouse & 19.10 & \cellcolor{tabthird}1.03 & 21.31 & 2.54 & \cellcolor{tabthird}21.49 & 4.12 & \cellcolor{tabsecond}21.75 & 6.42 & 21.35 & 13.18 & 17.73 & \cellcolor{tabsecond}0.45 & - & 12.90 & 13.06 & 1.24 & \cellcolor{tabfirst}22.31 & 17.19 & 14.55 & \cellcolor{tabfirst}0.28 & 16.10 & 18.64 & OOM & OOM & OOM & OOM \\
m60 & 25.93 & \cellcolor{tabthird}1.03 & 27.61 & 1.99 & 27.96 & 4.08 & \cellcolor{tabsecond}28.10 & 7.21 & \cellcolor{tabthird}28.08 & 6.21 & 17.97 & \cellcolor{tabsecond}0.46 & \cellcolor{tabfirst}28.46 & 7.04 & 13.28 & 1.27 & - & 13.87 & 12.96 & \cellcolor{tabfirst}0.29 & 27.97 & 18.46 & OOM & OOM & OOM & OOM \\
meetingroom & 23.76 & \cellcolor{tabsecond}1.25 & 24.91 & 2.12 & \cellcolor{tabfirst}25.12 & 3.54 & \cellcolor{tabthird}25.07 & 3.77 & \cellcolor{tabsecond}25.09 & 3.97 & OOM & OOM & OOM & OOM & 14.28 & \cellcolor{tabthird}1.80 & 24.87 & 11.86 & 13.78 & \cellcolor{tabfirst}0.44 & 16.29 & 9.50 & OOM & OOM & OOM & OOM \\
museum & 19.50 & \cellcolor{tabsecond}1.15 & \cellcolor{tabsecond}20.19 & 3.70 & \cellcolor{tabfirst}20.20 & 3.81 & 20.02 & 6.79 & \cellcolor{tabthird}20.17 & 7.34 & OOM & OOM & OOM & OOM & 12.14 & \cellcolor{tabthird}1.21 & 18.44 & 9.43 & 13.38 & \cellcolor{tabfirst}0.30 & 13.98 & 14.67 & OOM & OOM & OOM & OOM \\
palace & 13.40 & \cellcolor{tabsecond}1.68 & \cellcolor{tabthird}17.37 & \cellcolor{tabthird}4.03 & \cellcolor{tabfirst}17.98 & 8.34 & \cellcolor{tabsecond}17.64 & 14.73 & 17.33 & 13.24 & OOM & OOM & OOM & OOM & OOM & OOM & OOM & OOM & 10.48 & \cellcolor{tabfirst}0.85 & - & 0.85 & OOM & OOM & OOM & OOM \\
panther & 28.12 & 1.28 & 27.82 & 2.70 & 28.45 & 4.94 & 28.59 & 10.79 & \cellcolor{tabthird}28.61 & 10.65 & 19.43 & \cellcolor{tabsecond}0.46 & \cellcolor{tabfirst}29.05 & 8.10 & 14.48 & \cellcolor{tabthird}1.28 & \cellcolor{tabsecond}28.63 & 9.85 & 15.62 & \cellcolor{tabfirst}0.29 & 17.84 & 15.37 & OOM & OOM & OOM & OOM \\
playground & 25.09 & \cellcolor{tabthird}1.13 & 25.37 & 1.78 & 25.45 & 4.51 & \cellcolor{tabsecond}25.89 & 5.34 & 25.53 & 7.69 & 19.90 & \cellcolor{tabsecond}0.53 & \cellcolor{tabthird}25.59 & 5.72 & 16.02 & 1.22 & \cellcolor{tabfirst}26.02 & 6.72 & 14.69 & \cellcolor{tabfirst}0.31 & 17.37 & 7.89 & OOM & OOM & OOM & OOM \\
temple & 12.10 & 1.44 & \cellcolor{tabthird}16.02 & 1.55 & \cellcolor{tabfirst}19.01 & 2.48 & 15.43 & 4.19 & 15.39 & 3.88 & 15.21 & \cellcolor{tabsecond}0.52 & - & 8.96 & 12.94 & \cellcolor{tabthird}1.22 & \cellcolor{tabsecond}18.55 & 6.61 & 12.88 & \cellcolor{tabfirst}0.30 & 14.00 & 9.55 & OOM & OOM & OOM & OOM \\
train & 20.37 & \cellcolor{tabthird}1.16 & \cellcolor{tabthird}21.32 & 1.96 & \cellcolor{tabsecond}21.37 & 3.64 & \cellcolor{tabfirst}21.47 & 8.03 & 21.32 & 7.78 & 16.46 & \cellcolor{tabsecond}0.52 & 20.99 & 6.30 & 13.45 & 1.21 & 21.13 & 6.57 & 11.05 & \cellcolor{tabfirst}0.30 & 11.79 & 9.71 & OOM & OOM & OOM & OOM \\
truck & 12.35 & 1.65 & 24.14 & 1.39 & 24.21 & 2.51 & 24.31 & 4.69 & \cellcolor{tabthird}24.35 & 5.50 & 16.55 & \cellcolor{tabsecond}0.44 & \cellcolor{tabfirst}24.48 & 4.56 & 14.56 & \cellcolor{tabthird}0.86 & \cellcolor{tabsecond}24.42 & 3.88 & 11.45 & \cellcolor{tabfirst}0.22 & 12.20 & 6.83 & OOM & OOM & OOM & OOM \\
\bottomrule
\end{tabular}
}
\caption{Rendering and time results of the feed-forward refinements using \textbf{4096} feats on the Tanks \& Temples. Times are in minutes.}
\label{tab:ffn_refine_Tanks_Temples_4096feats}
\end{table*}

\begin{table*}%[]
\centering
\resizebox{\textwidth}{!}{
\begin{tabular}{lcc|cc|cc|cc|cc|cc|cc|cc|cc|cc|cc|cc|cc}
\toprule
& \multicolumn{10}{c|}{\textbf{Standard COLMAP reconstruction}} & \multicolumn{16}{c}{\textbf{Feed-forward recons}} \\
 & \multicolumn{2}{c|}{512} & \multicolumn{2}{c|}{1024} & \multicolumn{2}{c|}{2048} & \multicolumn{2}{c|}{4096} & \multicolumn{2}{c|}{8192} & \multicolumn{2}{c|}{DA3} & \multicolumn{2}{c|}{Refined DA3} & \multicolumn{2}{c|}{$\pi^3$} & \multicolumn{2}{c|}{Refined $\pi^3$} & \multicolumn{2}{c|}{Fast3R} & \multicolumn{2}{c|}{Refined Fast3R} & \multicolumn{2}{c|}{VGGT} & \multicolumn{2}{c}{Refined VGGT} \\
\textbf{Scene} & PSNR & Time & PSNR & Time & PSNR & Time & PSNR & Time & PSNR & Time & PSNR & Time & PSNR & Time & PSNR & Time & PSNR & Time & PSNR & Time & PSNR & Time & PSNR & Time & PSNR & Time \\
\midrule
auditorium & 23.03 & 2.67 & \cellcolor{tabthird}23.28 & 2.18 & \cellcolor{tabfirst}23.32 & 4.64 & 23.27 & 8.98 & \cellcolor{tabsecond}23.31 & 6.06 & 21.47 & \cellcolor{tabsecond}0.52 & 22.64 & 6.14 & 16.55 & \cellcolor{tabthird}1.21 & 22.36 & 6.73 & 17.90 & \cellcolor{tabfirst}0.30 & 18.72 & 5.17 & OOM & OOM & OOM & OOM \\
ballroom & \cellcolor{tabsecond}21.88 & 2.39 & 19.69 & 5.56 & \cellcolor{tabthird}19.78 & 9.46 & \cellcolor{tabfirst}23.11 & 18.60 & 17.13 & 18.28 & 13.08 & \cellcolor{tabsecond}0.60 & 15.33 & 8.73 & 12.43 & \cellcolor{tabthird}1.40 & 15.68 & 9.66 & 11.97 & \cellcolor{tabfirst}0.34 & 12.89 & 10.46 & OOM & OOM & OOM & OOM \\
barn & 27.44 & \cellcolor{tabsecond}2.58 & 27.66 & \cellcolor{tabthird}4.44 & \cellcolor{tabthird}27.86 & 8.91 & \cellcolor{tabfirst}27.96 & 13.51 & \cellcolor{tabsecond}27.95 & 13.56 & OOM & OOM & OOM & OOM & OOM & OOM & OOM & OOM & 10.28 & \cellcolor{tabfirst}0.53 & - & 0.53 & OOM & OOM & OOM & OOM \\
caterpillar & 23.11 & \cellcolor{tabthird}2.47 & 23.04 & 3.78 & 23.31 & 7.60 & \cellcolor{tabsecond}23.38 & 12.76 & \cellcolor{tabfirst}23.39 & 16.03 & OOM & OOM & OOM & OOM & 15.04 & \cellcolor{tabsecond}1.91 & \cellcolor{tabthird}23.34 & 10.80 & 12.29 & \cellcolor{tabfirst}0.47 & 12.89 & 13.77 & OOM & OOM & OOM & OOM \\
church & 20.75 & \cellcolor{tabsecond}9.61 & 20.70 & 16.47 & \cellcolor{tabthird}21.57 & 29.77 & \cellcolor{tabsecond}21.57 & 38.93 & \cellcolor{tabfirst}21.85 & 68.89 & OOM & OOM & OOM & OOM & OOM & OOM & OOM & OOM & 12.52 & \cellcolor{tabfirst}0.86 & 12.57 & \cellcolor{tabthird}15.52 & OOM & OOM & OOM & OOM \\
courthouse & \cellcolor{tabthird}21.06 & \cellcolor{tabfirst}17.72 & \cellcolor{tabsecond}21.15 & \cellcolor{tabsecond}28.72 & \cellcolor{tabfirst}21.41 & \cellcolor{tabthird}64.73 & 21.02 & 86.76 & 20.81 & 95.86 & OOM & OOM & OOM & OOM & OOM & OOM & OOM & OOM & OOM & OOM & OOM & OOM & OOM & OOM & OOM & OOM \\
courtroom & 22.07 & 1.92 & \cellcolor{tabthird}22.10 & 3.03 & \cellcolor{tabsecond}22.15 & 4.82 & 20.85 & 10.13 & \cellcolor{tabfirst}22.30 & 9.81 & 17.53 & \cellcolor{tabsecond}0.52 & 12.36 & 10.29 & 13.94 & \cellcolor{tabthird}1.22 & 20.56 & 7.25 & 14.80 & \cellcolor{tabfirst}0.30 & - & 0.30 & OOM & OOM & OOM & OOM \\
family & 24.10 & 0.46 & \cellcolor{tabsecond}24.39 & 0.73 & \cellcolor{tabthird}24.38 & 1.36 & 24.33 & 3.27 & 24.34 & 3.05 & 19.25 & \cellcolor{tabsecond}0.17 & 23.38 & 1.73 & 13.98 & 0.35 & \cellcolor{tabfirst}24.80 & 1.85 & 13.51 & \cellcolor{tabfirst}0.09 & - & 3.26 & 19.95 & \cellcolor{tabthird}0.19 & 23.41 & 1.19 \\
francis & 26.74 & 1.40 & 25.87 & 4.71 & \cellcolor{tabthird}27.12 & 3.98 & 26.99 & 6.92 & \cellcolor{tabsecond}27.15 & 6.98 & 22.34 & \cellcolor{tabsecond}0.52 & 27.10 & 11.86 & 17.38 & \cellcolor{tabthird}1.22 & \cellcolor{tabfirst}27.85 & 9.17 & 14.72 & \cellcolor{tabfirst}0.30 & 15.88 & 13.89 & OOM & OOM & OOM & OOM \\
horse & 22.00 & 0.35 & \cellcolor{tabsecond}23.65 & 0.54 & \cellcolor{tabthird}23.56 & 1.05 & 23.49 & 1.46 & 23.36 & 1.39 & 16.49 & \cellcolor{tabsecond}0.16 & 18.03 & 1.84 & 13.44 & 0.35 & \cellcolor{tabfirst}24.27 & 1.46 & 12.85 & \cellcolor{tabfirst}0.09 & 14.71 & 1.44 & 17.45 & \cellcolor{tabthird}0.19 & 19.45 & 1.12 \\
ignatius & 20.30 & 1.13 & \cellcolor{tabsecond}21.16 & 2.09 & \cellcolor{tabthird}21.05 & 3.24 & 19.27 & 8.69 & 20.29 & 13.87 & 17.20 & \cellcolor{tabsecond}0.41 & 20.44 & 4.65 & 12.95 & \cellcolor{tabthird}0.95 & \cellcolor{tabfirst}22.06 & 6.14 & 13.20 & \cellcolor{tabfirst}0.23 & 14.14 & 4.64 & OOM & OOM & OOM & OOM \\
lighthouse & 19.10 & \cellcolor{tabthird}1.03 & 21.31 & 2.54 & \cellcolor{tabsecond}21.49 & 4.12 & \cellcolor{tabfirst}21.75 & 6.42 & \cellcolor{tabthird}21.35 & 13.18 & 17.73 & \cellcolor{tabsecond}0.45 & - & 11.15 & 13.06 & 1.24 & 21.23 & 9.59 & 14.55 & \cellcolor{tabfirst}0.28 & 15.90 & 13.41 & OOM & OOM & OOM & OOM \\
m60 & 25.93 & \cellcolor{tabthird}1.03 & 27.61 & 1.99 & 27.96 & 4.08 & \cellcolor{tabsecond}28.10 & 7.21 & \cellcolor{tabthird}28.08 & 6.21 & 17.97 & \cellcolor{tabsecond}0.46 & \cellcolor{tabfirst}28.31 & 6.15 & 13.28 & 1.27 & - & 9.55 & 12.96 & \cellcolor{tabfirst}0.29 & - & 0.29 & OOM & OOM & OOM & OOM \\
meetingroom & 23.76 & \cellcolor{tabsecond}1.25 & 24.91 & 2.12 & \cellcolor{tabfirst}25.12 & 3.54 & \cellcolor{tabthird}25.07 & 3.77 & \cellcolor{tabsecond}25.09 & 3.97 & OOM & OOM & OOM & OOM & 14.28 & \cellcolor{tabthird}1.80 & 24.80 & 7.02 & 13.78 & \cellcolor{tabfirst}0.44 & 15.91 & 10.00 & OOM & OOM & OOM & OOM \\
museum & 19.50 & \cellcolor{tabsecond}1.15 & \cellcolor{tabsecond}20.19 & 3.70 & \cellcolor{tabfirst}20.20 & 3.81 & 20.02 & 6.79 & \cellcolor{tabthird}20.17 & 7.34 & OOM & OOM & OOM & OOM & 12.14 & \cellcolor{tabthird}1.21 & 17.09 & 6.55 & 13.38 & \cellcolor{tabfirst}0.30 & 15.35 & 7.67 & OOM & OOM & OOM & OOM \\
palace & 13.40 & \cellcolor{tabsecond}1.68 & \cellcolor{tabthird}17.37 & \cellcolor{tabthird}4.03 & \cellcolor{tabfirst}17.98 & 8.34 & \cellcolor{tabsecond}17.64 & 14.73 & 17.33 & 13.24 & OOM & OOM & OOM & OOM & OOM & OOM & OOM & OOM & 10.48 & \cellcolor{tabfirst}0.85 & - & 0.85 & OOM & OOM & OOM & OOM \\
panther & 28.12 & 1.28 & 27.82 & 2.70 & 28.45 & 4.94 & 28.59 & 10.79 & \cellcolor{tabthird}28.61 & 10.65 & 19.43 & \cellcolor{tabsecond}0.46 & \cellcolor{tabfirst}28.89 & 6.91 & 14.48 & \cellcolor{tabthird}1.28 & \cellcolor{tabsecond}28.80 & 6.26 & 15.62 & \cellcolor{tabfirst}0.29 & - & 0.29 & OOM & OOM & OOM & OOM \\
playground & 25.09 & \cellcolor{tabthird}1.13 & 25.37 & 1.78 & 25.45 & 4.51 & \cellcolor{tabsecond}25.89 & 5.34 & \cellcolor{tabthird}25.53 & 7.69 & 19.90 & \cellcolor{tabsecond}0.53 & 25.37 & 4.45 & 16.02 & 1.22 & \cellcolor{tabfirst}25.94 & 4.12 & 14.69 & \cellcolor{tabfirst}0.31 & 17.29 & 9.72 & OOM & OOM & OOM & OOM \\
temple & 12.10 & 1.44 & \cellcolor{tabthird}16.02 & 1.55 & \cellcolor{tabfirst}19.01 & 2.48 & 15.43 & 4.19 & 15.39 & 3.88 & 15.21 & \cellcolor{tabsecond}0.52 & 10.56 & 6.28 & 12.94 & \cellcolor{tabthird}1.22 & \cellcolor{tabsecond}17.95 & 5.03 & 12.88 & \cellcolor{tabfirst}0.30 & 12.55 & 5.63 & OOM & OOM & OOM & OOM \\
train & 20.37 & \cellcolor{tabthird}1.16 & \cellcolor{tabthird}21.32 & 1.96 & \cellcolor{tabsecond}21.37 & 3.64 & \cellcolor{tabfirst}21.47 & 8.03 & 21.32 & 7.78 & 16.46 & \cellcolor{tabsecond}0.52 & 20.97 & 4.92 & 13.45 & 1.21 & 21.14 & 4.90 & 11.05 & \cellcolor{tabfirst}0.30 & 11.66 & 5.03 & OOM & OOM & OOM & OOM \\
truck & 12.35 & 1.65 & 24.14 & 1.39 & 24.21 & 2.51 & 24.31 & 4.69 & \cellcolor{tabsecond}24.35 & 5.50 & 16.55 & \cellcolor{tabsecond}0.44 & \cellcolor{tabfirst}24.36 & 3.21 & 14.56 & \cellcolor{tabthird}0.86 & \cellcolor{tabthird}24.32 & 2.75 & 11.45 & \cellcolor{tabfirst}0.22 & 12.02 & 3.33 & OOM & OOM & OOM & OOM \\
\bottomrule
\end{tabular}
}
\caption{Rendering and time results of the feed-forward refinements using \textbf{2048} feats on the Tanks \& Temples. Times are in minutes.}
\label{tab:ffn_refine_Tanks_Temples_2048feats}
\end{table*}

%%%%%%%%%%%%%%%%%%%%%%%%%%%%%%%%%%%%%%%%%%%%%%%%%%%
%%%%%%%%%%%%%%%%%%%%%%%%%%%%%%%%%%%%%%%%%%%%%%%%%%%
%%%%%%%%%%%%%%%%%%%%%%%%%%%%%%%%%%%%%%%%%%%%%%%%%%%

\section{3D reconstruction statistics}
\label{sec:recons_stats}
We also report some information per scene from the 3D reconstructions obtained using the Mip-NeRF 360 dataset, for each feature extractor and for the different number of features. The information reported is: total number of keypoints, total number of matches, the number of registered views, and the total number of 3D points. Table~\ref{tab:reconsStats_Mip-NeRF360_colmap_LG} shows the reconstruction statistics for the COLMAP mapper and LightGlue matcher, and Table~\ref{tab:reconsStats_Mip-NeRF360_colmap_SMNNk-opt} for the COLMAP mapper and Mutual Nearest Neighbors matcher. This information can help to understand how good each learned detector is in registering images and also can help us to understand the impact on the runtime.

\begin{table*}%[]
\centering
\resizebox{\textwidth}{!}{
\begin{tabular}{l|l|cccc|cccc|cccc|cccc|cccc|cccc|cccc}
\hline
\multirow{2}{*}{Detector} & \multirow{2}{*}{Scene} & \multicolumn{4}{c|}{128} & \multicolumn{4}{c|}{256} & \multicolumn{4}{c|}{512} & \multicolumn{4}{c|}{1024} & \multicolumn{4}{c|}{2048} & \multicolumn{4}{c|}{4096} & \multicolumn{4}{c}{8192} \\
\cline{3-30}
 & & \multicolumn{1}{c}{\begin{tabular}[c]{@{}c@{}}2D \\ points\end{tabular}} & \multicolumn{1}{c}{Matches} & \multicolumn{1}{c}{\begin{tabular}[c]{@{}c@{}}3D\\ points\end{tabular}} & \multicolumn{1}{c|}{\begin{tabular}[c]{@{}c@{}}Reg.\\ views\end{tabular}} & \multicolumn{1}{c}{\begin{tabular}[c]{@{}c@{}}2D \\ points\end{tabular}} & \multicolumn{1}{c}{Matches} & \multicolumn{1}{c}{\begin{tabular}[c]{@{}c@{}}3D\\ points\end{tabular}} & \multicolumn{1}{c|}{\begin{tabular}[c]{@{}c@{}}Reg.\\ views\end{tabular}} & \multicolumn{1}{c}{\begin{tabular}[c]{@{}c@{}}2D \\ points\end{tabular}} & \multicolumn{1}{c}{Matches} & \multicolumn{1}{c}{\begin{tabular}[c]{@{}c@{}}3D\\ points\end{tabular}} & \multicolumn{1}{c|}{\begin{tabular}[c]{@{}c@{}}Reg.\\ views\end{tabular}} & \multicolumn{1}{c}{\begin{tabular}[c]{@{}c@{}}2D \\ points\end{tabular}} & \multicolumn{1}{c}{Matches} & \multicolumn{1}{c}{\begin{tabular}[c]{@{}c@{}}3D\\ points\end{tabular}} & \multicolumn{1}{c|}{\begin{tabular}[c]{@{}c@{}}Reg.\\ views\end{tabular}} & \multicolumn{1}{c}{\begin{tabular}[c]{@{}c@{}}2D \\ points\end{tabular}} & \multicolumn{1}{c}{Matches} & \multicolumn{1}{c}{\begin{tabular}[c]{@{}c@{}}3D\\ points\end{tabular}} & \multicolumn{1}{c|}{\begin{tabular}[c]{@{}c@{}}Reg.\\ views\end{tabular}} & \multicolumn{1}{c}{\begin{tabular}[c]{@{}c@{}}2D \\ points\end{tabular}} & \multicolumn{1}{c}{Matches} & \multicolumn{1}{c}{\begin{tabular}[c]{@{}c@{}}3D\\ points\end{tabular}} & \multicolumn{1}{c|}{\begin{tabular}[c]{@{}c@{}}Reg.\\ views\end{tabular}} & \multicolumn{1}{c}{\begin{tabular}[c]{@{}c@{}}2D \\ points\end{tabular}} & \multicolumn{1}{c}{Matches} & \multicolumn{1}{c}{\begin{tabular}[c]{@{}c@{}}3D\\ points\end{tabular}} & \multicolumn{1}{c}{\begin{tabular}[c]{@{}c@{}}Reg.\\ views\end{tabular}} \\
\hline
\multirow{9}{*}{ALIKED} & bicycle & 24.8K & 0.2M & 38 & 2 & 49.7K & 0.4M & 6.4K & 193 & 99.3K & 0.7M & 1.5K & 194 & 0.2M & 1.1M & 3.0K & 194 & 0.4M & 2.1M & 5.8K & 194 & 0.8M & 4.0M & 1.1K & 194 & 1.0M & 4.7M & 1.4K & 194 \\
 & bonsai & 37.4K & 0.5M & 4.5K & 282 & 74.8K & 1.1M & 8.7K & 292 & 0.1M & 1.9M & 1.5K & 292 & 0.3M & 3.8M & 2.3K & 292 & 0.6M & 7.8M & 3.7K & 292 & 0.8M & 10.8M & 4.8K & 292 & 0.8M & 10.8M & 4.8K & 292 \\
 & counter & 30.7K & 0.3M & 35 & 2 & 61.4K & 0.7M & 8.6K & 240 & 0.1M & 1.3M & 1.5K & 240 & 0.2M & 2.5M & 2.7K & 240 & 0.5M & 4.9M & 4.8K & 240 & 0.6M & 6.2M & 5.9K & 240 & 0.6M & 6.2M & 5.9K & 240 \\
 & flowers & 22.1K & 0.1M & 28 & 2 & 44.3K & 0.3M & 37 & 2 & 88.6K & 0.6M & 1.3K & 173 & 0.2M & 1.0M & 2.7K & 173 & 0.4M & 1.7M & 5.2K & 173 & 0.7M & 3.0M & 1.0K & 173 & 0.8M & 3.4M & 1.2K & 173 \\
 & garden & 23.7K & 0.2M & 3.1K & 185 & 47.4K & 0.5M & 6.4K & 185 & 94.7K & 0.9M & 1.2K & 185 & 0.2M & 1.9M & 2.1K & 185 & 0.4M & 3.9M & 3.8K & 185 & 0.8M & 8.0M & 7.1K & 185 & 0.8M & 8.7M & 7.7K & 185 \\
 & kitchen & 35.7K & 0.5M & 3.9K & 279 & 71.4K & 1.2M & 6.7K & 279 & 0.1M & 2.3M & 1.2K & 279 & 0.3M & 4.8M & 1.9K & 279 & 0.6M & 9.8M & 3.2K & 279 & 1.0M & 16.7M & 4.6K & 279 & 1.0M & 17.6M & 4.8K & 279 \\
 & room & 39.8K & 0.8M & 3.8K & 311 & 79.6K & 1.7M & 7.5K & 311 & 0.2M & 3.3M & 1.4K & 311 & 0.3M & 5.9M & 2.5K & 311 & 0.5M & 9.4M & 3.7K & 311 & 0.6M & 9.9M & 3.9K & 311 & 0.6M & 9.9M & 3.9K & 311 \\
 & stump & 16.0K & 56.2K & 17 & 2 & 32.0K & 0.1M & 50 & 2 & 64.0K & 0.2M & 8.5K & 125 & 0.1M & 0.4M & 1.8K & 125 & 0.3M & 0.7M & 3.7K & 125 & 0.5M & 1.5M & 7.3K & 125 & 0.6M & 1.8M & 9.0K & 125 \\
 & treehill & 18.0K & 90.0K & 233 & 16 & 36.1K & 0.2M & 4.2K & 141 & 72.2K & 0.4M & 9.2K & 141 & 0.1M & 0.7M & 1.9K & 141 & 0.3M & 1.2M & 3.7K & 141 & 0.6M & 2.3M & 7.0K & 141 & 0.6M & 2.4M & 7.3K & 141 \\
\hline
\multirow{9}{*}{DISK} & bicycle & 24.8K & 0.3M & 168 & 8 & 49.7K & 0.5M & 4.9K & 128 & 99.3K & 0.7M & 1.6K & 194 & 0.2M & 0.9M & 3.2K & 194 & 0.4M & 1.6M & 6.2K & 194 & 0.8M & 3.1M & 1.2K & 194 & 1.6M & 5.6M & 2.1K & 194 \\
 & bonsai & 37.4K & 0.9M & 4.0K & 235 & 74.8K & 1.4M & 9.9K & 284 & 0.1M & 2.2M & 1.8K & 292 & 0.3M & 3.8M & 2.7K & 292 & 0.6M & 7.1M & 4.2K & 292 & 1.2M & 13.5M & 7.5K & 292 & 1.8M & 18.7M & 1.2K & 292 \\
 & counter & 30.7K & 0.6M & 4.4K & 238 & 61.4K & 1.1M & 8.7K & 240 & 0.1M & 1.8M & 1.6K & 240 & 0.2M & 2.9M & 2.9K & 240 & 0.5M & 5.0M & 5.2K & 240 & 1.0M & 8.8M & 9.3K & 240 & 1.7M & 13.1M & 1.5K & 240 \\
 & flowers & 22.1K & 0.2M & 27 & 2 & 44.3K & 0.4M & 4.3K & 103 & 88.6K & 0.6M & 1.5K & 173 & 0.2M & 0.8M & 2.9K & 173 & 0.4M & 1.2M & 5.6K & 173 & 0.7M & 2.3M & 1.0K & 173 & 1.4M & 4.3M & 1.9K & 173 \\
 & garden & 23.7K & 0.3M & 3.3K & 185 & 47.4K & 0.6M & 6.7K & 185 & 94.7K & 1.0M & 1.3K & 185 & 0.2M & 1.6M & 2.3K & 185 & 0.4M & 3.0M & 4.2K & 185 & 0.8M & 6.5M & 7.7K & 185 & 1.5M & 13.0M & 1.4K & 185 \\
 & kitchen & 35.7K & 0.4M & 3.7K & 279 & 71.4K & 0.9M & 7.0K & 279 & 0.1M & 1.8M & 1.2K & 279 & 0.3M & 3.6M & 1.9K & 279 & 0.6M & 7.7M & 3.1K & 279 & 1.1M & 16.6M & 5.2K & 279 & 2.2M & 32.5M & 9.0K & 279 \\
 & room & 39.8K & 1.0M & 1.5K & 143 & 79.6K & 1.9M & 7.9K & 305 & 0.2M & 3.2M & 1.6K & 311 & 0.3M & 5.7M & 2.8K & 311 & 0.6M & 10.6M & 5.0K & 311 & 1.3M & 21.0M & 9.0K & 311 & 2.0M & 33.9M & 1.3K & 311 \\
 & stump & 16.0K & 95.2K & 75 & 3 & 32.0K & 0.2M & 1.6K & 41 & 64.0K & 0.2M & 8.3K & 118 & 0.1M & 0.3M & 1.8K & 125 & 0.3M & 0.5M & 3.5K & 125 & 0.5M & 1.1M & 7.1K & 125 & 1.0M & 1.9M & 1.4K & 125 \\
 & treehill & 18.0K & 0.1M & 49 & 2 & 36.1K & 0.2M & 2.0K & 52 & 72.2K & 0.4M & 1.1K & 141 & 0.1M & 0.5M & 2.2K & 141 & 0.3M & 0.9M & 4.0K & 141 & 0.6M & 1.9M & 7.6K & 141 & 1.1M & 3.8M & 1.4K & 141 \\
\hline
\multirow{9}{*}{\begin{tabular}[c]{@{}c@{}}DoG- \\ HardNet\end{tabular}} & bicycle & 16.5K & 67.4K & 422 & 32 & 34.1K & 0.2M & 2.2K & 99 & 70.3K & 0.5M & 6.8K & 163 & 0.1M & 0.9M & 1.9K & 194 & 0.3M & 1.7M & 4.5K & 194 & 0.6M & 3.1M & 9.6K & 194 & 1.3M & 5.3M & 2.0K & 194 \\
 & bonsai & 27.9K & 0.3M & 2.3K & 226 & 57.3K & 0.9M & 7.0K & 292 & 0.1M & 1.7M & 1.3K & 292 & 0.2M & 2.7M & 2.3K & 292 & 0.5M & 4.7M & 3.8K & 292 & 0.9M & 9.0M & 6.6K & 292 & 1.9M & 18.5M & 1.3K & 292 \\
 & counter & 22.5K & 0.1M & 1.1K & 94 & 46.8K & 0.5M & 6.5K & 240 & 97.1K & 1.2M & 1.3K & 240 & 0.2M & 2.2M & 2.6K & 240 & 0.4M & 4.1M & 5.0K & 240 & 0.8M & 7.6M & 9.3K & 240 & 1.7M & 13.5M & 1.7K & 240 \\
 & flowers & 14.5K & 80.3K & 51 & 2 & 29.9K & 0.2M & 348 & 16 & 61.8K & 0.5M & 5.2K & 120 & 0.1M & 0.8M & 1.9K & 173 & 0.3M & 1.4M & 4.2K & 173 & 0.6M & 2.5M & 8.7K & 173 & 1.1M & 4.1M & 1.8K & 173 \\
 & garden & 17.0K & 0.1M & 1.7K & 176 & 34.6K & 0.4M & 3.9K & 185 & 71.1K & 0.8M & 8.4K & 185 & 0.1M & 1.6M & 1.8K & 185 & 0.3M & 3.0M & 3.6K & 185 & 0.6M & 5.9M & 7.4K & 185 & 1.2M & 11.0M & 1.5K & 185 \\
 & kitchen & 25.1K & 0.3M & 1.7K & 171 & 51.1K & 0.7M & 4.9K & 279 & 0.1M & 1.6M & 9.2K & 279 & 0.2M & 3.3M & 1.7K & 279 & 0.4M & 6.5M & 2.9K & 279 & 0.9M & 12.4M & 4.9K & 279 & 1.7M & 22.6M & 9.0K & 279 \\
 & room & 29.3K & 0.3M & 1.4K & 158 & 61.6K & 0.9M & 6.3K & 311 & 0.1M & 2.2M & 1.2K & 311 & 0.3M & 4.5M & 2.5K & 311 & 0.5M & 8.4M & 5.1K & 311 & 1.1M & 14.8M & 9.7K & 311 & 1.7M & 25.8M & 1.5K & 311 \\
 & stump & 10.0K & 49.4K & 204 & 20 & 20.7K & 0.2M & 870 & 52 & 43.1K & 0.3M & 3.0K & 88 & 90.4K & 0.4M & 9.3K & 115 & 0.2M & 0.7M & 2.4K & 125 & 0.4M & 1.3M & 5.5K & 125 & 0.8M & 2.2M & 1.2K & 125 \\
 & treehill & 11.7K & 30.6K & 35 & 2 & 24.2K & 0.1M & 244 & 12 & 50.2K & 0.2M & 1.7K & 40 & 0.1M & 0.4M & 1.1K & 127 & 0.2M & 0.7M & 3.1K & 141 & 0.4M & 1.4M & 6.5K & 141 & 0.9M & 2.7M & 1.2K & 141 \\
\hline
\multirow{9}{*}{SIFT} & bicycle & 16.5K & 28.9K & 46 & 2 & 34.1K & 96.4K & 2.1K & 102 & 70.3K & 0.2M & 5.7K & 143 & 0.1M & 0.5M & 1.6K & 192 & 0.3M & 1.0M & 3.9K & 194 & 0.6M & 1.8M & 8.4K & 194 & 1.3M & 3.1M & 1.7K & 194 \\
 & bonsai & 27.9K & 0.2M & 2.2K & 224 & 57.3K & 0.5M & 6.6K & 292 & 0.1M & 0.9M & 1.3K & 292 & 0.2M & 1.8M & 2.4K & 292 & 0.5M & 3.3M & 4.1K & 292 & 0.9M & 6.7M & 7.0K & 292 & 1.9M & 13.8M & 1.3K & 292 \\
 & counter & 22.5K & 0.1M & 1.1K & 94 & 46.8K & 0.4M & 6.4K & 240 & 97.1K & 0.8M & 1.3K & 240 & 0.2M & 1.6M & 2.5K & 240 & 0.4M & 2.9M & 4.9K & 240 & 0.8M & 5.3M & 9.2K & 240 & 1.7M & 9.6M & 1.7K & 240 \\
 & flowers & 14.5K & 15.3K & 50 & 2 & 29.9K & 82.3K & 49 & 2 & 61.8K & 0.2M & 2.9K & 71 & 0.1M & 0.4M & 1.6K & 173 & 0.3M & 0.6M & 3.6K & 173 & 0.6M & 1.0M & 7.7K & 173 & 1.1M & 1.9M & 1.6K & 173 \\
 & garden & 17.0K & 0.1M & 1.7K & 177 & 34.6K & 0.3M & 3.8K & 185 & 71.1K & 0.7M & 8.1K & 185 & 0.1M & 1.3M & 1.7K & 185 & 0.3M & 2.5M & 3.5K & 185 & 0.6M & 4.7M & 7.0K & 185 & 1.2M & 8.5M & 1.4K & 185 \\
 & kitchen & 25.1K & 0.2M & 1.6K & 171 & 51.1K & 0.6M & 5.1K & 279 & 0.1M & 1.2M & 9.4K & 279 & 0.2M & 2.4M & 1.6K & 279 & 0.4M & 4.8M & 3.0K & 279 & 0.9M & 8.8M & 5.2K & 279 & 1.7M & 15.7M & 9.6K & 279 \\
 & room & 29.3K & 0.3M & 1.2K & 132 & 61.6K & 0.9M & 6.3K & 311 & 0.1M & 1.9M & 1.2K & 311 & 0.3M & 3.8M & 2.5K & 311 & 0.5M & 7.1M & 5.0K & 311 & 1.1M & 12.8M & 9.5K & 311 & 1.7M & 22.6M & 1.4K & 311 \\
 & stump & 10.0K & 7.0K & 205 & 20 & 20.7K & 28.7K & 869 & 51 & 43.1K & 89.0K & 2.6K & 81 & 90.4K & 0.2M & 7.7K & 112 & 0.2M & 0.3M & 1.9K & 125 & 0.4M & 0.6M & 4.3K & 125 & 0.8M & 1.2M & 9.3K & 125 \\
 & treehill & 11.7K & 5.2K & 33 & 2 & 24.2K & 26.0K & 341 & 15 & 50.2K & 84.9K & 1.7K & 40 & 0.1M & 0.2M & 1.0K & 136 & 0.2M & 0.4M & 2.5K & 141 & 0.4M & 0.9M & 5.5K & 141 & 0.9M & 1.8M & 1.0K & 141 \\
\hline
\multirow{9}{*}{SuperPoint} & bicycle & 24.8K & 78.0K & 33 & 2 & 49.7K & 0.3M & 6.0K & 192 & 99.3K & 0.7M & 1.4K & 194 & 0.2M & 1.6M & 2.9K & 194 & 0.4M & 2.6M & 6.1K & 194 & 0.6M & 3.5M & 9.9K & 194 & 0.6M & 3.5M & 1.0K & 194 \\
 & bonsai & 37.4K & 0.5M & 4.8K & 285 & 74.8K & 1.3M & 8.8K & 292 & 0.1M & 2.5M & 1.6K & 292 & 0.3M & 4.5M & 2.9K & 292 & 0.6M & 7.3M & 5.8K & 292 & 0.9M & 9.9M & 9.1K & 292 & 0.9M & 9.9M & 9.2K & 292 \\
 & counter & 30.7K & 0.2M & 34 & 2 & 61.4K & 0.6M & 8.7K & 240 & 0.1M & 1.3M & 1.7K & 240 & 0.2M & 2.9M & 3.1K & 240 & 0.5M & 5.4M & 5.9K & 240 & 0.9M & 8.1M & 1.1K & 240 & 0.9M & 8.1M & 1.1K & 240 \\
 & flowers & 22.1K & 71.8K & 50 & 2 & 44.3K & 0.3M & 4.5K & 130 & 88.6K & 0.6M & 1.4K & 173 & 0.2M & 1.2M & 2.8K & 173 & 0.4M & 2.2M & 5.9K & 173 & 0.6M & 3.1M & 1.1K & 173 & 0.7M & 3.2M & 1.1K & 173 \\
 & garden & 23.7K & 0.2M & 2.8K & 185 & 47.4K & 0.5M & 5.5K & 185 & 94.7K & 1.1M & 1.1K & 185 & 0.2M & 2.4M & 2.3K & 185 & 0.4M & 4.4M & 4.7K & 185 & 0.6M & 6.5M & 8.2K & 185 & 0.6M & 6.6M & 8.3K & 185 \\
 & kitchen & 35.7K & 0.5M & 3.8K & 279 & 71.4K & 1.4M & 6.9K & 279 & 0.1M & 3.1M & 1.3K & 279 & 0.3M & 6.2M & 2.5K & 279 & 0.6M & 10.5M & 5.1K & 279 & 0.9M & 14.5M & 8.4K & 279 & 0.9M & 15.1M & 8.9K & 279 \\
 & room & 39.8K & 0.4M & 344 & 25 & 79.6K & 1.3M & 8.2K & 311 & 0.2M & 3.1M & 1.5K & 311 & 0.3M & 6.3M & 2.9K & 311 & 0.6M & 10.9M & 5.7K & 311 & 0.8M & 13.3M & 7.3K & 311 & 0.8M & 13.3M & 7.3K & 311 \\
 & stump & 16.0K & 16.9K & 25 & 2 & 32.0K & 94.3K & 892 & 29 & 64.0K & 0.2M & 7.8K & 125 & 0.1M & 0.4M & 1.8K & 125 & 0.3M & 0.8M & 3.8K & 125 & 0.5M & 1.3M & 7.1K & 125 & 0.5M & 1.4M & 7.5K & 125 \\
 & treehill & 18.0K & 48.1K & 35 & 2 & 36.1K & 0.2M & 3.8K & 131 & 72.2K & 0.4M & 1.0K & 141 & 0.1M & 0.7M & 2.2K & 141 & 0.3M & 1.1M & 4.3K & 141 & 0.4M & 1.4M & 6.2K & 141 & 0.4M & 1.4M & 6.5K & 141 \\
\hline
\multirow{9}{*}{XFeat} & bicycle & 24.8K & 0.1M & 20 & 2 & 49.7K & 0.3M & 46 & 2 & 99.3K & 0.5M & 1.2K & 192 & 0.2M & 0.6M & 2.6K & 194 & 0.4M & 1.0M & 5.3K & 194 & 0.8M & 1.7M & 1.0K & 194 & 1.6M & 3.2M & 2.0K & 194 \\
 & bonsai & 37.4K & 0.5M & 16 & 2 & 74.8K & 0.8M & 1.1K & 292 & 0.1M & 1.2M & 2.0K & 292 & 0.3M & 2.0M & 3.7K & 292 & 0.6M & 4.0M & 6.4K & 292 & 1.2M & 8.0M & 1.1K & 292 & 2.4M & 15.5M & 2.1K & 292 \\
 & counter & 30.7K & 0.3M & 30 & 2 & 61.4K & 0.6M & 1.0K & 240 & 0.1M & 0.9M & 1.9K & 240 & 0.2M & 1.5M & 3.6K & 240 & 0.5M & 2.9M & 6.6K & 240 & 1.0M & 5.7M & 1.2K & 240 & 2.0M & 11.0M & 2.1K & 240 \\
 & flowers & 22.1K & 0.1M & - & - & 44.3K & 0.3M & 33 & 2 & 88.6K & 0.4M & 67 & 2 & 0.2M & 0.4M & 1.4K & 124 & 0.4M & 0.5M & 4.0K & 173 & 0.7M & 0.8M & 7.7K & 173 & 1.4M & 1.5M & 1.6K & 173 \\
 & garden & 23.7K & 80.6K & 23 & 2 & 47.4K & 0.3M & 55 & 2 & 94.7K & 0.5M & 1.6K & 185 & 0.2M & 0.9M & 3.2K & 185 & 0.4M & 1.7M & 6.1K & 185 & 0.8M & 3.6M & 1.1K & 185 & 1.5M & 7.5M & 2.0K & 185 \\
 & kitchen & 35.7K & 0.4M & 51 & 2 & 71.4K & 0.7M & 9.8K & 279 & 0.1M & 1.5M & 1.7K & 279 & 0.3M & 3.2M & 3.0K & 279 & 0.6M & 6.9M & 5.1K & 279 & 1.1M & 14.0M & 8.8K & 279 & 2.3M & 26.7M & 1.5K & 279 \\
 & room & 39.8K & 0.4M & 35 & 2 & 79.6K & 0.9M & 1.1K & 311 & 0.2M & 1.8M & 1.9K & 311 & 0.3M & 3.9M & 3.5K & 311 & 0.6M & 8.3M & 6.1K & 311 & 1.3M & 17.2M & 1.1K & 311 & 2.5M & 33.6M & 2.0K & 311 \\
 & stump & 16.0K & 57.4K & - & - & 32.0K & 0.1M & 40 & 2 & 64.0K & 0.2M & 58 & 2 & 0.1M & 0.2M & 7.0K & 94 & 0.3M & 0.2M & 1.9K & 125 & 0.5M & 0.4M & 4.3K & 125 & 1.0M & 0.8M & 9.4K & 125 \\
 & treehill & 18.0K & 78.9K & - & - & 36.1K & 0.2M & 28 & 2 & 72.2K & 0.3M & 65 & 2 & 0.1M & 0.4M & 1.6K & 141 & 0.3M & 0.5M & 3.2K & 141 & 0.6M & 0.9M & 6.2K & 141 & 1.2M & 1.8M & 1.2K & 141 \\
\hline
\end{tabular}
}
\caption{3D reconstruction stats on \textbf{MipNeRF 360} dataset using \textbf{COLMAP} mapper and \textbf{LightGlue} matcher}
\label{tab:reconsStats_Mip-NeRF360_colmap_LG}
\end{table*}

\begin{table*}%[]
\centering
\resizebox{\textwidth}{!}{
\begin{tabular}{l|l|cccc|cccc|cccc|cccc|cccc|cccc|cccc}
\hline
\multirow{2}{*}{Detector} & \multirow{2}{*}{Scene} & \multicolumn{4}{c|}{128} & \multicolumn{4}{c|}{256} & \multicolumn{4}{c|}{512} & \multicolumn{4}{c|}{1024} & \multicolumn{4}{c|}{2048} & \multicolumn{4}{c|}{4096} & \multicolumn{4}{c}{8192} \\
\cline{3-30}
 & & \multicolumn{1}{c}{\begin{tabular}[c]{@{}c@{}}2D \\ points\end{tabular}} & \multicolumn{1}{c}{Matches} & \multicolumn{1}{c}{\begin{tabular}[c]{@{}c@{}}3D\\ points\end{tabular}} & \multicolumn{1}{c|}{\begin{tabular}[c]{@{}c@{}}Reg.\\ views\end{tabular}} & \multicolumn{1}{c}{\begin{tabular}[c]{@{}c@{}}2D \\ points\end{tabular}} & \multicolumn{1}{c}{Matches} & \multicolumn{1}{c}{\begin{tabular}[c]{@{}c@{}}3D\\ points\end{tabular}} & \multicolumn{1}{c|}{\begin{tabular}[c]{@{}c@{}}Reg.\\ views\end{tabular}} & \multicolumn{1}{c}{\begin{tabular}[c]{@{}c@{}}2D \\ points\end{tabular}} & \multicolumn{1}{c}{Matches} & \multicolumn{1}{c}{\begin{tabular}[c]{@{}c@{}}3D\\ points\end{tabular}} & \multicolumn{1}{c|}{\begin{tabular}[c]{@{}c@{}}Reg.\\ views\end{tabular}} & \multicolumn{1}{c}{\begin{tabular}[c]{@{}c@{}}2D \\ points\end{tabular}} & \multicolumn{1}{c}{Matches} & \multicolumn{1}{c}{\begin{tabular}[c]{@{}c@{}}3D\\ points\end{tabular}} & \multicolumn{1}{c|}{\begin{tabular}[c]{@{}c@{}}Reg.\\ views\end{tabular}} & \multicolumn{1}{c}{\begin{tabular}[c]{@{}c@{}}2D \\ points\end{tabular}} & \multicolumn{1}{c}{Matches} & \multicolumn{1}{c}{\begin{tabular}[c]{@{}c@{}}3D\\ points\end{tabular}} & \multicolumn{1}{c|}{\begin{tabular}[c]{@{}c@{}}Reg.\\ views\end{tabular}} & \multicolumn{1}{c}{\begin{tabular}[c]{@{}c@{}}2D \\ points\end{tabular}} & \multicolumn{1}{c}{Matches} & \multicolumn{1}{c}{\begin{tabular}[c]{@{}c@{}}3D\\ points\end{tabular}} & \multicolumn{1}{c|}{\begin{tabular}[c]{@{}c@{}}Reg.\\ views\end{tabular}} & \multicolumn{1}{c}{\begin{tabular}[c]{@{}c@{}}2D \\ points\end{tabular}} & \multicolumn{1}{c}{Matches} & \multicolumn{1}{c}{\begin{tabular}[c]{@{}c@{}}3D\\ points\end{tabular}} & \multicolumn{1}{c}{\begin{tabular}[c]{@{}c@{}}Reg.\\ views\end{tabular}} \\
\hline
\multirow{9}{*}{ALIKED} & bicycle & 24.8K & 0.6M & 35 & 2 & 49.7K & 1.0M & 6.1K & 188 & 99.3K & 2.0M & 1.5K & 193 & 0.2M & 3.7M & 2.9K & 193 & 0.4M & 6.8M & 5.7K & 194 & 0.8M & 11.3M & 1.1K & 194 & 1.0M & 12.4M & 1.3K & 194 \\
 & bonsai & 37.4K & 1.3M & 5.0K & 292 & 74.8K & 2.4M & 9.2K & 292 & 0.1M & 4.3M & 1.8K & 292 & 0.3M & 7.7M & 2.7K & 292 & 0.6M & 13.1M & 4.5K & 292 & 0.8M & 16.1M & 6.0K & 292 & 0.8M & 16.1M & 6.0K & 292 \\
 & counter & 30.7K & 0.9M & 276 & 13 & 61.4K & 1.6M & 9.4K & 240 & 0.1M & 3.2M & 1.7K & 236 & 0.2M & 5.9M & 3.2K & 240 & 0.5M & 9.5M & 5.7K & 240 & 0.6M & 10.7M & 7.2K & 240 & 0.6M & 10.7M & 7.3K & 240 \\
 & flowers & 22.1K & 0.4M & 41 & 2 & 44.3K & 0.8M & 25 & 2 & 88.6K & 1.6M & 1.0K & 139 & 0.2M & 3.0M & 2.6K & 173 & 0.4M & 5.6M & 5.1K & 173 & 0.7M & 9.7M & 9.8K & 173 & 0.8M & 10.9M & 1.1K & 173 \\
 & garden & 23.7K & 0.6M & 3.3K & 185 & 47.4K & 1.1M & 6.7K & 185 & 94.7K & 2.1M & 1.3K & 185 & 0.2M & 4.1M & 2.3K & 185 & 0.4M & 7.6M & 4.1K & 185 & 0.8M & 13.3M & 7.6K & 185 & 0.8M & 14.1M & 8.2K & 185 \\
 & kitchen & 35.7K & 1.0M & 4.3K & 279 & 71.4K & 1.9M & 8.3K & 279 & 0.1M & 3.4M & 1.5K & 279 & 0.3M & 5.8M & 2.5K & 279 & 0.6M & 9.7M & 4.4K & 279 & 1.0M & 13.8M & 6.6K & 279 & 1.0M & 14.3M & 6.6K & 279 \\
 & room & 39.8K & 1.6M & 3.9K & 311 & 79.6K & 3.0M & 7.6K & 311 & 0.2M & 5.5M & 1.4K & 306 & 0.3M & 9.1M & 2.7K & 311 & 0.5M & 12.3M & 4.2K & 311 & 0.6M & 12.6M & 4.2K & 310 & 0.6M & 12.6M & 4.4K & 311 \\
 & stump & 16.0K & 0.2M & 28 & 2 & 32.0K & 0.4M & 30 & 2 & 64.0K & 0.8M & 8.3K & 125 & 0.1M & 1.6M & 1.9K & 125 & 0.3M & 3.0M & 3.7K & 125 & 0.5M & 5.4M & 7.0K & 125 & 0.6M & 6.1M & 8.4K & 125 \\
 & treehill & 18.0K & 0.3M & 44 & 2 & 36.1K & 0.6M & 4.4K & 141 & 72.2K & 1.1M & 9.6K & 141 & 0.1M & 2.1M & 1.9K & 141 & 0.3M & 3.8M & 3.8K & 141 & 0.6M & 6.2M & 6.8K & 141 & 0.6M & 6.3M & 7.1K & 141 \\
\hline
\multirow{9}{*}{DISK} & bicycle & 24.8K & 0.6M & 84 & 4 & 49.7K & 1.1M & 3.8K & 103 & 99.3K & 2.0M & 1.2K & 163 & 0.2M & 3.7M & 2.9K & 194 & 0.4M & 6.6M & 5.6K & 194 & 0.8M & 10.9M & 1.0K & 194 & 1.6M & 14.5M & 1.9K & 194 \\
 & bonsai & 37.4K & 1.4M & 4.3K & 247 & 74.8K & 2.6M & 9.2K & 252 & 0.1M & 4.8M & 1.6K & 279 & 0.3M & 8.7M & 3.1K & 292 & 0.6M & 15.6M & 4.9K & 292 & 1.2M & 24.8M & 8.9K & 292 & 1.8M & 29.4M & 1.4K & 292 \\
 & counter & 30.7K & 0.9M & 30 & 2 & 61.4K & 1.6M & 9.0K & 240 & 0.1M & 2.8M & 1.7K & 239 & 0.2M & 5.1M & 3.3K & 240 & 0.5M & 8.7M & 6.1K & 240 & 1.0M & 13.5M & 1.1K & 240 & 1.7M & 17.3M & 1.8K & 240 \\
 & flowers & 22.1K & 0.6M & 66 & 2 & 44.3K & 1.1M & 546 & 14 & 88.6K & 2.2M & 6.4K & 80 & 0.2M & 4.0M & 2.5K & 159 & 0.4M & 7.2M & 5.4K & 172 & 0.7M & 11.8M & 1.0K & 173 & 1.4M & 15.1M & 1.8K & 173 \\
 & garden & 23.7K & 0.6M & 3.2K & 179 & 47.4K & 1.1M & 6.7K & 185 & 94.7K & 2.0M & 1.3K & 185 & 0.2M & 3.7M & 2.4K & 185 & 0.4M & 6.6M & 4.3K & 185 & 0.8M & 11.1M & 7.7K & 185 & 1.5M & 15.9M & 1.4K & 185 \\
 & kitchen & 35.7K & 1.0M & 2.5K & 171 & 71.4K & 1.8M & 7.9K & 279 & 0.1M & 3.4M & 1.6K & 279 & 0.3M & 6.2M & 2.8K & 279 & 0.6M & 11.3M & 4.5K & 279 & 1.1M & 20.5M & 8.1K & 279 & 2.2M & 32.2M & 1.3K & 279 \\
 & room & 39.8K & 1.4M & 1.3K & 135 & 79.6K & 2.5M & 8.1K & 307 & 0.2M & 4.6M & 1.5K & 306 & 0.3M & 8.2M & 2.7K & 302 & 0.6M & 14.3M & 5.4K & 311 & 1.3M & 23.5M & 1.0K & 311 & 2.0M & 32.0M & 1.7K & 311 \\
 & stump & 16.0K & 0.3M & 30 & 2 & 32.0K & 0.5M & 415 & 12 & 64.0K & 1.0M & 4.9K & 68 & 0.1M & 1.8M & 1.6K & 118 & 0.3M & 3.2M & 3.4K & 125 & 0.5M & 5.3M & 6.6K & 125 & 1.0M & 6.7M & 1.2K & 125 \\
 & treehill & 18.0K & 0.3M & 50 & 2 & 36.1K & 0.6M & 1.0K & 26 & 72.2K & 1.2M & 7.7K & 105 & 0.1M & 2.2M & 2.1K & 141 & 0.3M & 3.9M & 3.9K & 141 & 0.6M & 6.6M & 7.0K & 141 & 1.1M & 9.5M & 1.2K & 141 \\
\hline
\multirow{9}{*}{\begin{tabular}[c]{@{}c@{}}DoG- \\ HardNet\end{tabular}} & bicycle & 16.5K & 10.7K & 41 & 2 & 34.1K & 35.0K & 903 & 45 & 70.3K & 0.1M & 3.3K & 108 & 0.1M & 0.5M & 8.6K & 176 & 0.3M & 1.1M & 2.0K & 193 & 0.6M & 2.1M & 4.6K & 194 & 1.3M & 4.1M & 9.8K & 194 \\
 & bonsai & 27.9K & 73.5K & 204 & 17 & 57.3K & 0.3M & 6.1K & 292 & 0.1M & 1.1M & 1.3K & 292 & 0.2M & 2.1M & 2.9K & 292 & 0.5M & 3.7M & 5.6K & 292 & 0.9M & 6.9M & 1.0K & 292 & 1.9M & 13.6M & 2.0K & 292 \\
 & counter & 22.5K & 47.1K & 34 & 2 & 46.8K & 0.2M & 3.4K & 149 & 97.1K & 0.9M & 1.2K & 240 & 0.2M & 1.7M & 2.5K & 240 & 0.4M & 3.1M & 5.4K & 240 & 0.8M & 5.7M & 1.1K & 240 & 1.7M & 10.4M & 2.0K & 240 \\
 & flowers & 14.5K & 1.3K & 28 & 2 & 29.9K & 11.8K & 79 & 3 & 61.8K & 0.1M & 144 & 4 & 0.1M & 0.5M & 3.5K & 83 & 0.3M & 1.0M & 2.3K & 173 & 0.6M & 1.9M & 5.4K & 173 & 1.1M & 3.8M & 9.9K & 173 \\
 & garden & 17.0K & 51.0K & 969 & 101 & 34.6K & 0.2M & 3.7K & 184 & 71.1K & 0.5M & 7.7K & 185 & 0.1M & 1.0M & 1.6K & 185 & 0.3M & 2.0M & 3.3K & 185 & 0.6M & 3.9M & 7.0K & 185 & 1.2M & 7.3M & 1.4K & 185 \\
 & kitchen & 25.1K & 91.1K & 800 & 82 & 51.1K & 0.2M & 3.5K & 171 & 0.1M & 0.6M & 6.9K & 171 & 0.2M & 1.5M & 2.2K & 279 & 0.4M & 2.8M & 4.5K & 279 & 0.9M & 4.8M & 8.6K & 279 & 1.7M & 8.5M & 1.8K & 279 \\
 & room & 29.3K & 0.2M & 30 & 2 & 61.6K & 0.6M & 6.2K & 311 & 0.1M & 1.7M & 1.2K & 311 & 0.3M & 3.3M & 2.5K & 311 & 0.5M & 6.3M & 5.3K & 311 & 1.1M & 11.2M & 1.0K & 311 & 1.7M & 17.7M & 1.6K & 311 \\
 & stump & 10.0K & 1.9K & 31 & 2 & 20.7K & 6.9K & 46 & 2 & 43.1K & 52.6K & 180 & 7 & 90.4K & 0.2M & 3.0K & 71 & 0.2M & 0.4M & 9.2K & 116 & 0.4M & 0.9M & 2.2K & 125 & 0.8M & 1.7M & 5.4K & 125 \\
 & treehill & 11.7K & 1.1K & 26 & 2 & 24.2K & 8.1K & 74 & 3 & 50.2K & 60.5K & 394 & 12 & 0.1M & 0.2M & 3.7K & 77 & 0.2M & 0.5M & 1.4K & 141 & 0.4M & 1.1M & 3.2K & 141 & 0.9M & 2.3M & 6.6K & 141 \\
\hline
\multirow{9}{*}{SIFT} & bicycle & 16.5K & 37.0K & 32 & 2 & 34.1K & 0.3M & 738 & 38 & 70.3K & 0.7M & 3.3K & 104 & 0.1M & 1.4M & 8.9K & 140 & 0.3M & 2.9M & 2.2K & 181 & 0.6M & 5.6M & 4.3K & 192 & 1.3M & 10.7M & 7.8K & 193 \\
 & bonsai & 27.9K & 0.5M & 139 & 11 & 57.3K & 1.3M & 4.5K & 184 & 0.1M & 2.5M & 1.4K & 269 & 0.2M & 4.1M & 3.2K & 292 & 0.5M & 6.8M & 5.9K & 292 & 0.9M & 12.3M & 1.0K & 292 & 1.9M & 24.1M & 1.9K & 292 \\
 & counter & 22.5K & 0.3M & 34 & 2 & 46.8K & 0.9M & 2.7K & 115 & 97.1K & 1.7M & 1.3K & 239 & 0.2M & 3.3M & 2.7K & 240 & 0.4M & 6.2M & 5.3K & 240 & 0.8M & 11.6M & 1.0K & 240 & 1.7M & 21.4M & 1.9K & 240 \\
 & flowers & 14.5K & 33.9K & 38 & 2 & 29.9K & 0.3M & 70 & 3 & 61.8K & 0.6M & 172 & 4 & 0.1M & 1.3M & 0 & 3 & 0.3M & 2.5M & 9.5K & 73 & 0.6M & 4.9M & 4.0K & 166 & 1.1M & 9.5M & 7.3K & 172 \\
 & garden & 17.0K & 0.1M & 1.0K & 100 & 34.6K & 0.5M & 3.7K & 181 & 71.1K & 0.9M & 8.1K & 185 & 0.1M & 1.8M & 1.8K & 185 & 0.3M & 3.4M & 3.7K & 185 & 0.6M & 6.5M & 7.0K & 185 & 1.2M & 12.2M & 1.3K & 185 \\
 & kitchen & 25.1K & 0.2M & 90 & 4 & 51.1K & 0.6M & 3.6K & 172 & 0.1M & 1.3M & 1.2K & 279 & 0.2M & 2.5M & 2.5K & 279 & 0.4M & 4.5M & 5.0K & 279 & 0.9M & 7.7M & 9.5K & 279 & 1.7M & 14.3M & 1.8K & 279 \\
 & room & 29.3K & 0.7M & 1.2K & 127 & 61.6K & 1.6M & 6.6K & 311 & 0.1M & 3.2M & 1.4K & 311 & 0.3M & 6.2M & 2.6K & 306 & 0.5M & 11.6M & 5.7K & 311 & 1.1M & 20.1M & 1.1K & 311 & 1.7M & 31.2M & 1.7K & 311 \\
 & stump & 10.0K & 8.4K & 31 & 2 & 20.7K & 0.1M & 36 & 2 & 43.1K & 0.3M & 1.5K & 52 & 90.4K & 0.6M & 4.2K & 83 & 0.2M & 1.2M & 1.3K & 103 & 0.4M & 2.5M & 2.6K & 115 & 0.8M & 4.9M & 4.8K & 125 \\
 & treehill & 11.7K & 12.9K & 25 & 2 & 24.2K & 0.2M & 67 & 3 & 50.2K & 0.4M & 481 & 15 & 0.1M & 0.7M & 4.0K & 59 & 0.2M & 1.5M & 1.4K & 117 & 0.4M & 3.1M & 3.1K & 141 & 0.9M & 6.0M & 5.8K & 141 \\
\hline
\multirow{9}{*}{SuperPoint} & bicycle & 24.8K & 86.6K & 27 & 2 & 49.7K & 0.4M & 4.7K & 186 & 99.3K & 0.8M & 1.0K & 194 & 0.2M & 1.4M & 2.1K & 194 & 0.4M & 2.3M & 4.2K & 194 & 0.6M & 2.8M & 6.2K & 194 & 0.6M & 2.8M & 6.3K & 194 \\
 & bonsai & 37.4K & 0.3M & 3.4K & 217 & 74.8K & 1.2M & 8.6K & 292 & 0.1M & 2.3M & 1.6K & 292 & 0.3M & 4.1M & 3.1K & 292 & 0.6M & 5.8M & 6.1K & 292 & 0.9M & 6.8M & 9.3K & 292 & 0.9M & 6.8M & 9.4K & 292 \\
 & counter & 30.7K & 0.3M & 28 & 2 & 61.4K & 0.8M & 7.6K & 240 & 0.1M & 1.6M & 1.6K & 240 & 0.2M & 3.1M & 3.2K & 240 & 0.5M & 4.8M & 6.2K & 240 & 0.9M & 5.8M & 1.1K & 240 & 0.9M & 5.9M & 1.1K & 240 \\
 & flowers & 22.1K & 79.3K & 34 & 2 & 44.3K & 0.4M & 26 & 2 & 88.6K & 0.7M & 8.5K & 173 & 0.2M & 1.2M & 2.3K & 173 & 0.4M & 2.1M & 4.6K & 173 & 0.6M & 2.8M & 7.5K & 173 & 0.7M & 2.8M & 7.7K & 173 \\
 & garden & 23.7K & 0.2M & 2.7K & 185 & 47.4K & 0.6M & 5.7K & 185 & 94.7K & 1.1M & 1.1K & 185 & 0.2M & 2.1M & 2.2K & 185 & 0.4M & 3.4M & 4.5K & 185 & 0.6M & 4.5M & 7.5K & 185 & 0.6M & 4.6M & 7.6K & 185 \\
 & kitchen & 35.7K & 0.4M & 2.4K & 172 & 71.4K & 1.2M & 6.8K & 279 & 0.1M & 2.2M & 1.3K & 279 & 0.3M & 3.7M & 2.5K & 279 & 0.6M & 4.9M & 5.3K & 279 & 0.9M & 5.7M & 8.2K & 279 & 0.9M & 5.9M & 8.2K & 279 \\
 & room & 39.8K & 0.6M & 2.3K & 167 & 79.6K & 1.6M & 8.0K & 311 & 0.2M & 3.3M & 1.4K & 311 & 0.3M & 5.8M & 2.9K & 311 & 0.6M & 8.3M & 5.6K & 311 & 0.8M & 9.3M & 7.1K & 311 & 0.8M & 9.3M & 6.9K & 311 \\
 & stump & 16.0K & 35.0K & 24 & 2 & 32.0K & 0.2M & 50 & 2 & 64.0K & 0.3M & 3.0K & 95 & 0.1M & 0.6M & 9.3K & 125 & 0.3M & 1.0M & 2.3K & 125 & 0.5M & 1.5M & 4.0K & 125 & 0.5M & 1.5M & 4.1K & 125 \\
 & treehill & 18.0K & 66.3K & 34 & 2 & 36.1K & 0.2M & 939 & 41 & 72.2K & 0.4M & 6.0K & 141 & 0.1M & 0.8M & 1.3K & 141 & 0.3M & 1.2M & 2.6K & 141 & 0.4M & 1.5M & 3.3K & 141 & 0.4M & 1.5M & 3.4K & 141 \\
\hline
\multirow{9}{*}{XFeat} & bicycle & 24.8K & 0.3M & - & - & 49.7K & 0.7M & 52 & 2 & 99.3K & 1.2M & 1.0K & 185 & 0.2M & 2.2M & 2.4K & 193 & 0.4M & 4.1M & 4.8K & 193 & 0.8M & 7.3M & 9.1K & 194 & 1.6M & 12.9M & 1.7K & 194 \\
 & bonsai & 37.4K & 0.8M & 37 & 2 & 74.8K & 1.5M & 1.1K & 292 & 0.1M & 2.6M & 2.0K & 289 & 0.3M & 4.6M & 4.0K & 292 & 0.6M & 7.7M & 7.0K & 292 & 1.2M & 12.3M & 1.3K & 292 & 2.4M & 18.4M & 2.3K & 292 \\
 & counter & 30.7K & 0.6M & 14 & 2 & 61.4K & 1.1M & 9.6K & 240 & 0.1M & 2.0M & 2.0K & 240 & 0.2M & 3.5M & 3.8K & 240 & 0.5M & 6.1M & 7.0K & 240 & 1.0M & 9.9M & 1.3K & 240 & 2.0M & 14.7M & 2.3K & 240 \\
 & flowers & 22.1K & 0.3M & - & - & 44.3K & 0.6M & 33 & 2 & 88.6K & 1.2M & 70 & 2 & 0.2M & 2.1M & 36 & 2 & 0.4M & 3.8M & 2.6K & 138 & 0.7M & 6.9M & 7.5K & 173 & 1.4M & 12.2M & 1.5K & 173 \\
 & garden & 23.7K & 0.4M & 25 & 2 & 47.4K & 0.7M & 55 & 2 & 94.7K & 1.4M & 1.6K & 185 & 0.2M & 2.6M & 3.2K & 185 & 0.4M & 4.8M & 6.1K & 185 & 0.8M & 8.9M & 1.1K & 185 & 1.5M & 16.2M & 2.0K & 185 \\
 & kitchen & 35.7K & 0.8M & 5.4K & 279 & 71.4K & 1.5M & 1.0K & 279 & 0.1M & 2.7M & 1.9K & 279 & 0.3M & 4.7M & 3.3K & 279 & 0.6M & 8.1M & 5.8K & 279 & 1.1M & 13.1M & 1.0K & 279 & 2.3M & 19.9M & 1.8K & 279 \\
 & room & 39.8K & 0.9M & 32 & 2 & 79.6K & 1.8M & 1.0K & 311 & 0.2M & 3.2M & 1.8K & 311 & 0.3M & 5.7M & 3.4K & 305 & 0.6M & 9.8M & 6.2K & 311 & 1.3M & 16.2M & 1.1K & 310 & 2.5M & 25.2M & 2.2K & 311 \\
 & stump & 16.0K & 0.2M & 25 & 2 & 32.0K & 0.3M & 36 & 2 & 64.0K & 0.6M & 49 & 2 & 0.1M & 1.1M & 3.0K & 36 & 0.3M & 2.0M & 1.7K & 105 & 0.5M & 3.6M & 3.9K & 124 & 1.0M & 6.4M & 7.5K & 125 \\
 & treehill & 18.0K & 0.2M & - & - & 36.1K & 0.4M & 27 & 2 & 72.2K & 0.7M & 1 & 2 & 0.1M & 1.2M & 8.2K & 81 & 0.3M & 2.2M & 3.0K & 141 & 0.6M & 4.0M & 5.6K & 141 & 1.2M & 7.1M & 1.0K & 141 \\
\hline
\end{tabular}
}
\caption{3D reconstruction stats on \textbf{MipNeRF 360} dataset using \textbf{COLMAP} mapper and \textbf{Mutual Nearest Neighbors} matcher}
\label{tab:reconsStats_Mip-NeRF360_colmap_SMNNk-opt}
\end{table*}

\end{appendix}

\end{document}